\documentclass[acmsmall]{acmart}

\usepackage{gensymb}
\usepackage{amsmath,bbm,bm}
\usepackage[utf8]{inputenc} % allow utf-8 input
\usepackage[T1]{fontenc}    % use 8-bit T1 fonts
\usepackage{url}            % simple URL typesetting
\usepackage{booktabs}       % professional-quality tables
\usepackage{amsfonts}       % blackboard math symbols
\usepackage{nicefrac}       % compact symbols for 1/2, etc.
\usepackage{microtype}      % microtypography
\usepackage{graphicx}
\usepackage[font=footnotesize]{subcaption}
\usepackage{multirow}
\usepackage{comment}
\usepackage{color}
\usepackage{etoolbox}
\PassOptionsToPackage{table,xcdraw}{xcolor}

\usepackage[flushleft]{threeparttable} %QC
\usepackage{siunitx} %QC

\newtoggle{comments}
\settoggle{comments}{true} %set true to show all comments, set false to hide all comments
\iftoggle{comments}{
  \newcommand {\alberto}[1]{{\color{orange}{~Alberto: #1}\normalfont}}
  \newcommand {\bjin}[1]{{\color{blue}{~Baihong: #1}\normalfont}}
  \newcommand {\qiushi}[1]{{\color{violet}{~Qiushi: #1}\normalfont}}
  \newcommand {\yingshui}[1]{{\color{green}{~Yingshui: #1}\normalfont}}
  \newcommand {\red}[1]{{\color{red}{#1}\normalfont}}
}{
  \newcommand {\alberto}[1]{{}}
  \newcommand {\bjin}[1]{{}}
  \newcommand {\qiushi}[1]{{}}
  \newcommand {\yingshui}[1]{{}}
  \newcommand {\red}[1]{{}}
}

\newcommand{\expctover}[2]{\mathbb{E}_{#1}\!\left[#2\right]}

\usepackage{acronym}
\acrodef{i.d.}{in-distribution}
\acrodef{o.o.d.}{out-of-distribution}
\acrodef{FDD}{Fault Detection and Diagnosis}
\acrodef{LDA}{Linear Discriminant Analysis}
\acrodef{ML}{Machine Learning}
\acrodef{CFAR}{Constant False Alarm Rate}
\acrodef{RF}{Random Forest}
\acrodef{LR}{Logistic Regression}
\acrodef{NN}{Neural Network}
\acrodef{DT}{Decision Tree}
\acrodef{OC-SVM}{One-Class Support Vector Machine}
\acrodef{AE}{Autoencoder}

\acrodef{FPR}{False Positive Rate}
\acrodef{FNR}{False Negative Rate}
\acrodef{SL}{Severity Level}
\acrodef{ROC}{Receiver Operating Characteristic}
\acrodef{CI}{confidence interval}
\acrodef{IS}{Intermediate-Severity}
\acrodef{RBF}{Radial Basis Function}
\acrodef{PCA}{Principle Component Analysis}
\acrodef{LSTM}{Long Short-Term Memory}
\acrodef{RUL}{Remaining Useful Life}
\acrodef{AHU}{Air Handling Unit}
\acrodef{FNP}{False Negative Precision}
\acrodef{AUC}{Area Under Curve}
\acrodef{FDD}{Fault detection}
\acrodef{CNN}{Convolutional Neural Network}
\acrodef{CPS}{Cyber-Physical System}
\acrodef{SACV}{Stratification-Aware Cross-Validation}
\acrodef{HVAC}{Heating, Ventilation and Air Conditioning}
\acrodef{HIF}{High Impedance Fault}
\acrodef{VAV}{Variable Air Volume}

\AtBeginDocument{%
  \providecommand\BibTeX{{%
    \normalfont B\kern-0.5em{\scshape i\kern-0.25em b}\kern-0.8em\TeX}}}

\thanks{
  $^\ast$Y.~Tan and B.~Jin contributed equally to this paper.\\
  This work is supported by the National Research Foundation of Singapore through a grant to the Berkeley Education Alliance for Research in Singapore (BEARS) for the Singapore-Berkeley Building Efficiency and Sustainability in the Tropics (SinBerBEST) program, and by the Defence Science \& Technology Agency (DSTA) of Singapore.
}

\begin{document}

%%
%% The "title" command has an optional parameter,
%% allowing the author to define a "short title" to be used in page headers.
\title[Generalizing Fault Detection Against Domain Shifts Using SACV]{Generalizing Fault Detection Against Domain Shifts Using Stratification-Aware Cross-Validation}

%%
%% The "author" command and its associated commands are used to define
%% the authors and their affiliations.
%% Of note is the shared affiliation of the first two authors, and the
%% "authornote" and "authornotemark" commands
%% used to denote shared contribution to the research.
\author{Yingshui Tan}
\affiliation{%
  \institution{University of California, Berkeley}
%   \streetaddress{Cory Hall}
  \city{Berkeley, CA~94720}
%   \state{California}
  \country{USA}
  \postcode{94720}
}
\email{tys@eecs.berkeley.edu}

\author{Baihong Jin}
\affiliation{%
  \institution{University of California, Berkeley}
%   \streetaddress{Cory Hall}
  \city{Berkeley, CA~94720}
  \country{USA}}
\email{bjin@eecs.berkeley.edu}

\author{Qiushi Cui}
\affiliation{%
  \institution{Arizona State University}
  \city{Tempe, AZ~85281}
  \country{USA}
}
\email{qiushi.cui@asu.edu}

\author{Xiangyu Yue}
\affiliation{%
  \institution{University of California, Berkeley}
%   \streetaddress{Soda Hall}
\city{Berkeley, CA~94720}
  \country{USA}}
\email{xyyue@eecs.berkeley.edu}

\author{Alberto Sangiovanni Vincentelli}
\affiliation{%
  \institution{University of California, Berkeley}
%   \streetaddress{Cory Hall}
  \city{Berkeley, CA~94720}
  \country{USA}
%   \postcode{94720}
}
\email{alberto@eecs.berkeley.edu}

%%
%% By default, the full list of authors will be used in the page
%% headers. Often, this list is too long, and will overlap
%% other information printed in the page headers. This command allows
%% the author to define a more concise list
%% of authors' names for this purpose.
\renewcommand{\shortauthors}{Tan, Jin, Cui, Yue and Sangiovanni Vincentelli}

%%
%% The abstract is a short summary of the work to be presented in the
%% article.
\begin{abstract}
\ac{ML}-based fault detection techniques have attracted much interest in many \ac{CPS} application domains, including power grids and smart buildings. An often overlooked objective when training \ac{ML}-based fault detectors is the ability of generalizing to unseen environments (e.g., fault types, and locations). Indeed, domain shifts from the training distribution can have an adverse effect on the test-time performance of the trained \ac{ML} models.  In this paper we show that the risks due to domain shifts can be mitigated via a carefully-designed cross-validation process.
\end{abstract}

%%
%% The code below is generated by the tool at http://dl.acm.org/ccs.cfm.
%% Please copy and paste the code instead of the example below.
%%


%%
%% Keywords. The author(s) should pick words that accurately describe
%% the work being presented. Separate the keywords with commas.

%% A "teaser" image appears between the author and affiliation
%% information and the body of the document, and typically spans the
%% page.
% \begin{teaserfigure}
%   \includegraphics[width=\textwidth]{sampleteaser}
%   \caption{Seattle Mariners at Spring Training, 2010.}
%   \Description{Enjoying the baseball game from the third-base
%   seats. Ichiro Suzuki preparing to bat.}
%   \label{fig:teaser}
% \end{teaserfigure}

%%
%% This command processes the author and affiliation and title
%% information and builds the first part of the formatted document.
\maketitle

% If your title is lengthy, you must define a short version to be used
% in the page headers, to prevent overlapping text. The \verb|title|
% command has a ``short title'' parameter:
% \begin{verbatim}
%   \title[short title]{full title}
% \end{verbatim}

%%
%% The acknowledgments section is defined using the "acks" environment
%% (and NOT an unnumbered section). This ensures the proper
%% identification of the section in the article metadata, and the
%% consistent spelling of the heading.

\section{Introduction}\label{sec:intro}
\acresetall

Fault detection methods can be broadly categorized~\cite{Dai2013from,Gao2015asurvey,jia2018design} into i)~\textit{model-based}, ii)~\textit{signal-based}, and iii)~\textit{data-driven} approaches. Model-based methods~\cite{isermann2005model} rely on building explicit physical models at the device levels and use correlation tests on the input-output data to detect faults~\cite{Qin2005afault,Saththasivam2008predictive,Ru2008variable}, which can be effective if high-fidelity physical models are available; however, developing detailed models is a time-consuming and daunting process, especially for complex and highly diversified \acp{CPS} such as buildings. Authors of~\cite{Dumidu2014mining} point out that model-based methods are not as practical as data-driven methods in terms of applying the fault detection techniques to real buildings. A recent study by Hu~et~al.~\cite{zhou2017quantitative} shows that a data-driven method outperforms a physical-based method in identifying the thermal dynamics of an office environment.

Signal-based fault detection methods~\cite{Shun2014amodel,harihara2003signal,chong2011signal} aim to find indicative sensor measurement signatures to detect faults. Again, finding well-performing signatures oftentimes requires not only advanced signal processing and transformation techniques but also deep insights into the systems under study, which can be a daunting task especially for complex \acp{CPS}. For model developers, it is highly appealing to have an end-to-end approach that can directly learn from data and produce well-performing \ac{ML} models. However, the domain shift~\cite{luo2019taking} (a.k.a. distribution shift~\cite{sun2020test}, concept drift~\cite{vorburger2006entropy}) problem presents a major challenge for the adoption of data-driven methods in practice. Although models trained with supervised learning tend to perform well on known (in-distribution) data patterns, the unseen, \ac{o.o.d.} data may lead to unexpected prediction behaviors. In order to train a well-performing model, large amount of labeled, diversified data is typically needed, which is not always easy to obtain, especially for fault detection tasks where the fault data usually constitute only a small fraction of the collected data.

In fault detection applications, the prediction task is usually to differentiate a ``normal'' or fault-free class (hereinafter referred to as the negative class) from a set of fault classes (hereinafter referred to as the the positive class), which is often cast as a binary classification problem. In other words, the positive class is often \textit{stratified}, which may cause severe consequences especially for safety critical applications such as fault detection and medical diagnosis. Worse still, if some strata are missing from the training distribution but appear in the test distribution (a.k.a. \ac{o.o.d.}), regular \ac{ML} training pipelines offer no guarantee on such \ac{o.o.d.} data. In other words, many false negative decisions may occur. For example, if an unseen fault type occurs or if an industrial machine is operating under a different environment, a fault detection model may fail to identify faulty conditions.

The unseen nature of domain shifts presents a major challenge to training generalizable \ac{ML} models, especially in the lack of domain knowledge. On the other hand, we wish to make best use of available data (although not comprehensive enough to capture all possible variations) to obtain \ac{ML} models as robust as possible against domain shifts. Our solution is to use a stratification-aware cross-validation strategy during model selection, which helps \textit{reject} models that are not robust even on \ac{i.d.} data.
We believe this strategy is an easy-to-use recipe for developing supervised ensemble fault detection models that are more immune to the above-mentioned domain shift phenomena. We summarize our contributions in this paper as follows:
\begin{itemize}
    \item We propose a \textit{stratification-aware} cross-validation strategy for training \ac{ML} models on stratified data to encourage improved robustness against unknown domain shift in test sets.
    \item The efficacy of the proposed method is demonstrated in three case studies: a power distribution system, a commercial building chiller system, and a commercial building \ac{AHU} system. The results showed that our stratification-aware cross-validation strategy leads to substantial improvement on detecting \ac{o.o.d.} faults.
    \item On top of that, we applied ensemble learning in an uncertainty-informed fault detection framework to identify false negatives which demonstrated significant performance boost when domain experts can help correct the decisions on the high-uncertainty negative examples identified by our algorithm. 
\end{itemize}
The remainder of this paper is organized as follows. We formulate the fault detection and diagnosis problems in Sec.~\ref{sec:formulation}. Next, in Sec.~\ref{sec:methodology}, we  describe in details our methodology. The power network dataset used in our empirical study are briefly described in Sec.~\ref{sec:dataset}, and in Sec.~\ref{sec:experiment} experimental results will be presented. In Sec.~\ref{sec:related-work}, we review related research topics found in the literature. We summarize the findings in this paper and discuss future work in Sec.~\ref{sec:conclusion}.

\section{Background and Problem Formulation}\label{sec:formulation}

\subsection{Generalizable Fault Detection on Stratified Data}\label{sec:generalizable-fault-detection}
We formulate the fault detection problem under a \textit{binary classification} setting.  
A fault detection model aims at differentiating the fault conditions from the normal condition by monitoring the system state. Let ${z}\in\{0,1\}$ represent the ground-truth label of system state $\bm{x}\in\mathbb{R}^d$, where $z=0$ stands for the normal condition and $z=1$ the fault condition. A \textit{fault detector} is a rule or function that predicts a label $\hat{z}\in\{0,1\}$ given input $x$. 
Let $\mathcal{X}$ be the set of data points, and $\mathcal{M}$ be a model class of classification models. Suppose a classification model $M\in\mathcal{M}$ defines an \textit{anomaly score} function $s^M:\mathcal{X}\rightarrow\mathbb{R}$ that characterizes how likely $x$ corresponds to a fault state; a larger $s^M(x)$ implies a higher chance of a data point $\bm{x}$ being a fault. The classifier's decision on whether or not $x$ corresponds to a fault can be made by introducing a \textit{decision threshold} $\tau^M$ to dichotomies the anomaly score $s^M(x)$. We can define the classifier's predicted label 
\begin{align*}
    \hat{z} = \mathbbm{1}\{s^M(x) > \tau^M\}.
\end{align*}
For evaluating the performance of $M$, we can define the \ac{FNR} and \ac{FPR} of the model $M$ on the test data distribution as follows:
\begin{align}
    \text{FNR}(s^M,\tau^M) &= \expctover{}{\hat{z}=0 \mid z=0},\\
    \text{FPR}(s^M,\tau^M) &= \expctover{}{\hat{z}=1 \mid z=1}.
\end{align}
Let $\mathcal{X}^\text{dev}$ be the subset of labeled training data points that are available to us at training time. Ideally, the goal is to learn an anomaly score function $s^\ast$ by minimizing the classification error on $\mathcal{X}^\text{dev}$, and then decide a corresponding threshold $\tau^\ast$, such that the resulting model $M\doteq(s^\ast,\tau^\ast)$ can optimize both the \ac{FNR} and the \ac{FPR} on the (unseen) test data distribution $\mathcal{D}_\text{test}$.

Different from the traditional assumption that the training set and the test set are sampled from the same distribution, in this paper we assume that the test data distribution $\mathcal{D}_\text{test}$ not only comprises of the \ac{i.d.} data $\mathcal{D}_\text{test}^\text{i.d.}$ but also the \ac{o.o.d.} data $\mathcal{D}_\text{test}^\text{o.o.d.}$ with domain shift. Our goal is to train a binary classification model $\mathcal{M}$ using the development set data $\mathcal{D}_\text{dev}$ such that $\mathcal{M}$ achieves the best precision-recall trade-off on the test data $\mathcal{D}_\text{test}$, including both the \ac{i.d.} and the \ac{o.o.d.} portions. In this study, we assume that the data distributions we will be dealing with follow a stratified structure; in other words, the fault data are structured as a set of subgroups (strata). Suppose that the development set data consists of $K^\text{i.d.}$ subgroups in total; the same data subgroups also make up the \ac{i.d.} test set $\mathcal{D}_\text{test}^\text{i.d.}$. The \ac{o.o.d.} test set $\mathcal{D}_\text{test}^\text{o.o.d.}$ contains $K^\text{o.o.d.}$ subgroups that do not appear in the development set.

\paragraph{Setting a proper detection threshold $\tau$}
In real practice, one always has to make a trade-off between \ac{FNR} and \ac{FPR} when determining a proper value for the decision threshold $\tau$ (a.k.a. the operating point). One approach for determining $\tau$ is to directly set it to a predefined value (e.g., $0.5$, an often used threshold value). This is usually not a bad approach, if most data points are well separated by the classifier and receive anomaly scores $s(x)$ that are close to either $0$ or $1$. In fact, under such scenarios, it will not make a huge difference to pick a $\tau$ value other than $0.5$ as long as similar \ac{FPR} and \ac{FNR} can be achieved on the development set (\ac{i.d.} data). Thanks to the flexibility, we can pick a lower $\tau$ without affecting the classifier's performance on the \ac{i.d.} data, which will give us a classifier with higher sensitivity that is better at telling unseen anomalies. In practice, we can select $\tau$ such that the \ac{FPR} on the development set is under a predefined level $q$. This approach is also known as \ac{CFAR} detection scheme~\cite{richards2005fundamentals} in radar applications.

\begin{figure}[t]
    \centering
    \includegraphics[width=\linewidth]{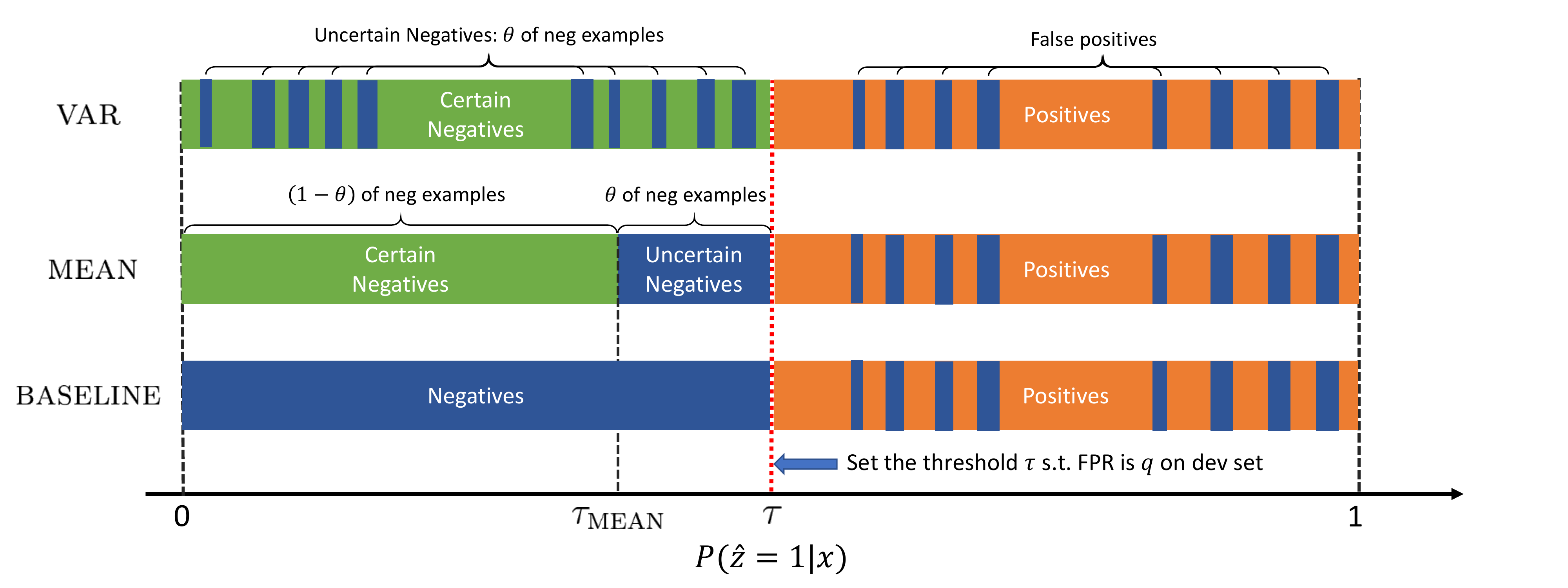}
    \caption{Illustration showing the concepts in an uncertainty-informed decision framework.}
    \label{fig:decision}
\end{figure}

The above-mentioned decision scheme is illustrated in Fig.~\ref{fig:decision} as the \textsc{baseline} scheme. The detection threshold $\tau$ identifies positive examples (shown as yellow in the diagram) that are likely to be fault states. Although there will usually be false positives among these identified positive examples, the false negative decisions can be however a more serious concern in anomaly detection since they are anomalous instances mistaken as normal. Following the approaches proposed in our previous works~\cite{jin2020ensemble,Tan2020ExploitingUF}, we will utilize the decision uncertainty information from ensemble classifiers to identify potential false negatives in an \textit{uncertainty-informed} decision scheme as to be described below.

\subsection{Uncertainty-Informed Fault Detection}\label{sec:uncertainty-fault-detection}
To further improve the fault detection performance, we adopt an \textit{uncertainty-informed} diagnostic scheme~\cite{Tan2020ExploitingUF,Jin2020incipient} that exploits prediction uncertainties in a human-AI collaborative setting. Under this scheme, the trained model $\mathcal{M}$ described in Sec.~\ref{sec:generalizable-fault-detection} is first used to screen the data, and detect as positive the cases that are likely to be anomalous. These positive cases will then be referred to human experts (e.g., technicians or maintenance specialists) for further inspection; human experts then will confirm these cases as positive and take necessary maintenance or repair actions if they agree with the \ac{ML} model's decisions.

To identify high-uncertainty examples that are likely to be false negative decisions, we use an \textit{uncertainty metric} $U$ to rank the negative examples\footnote{Examples that are classified as negative by a classification model, i.e. $\{x_i\,\vert\,\hat{z}_i=0\}$.}. To ease later exposition, here we suppose that an ensemble model of size $T$ is used, and denote the predictions of individual ensemble members on $x_i$ as $y_i^{(1)},y_i^{(2)},\ldots,y_i^{(T)}$. The uncertainty metric $U: \mathbb{R}^K\rightarrow \mathbb{R}$ takes as input the ensemble predictions $\{\hat{y}_i^{(k)}\}$ on $x_i$, and outputs an real-valued \textit{uncertainty score} $u(x_i)\doteq U\left(y_i^{(1)},y_i^{(2)},\ldots,y_i^{(T)}\right)$.
To resolve a dichotomy between ``uncertain'' and ``certain'', we introduce a threshold $\tilde{u}$ on $u(x)$: if $u(x)>\tilde{u}$ then $x$ is deemed an uncertain input example and otherwise a certain one. We then need external resources such as human experts to inspect these uncertain negatives and determine their true states; however, due to budget constraints such resources are often limited. Therefore, we need to control the fraction of uncertain negatives. We define the \textit{uncertain negative ratio} as the fraction of uncertain examples among negative examples, and bound the ratio to be below a level of $\theta$ on the development set.
To evaluate how the identified uncertain negatives overlap with the actual false negatives, we use the following performance measure.
The \begin{definition}[False Negative Precision~\cite{Jin2020incipient}]
We define the false negative precision to be the fraction of false negatives among identified uncertain negative inputs by a given uncertainty metric $U$ and uncertain negative ratio $\theta$. Written in mathematical form,

\begin{align}
    \text{FN-precision}(U,q) \doteq \frac{\left\vert\left\{x_i\,\vert\,i\in\mathcal{I}^{-}_{q},z_i=1\right\}\right\vert}
    {\left\vert\mathcal{I}^{-}_{q}\right\vert}\in[0,1],
\end{align}
where $\mathcal{I}^{-}_{q}$ is the index set of identified uncertain negative examples.
\end{definition}

The FN-precision metric can be interpreted as the ratio of identified uncertain examples being actual false negatives. The higher the FN-precision value, the fewer false alarms are raised by $(U,q)$. We can similarly define a ``false negative recall'' metric that measures the fraction of false negatives identified by the algorithm; however, in this study we choose to directly report the total number of false negatives instead. In our empirical study to be described later, we will compare two  commonly used uncertainty metrics, \textsc{mean} and \textsc{var}, to see which one is more suitable for the fault detection task under domain shift.

\section{Methodology}\label{sec:methodology}
\textit{Validation} is a classic and almost a must-have procedure for model selection in a modern \ac{ML} pipeline. The goal of validation is to obtain an accurate estimate of a trained model's prediction performance on the test set, under the typical assumption that the training set and the test set are sampled from the same data distribution. By using validation during a model selection procedure, we can reject model instances that overfit to the training data or lead to unsatisfactory performance.

Holdout validation (hereinafter abbreviated as ``holdout'') is one of the simplest validation strategies in \ac{ML}. Part of the development set data is held out as the validation set, and the rest is used for training the models. The holdout validation involves only a single run, and hence part of the data is never used for training and may cause misleading results. Cross-validation alleviates the problem by involving multiple validation runs, and then combine the results of the runs together (to be discussed in details in Sec.~\ref{sec:final-model}). The $k$-fold cross-validation method (hereinafter abbreviated as ``$k$-fold'') partitions the development set data into $k$ equal-sized folds. In a rotated fashion, each time a fold is held out as the validation set and the rest is used for training. 
% In this way, each data point joins the training process for $k-1$ times out of $k$ validation runs.
Under both holdout and $k$-fold strategies, the development set is split \textit{randomly} into a training set and a validation set. Since the split is random, we can expect that the $K^\text{i.d.}$ subgroups of the development set will all be represented in both the training and the validation set. If the cross-validation procedure is properly implemented, we can expect the resulting model will perform well on the \ac{i.d.} data, i.e. these $K^\text{i.d.}$ subgroups in the development set. However, such cross-validation strategy does not take into account the resulting model's generalization behavior on \ac{o.o.d.} test data, and therefore the resulting classifier may not perform well on $\mathcal{D}_\text{test}^\text{o.o.d.}$.

\subsection{\acf{SACV} Strategy for Model Selection}
\begin{figure}
    \centering
    \includegraphics[width=\linewidth]{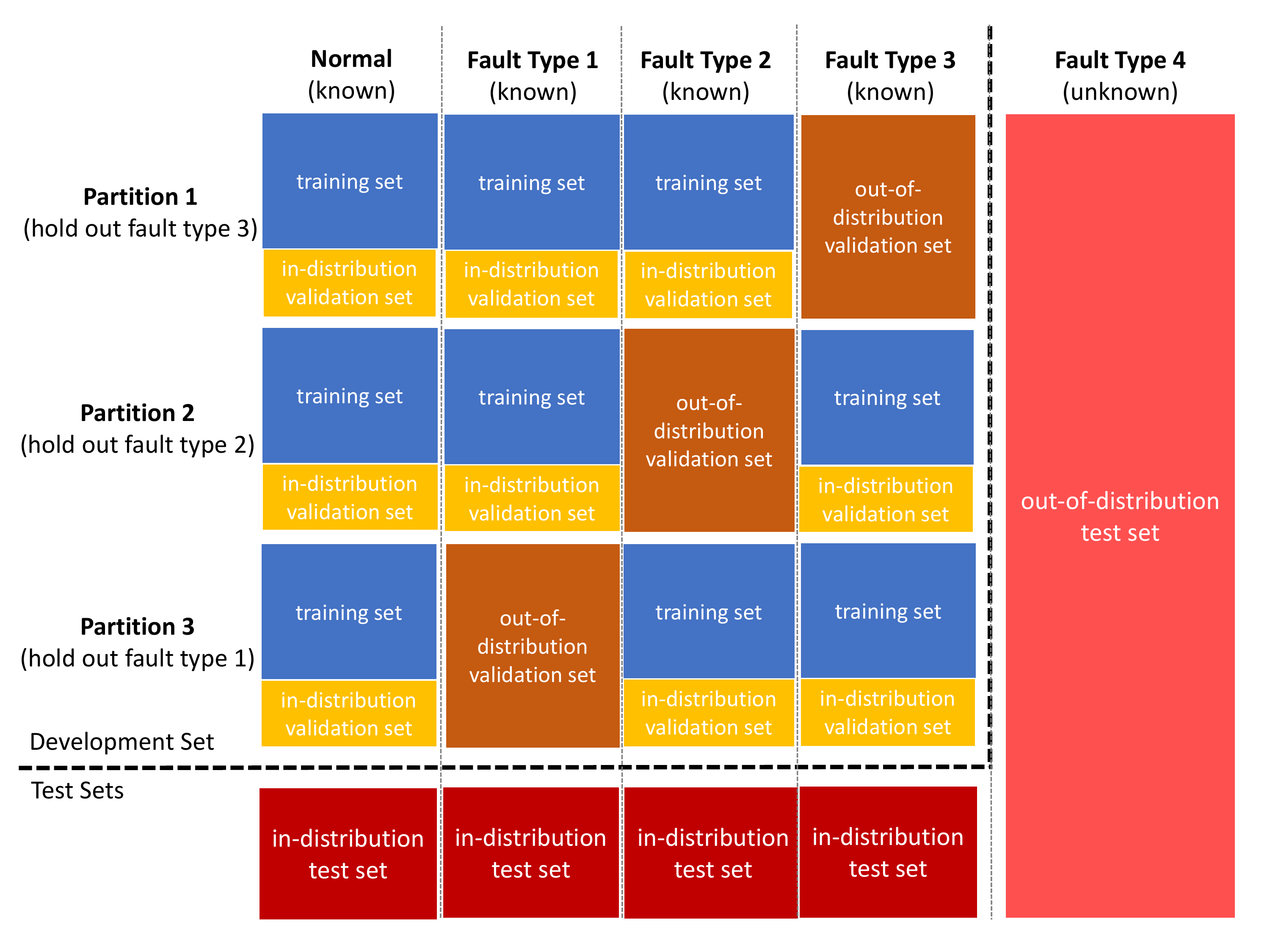}
    \caption{An illustration showing how \ac{SACV} partitions a dataset during cross-validation. In this example, the dataset is made up of four fault types (subgroups), and three out of the four appear in the development set. Our goal is to train a classifier using the development set data to achieve good detection performance on both the unseen \ac{i.d.} (dark red) and the \ac{o.o.d.} (light red) test data.}
    \label{fig:partition}
\end{figure}

To address the issue mentioned above, we propose a \acf{SACV} strategy that explicitly emphasizes and prioritizes the model's generalization performance on test data under domain shift. When an \ac{SACV} strategy is employed, one by one, a subgroup (stratum) of the development set data is selected as the \ac{o.o.d.} validation set; then part of the rest $K^\text{i.d.}-1$ subgroups will be used as the training set, and the remaining portion will be used as the \ac{i.d.} validation set, as illustrated in Fig.~\ref{fig:partition}.

A different technique with similar name is the \textit{stratified $k$-fold cross-validation}, which also deals with stratified data but should not be confused with our proposed \ac{SACV} strategy. In stratified $k$-fold cross-validation, the folds are made by preserving the portion of samples for each class (or stratum). As a result, instead of returning randomly sampled folds, stratified $k$-fold cross-validation returns \textit{stratified folds}. Similar to stratified $k$-fold cross-validation, our proposed strategy also takes data stratification into consideration; however, we deliberately exclude one or more stratum from the training set and keep them solely in the validation set so that we can directly measure a trained model's generalization performance at training time.

The primary objectives of cross-validation are 1) assessing model validity and 2) hyperparameter tuning. During cross-validation, we search through the hyperparameter space and evaluate the performance of each configuration. Suppose a total of $R$ hyperparameter configurations, respectively denoted by $\mathcal{H}_1,\mathcal{H}_2,\ldots,\mathcal{H}_R$, are evaluated and ranked during cross-validation. In our empirical study, we will retain the top-$r$ hyperparameter configurations, instead of the single best-performing one, and report their performance indices.

\subsubsection{Combining Results from Multiple Validation Runs}\label{sec:final-model}
To finalize model selection, the conventional method (hereinafter referred to as \textsc{refit-all}) is to refit the model using the entire development set data and the selected hyperparameter configuration $\mathcal{H}^{\ast}$. Another method is to combine the $K^\text{i.d.}$ models, e.g., by using simple average, that are created during cross-validation in a ensemble. The idea is similar to sample Bagging~\cite{breiman1996bagging}; as a result, we will name this approach \textsc{combine}. Later, we will compare \textsc{refit-all} and \textsc{combine} in our empirical study.

\subsection{Ensemble Learning and Uncertainty Estimation}\label{sec:uncertainty-estimation}
It has long been observed that ensemble learning, a meta-learning method that combines the predictions of multiple diversified learners, can help boost the prediction performance of \ac{ML} models. In addition, recent literature shows that ensemble models can also be used for estimating prediction uncertainties, which is crucial for us to decide whether or not to trust the decisions made by \ac{ML} models.   

\begin{figure}
    \centering
    \includegraphics[width=0.80\linewidth]{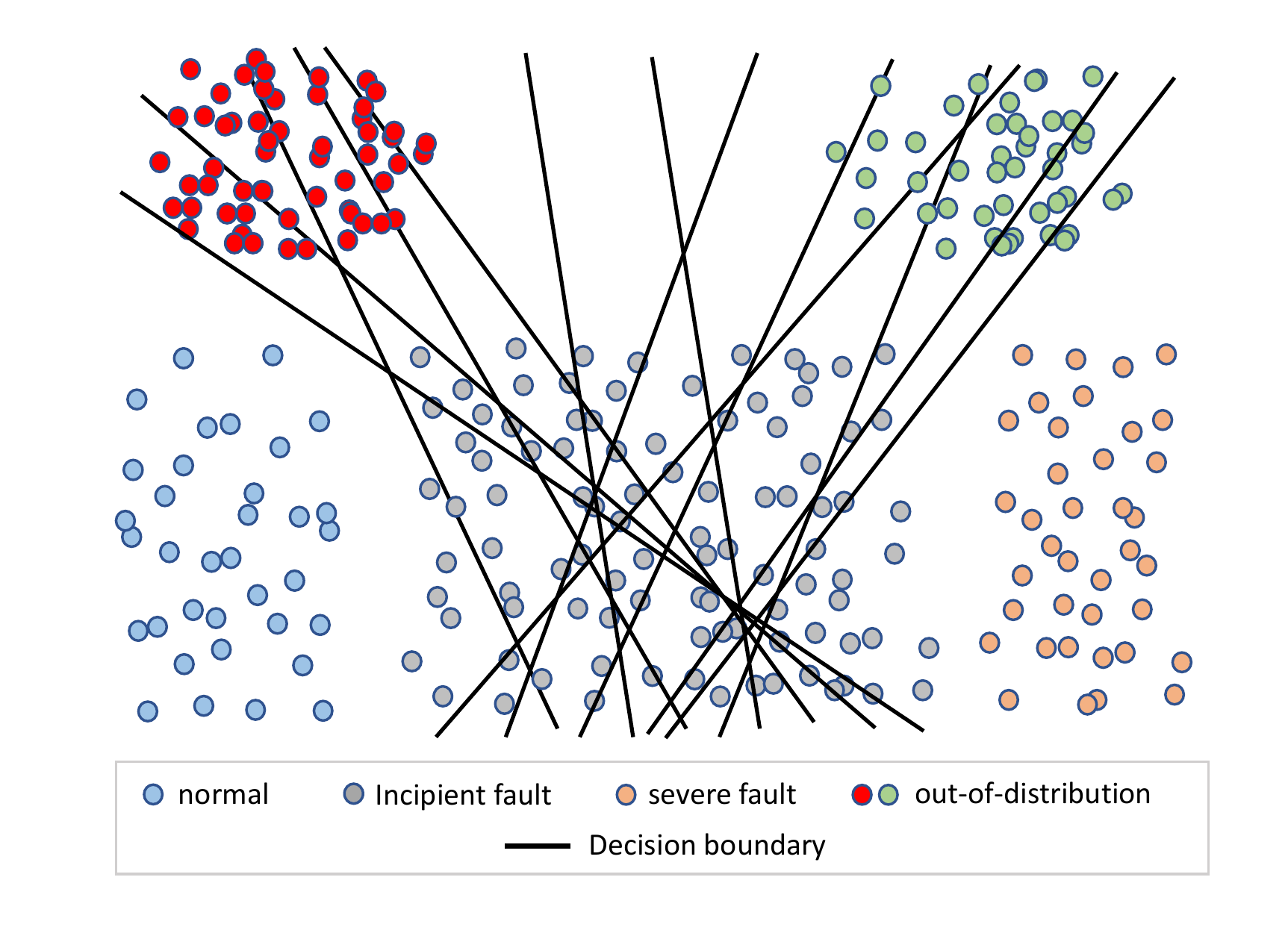}
    \caption{An illustration showing how ensemble classifiers help detect incipient fault data~\cite{Jin2020incipient,Tan2020ExploitingUF} and \ac{o.o.d.} fault data.}
    \label{fig:ensemble-illustration}
\end{figure}

Diversity is recognized as one of the key factors that contribute to the success of ensemble approaches~\cite{brown2005diversity}; the diversity allows individual classifiers to generate different decision boundaries. As illustrated in Fig.~\ref{fig:ensemble-illustration}, the diversity among ensemble members is crucial for improving the detection performance on \ac{o.o.d.} data instances. For the ensemble methods to work, the individual classifiers must exhibit \textit{diversity} among themselves, such that the resulting ensemble can hopefully give a high prediction uncertainty on \ac{o.o.d.} data points.

In our empirical study to be described later, we employed the bagging~\cite{breiman1996bagging} (or bootstrap aggregation) approach for creating diversity among ensemble members. The core idea is to construct a family of models by randomly subsetting the development set (a.k.a. \textit{sample bagging}~\cite{breiman1996bagging}). A later variant called \textit{feature bagging}~\cite{ho1998random} selects a random subset of the features for training each member classifier in an ensemble. One famous application of Bagging in \ac{ML} is the \ac{RF} model. In our empirical study, we only used sample bagging for inducing diversity among ensemble classifiers. In this study, only \textit{homogeneous} base learners, i.e. models of the same type, are used to construct ensembles. The case of heterogeneous ensembles is an interesting setting and we leave it for future investigation. 

Next, we will briefly introduce ensemble mean (hereinafter referred to as \textsc{mean}) and ensemble variance (hereinafter referred to as \textsc{var}), the two commonly used uncertainty metrics to be compared and evaluated in this paper. 

\paragraph{Ensemble Mean (\textsc{mean})}
An intuitive metric that measures the confidence of a classifier on input $x$ is to see how close the prediction $\hat{y}$ is to the decision threshold $\tau$; same with an ensemble classifier that outputs a prediction $\hat{y}^e$ by combining the individual classifier's predictions. Here we use the superscripts in $\hat{y}^{e}$ and $\tau^{e}$ to signify values associated with an ensemble classifier; in the special case where $K=1$, the ensemble classifier degenerates to a single learner model.
The smaller the gap $\left\vert\hat{y}_i^{e}-\tau^{e}\right\vert$ is, the higher the uncertainty with $x_i$. Since we prefer the convention that larger function values of $u^\textsc{mean}(x_i)$ corresponds to larger uncertainties, we define the uncertainty score under the margin metric can be formulated as 
\begin{align}
    u^\textsc{mean}(x_i) \doteq 1 - \left\vert\hat{y}_i^{e}-\tau^{e}\right\vert,
\end{align}\label{eqn:mean-metric}
where a constant $1$ is added to the definition so that the uncertainty value $u^\textsc{mean}(x)$ is always positive. Since the ensemble prediction $\hat{y}_i^{e}$ is obtained by taking the average of the individual outputs of classifiers in the ensemble, we will hereinafter refer to this metric as \textsc{mean}.

\paragraph{Ensemble Variance (\textsc{var})}
The variance (or standard deviation) metric~\cite{leibig2017leveraging,jin2019detecting} measures how spread out the individual learners' predictions are from the ensemble prediction $\hat{y}_i^{e}$. The uncertainty score of input $x_i$ based on \textit{sample variance} can be written as
\begin{align}
    u^\textsc{var}(x_i)\doteq\frac{1}{K-1}\sum_{k=1}^K \left[\hat{y}_i^{(k)} - \hat{y}_i^{e}\right]
\end{align}\label{eqn:var-metric}

A problem with \textsc{var} is that it focuses mainly on the disagreement among ensemble predictions but do not take in consideration the magnitude of $\hat{y}_i^{e}$. Consider a scenario where the all ensemble members predict a probability of $0.5$. Both \textsc{var} and \textsc{kl} will produce an uncertainty score of $0$ and thus will not be able to capture any decision uncertainties; in fact, this case where all learners give an output of $0.5$ is highly uncertain. 

A theoretical analysis for comparing between the two uncertainty metrics \textsc{mean} and \textsc{var} is given in our previous works~\cite{Tan2020ExploitingUF,Jin2020incipient}, but on uncertain examples known as \textit{incipient anomalies} that exhibit mild symptoms of known anomaly (faults or diseases) types. The results showed that \textsc{mean} is a more \textit{robust} uncertainty metric than \textsc{var} in the sense that the performance lower bound given by \textsc{mean} is higher than that of \textsc{var}. It is still unclear which uncertainty metric is likely to perform better on \ac{o.o.d.} strata; we plan to give an answer to this question in our empirical study to be presented later. 

\begin{figure}
    \centering
    \includegraphics[width=0.85\linewidth]{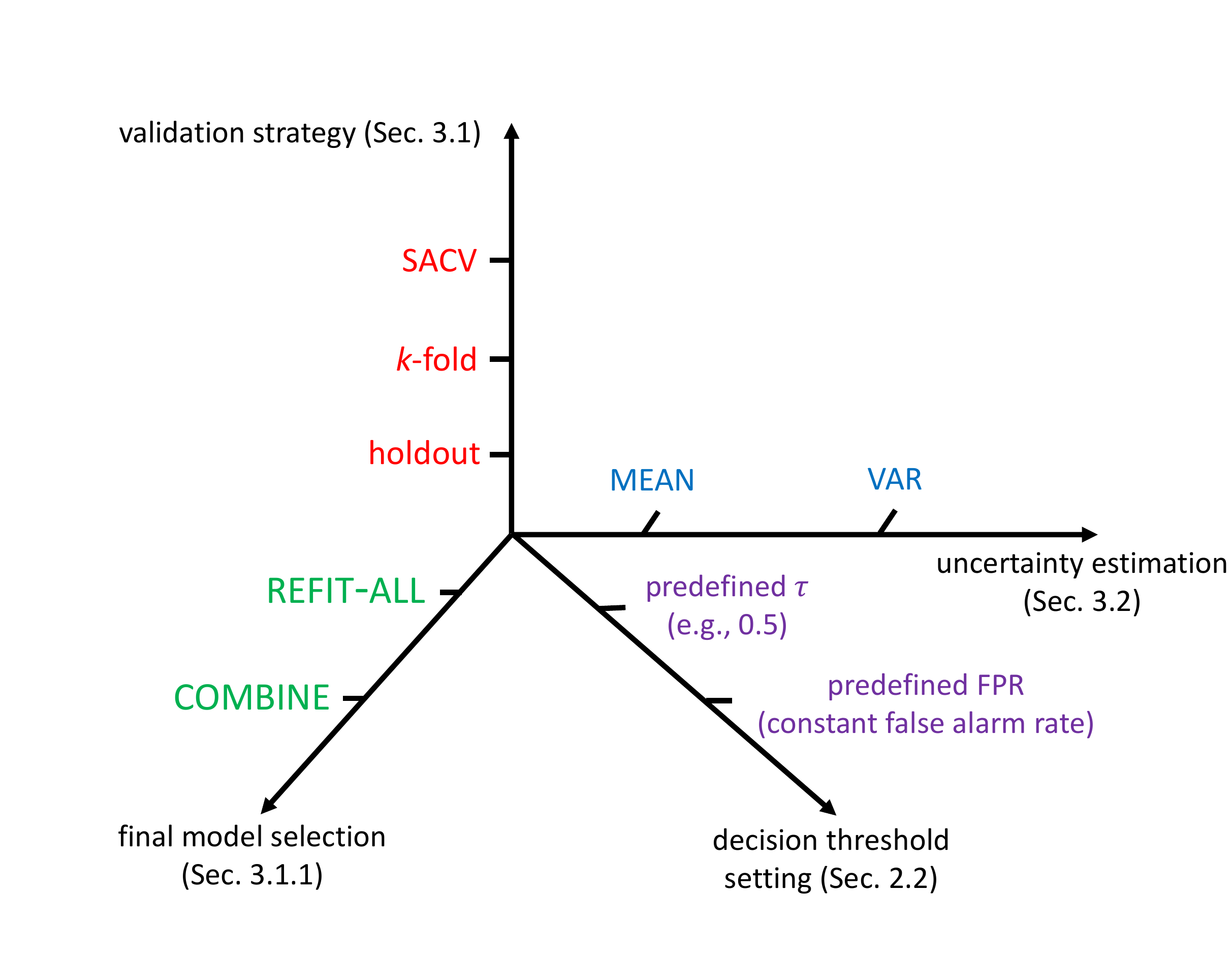}
    \caption{An illustration showing the concepts and techniques compared in this study. Orthogonal concepts are put onto different axes.}
    \label{fig:concept-coord}
\end{figure}

We show in Fig.~\ref{fig:concept-coord} the relationship among the various concepts introduced above. Note that techniques on different axes are orthogonal, and thus can be applied together. 

\section{Datasets}\label{sec:dataset}

In this section, we give a brief overview of the three datasets to be used in our empirical study; further details about the three datasets will be given in the appendix. We will also describe how we partitioned the datasets in our experiments into development sets and test sets. 

\paragraph{ASHRAE RP-1043 Chiller Faults Dataset (``chiller dataset'')}\label{sec:rp-1043}
We used the ASHRAE~RP-1043 Dataset~\cite{comstock1999development} to examine the proposed approach. In the chiller dataset, sensor measurements of a 90-ton centrifugal water-cooled chiller were recorded under both fault-free and various fault conditions. In this study, we included the six faults (FT-FWE, FT-FWC, FT-RO, FT-RL, FT-CF, FT-NC) used in our previous study~\cite{jin2019detecting} as the fault (positive) class. 
% We consider SL3 and SL4 cases as severe faults, and SL1 and SL2 cases as incipient faults. For feature selection, we also followed our previous study~\cite{jin2019detecting} and used the sixteen key features therein for training our models. To give the readers an intuitive view about the distribution of the chiller data, we employed the \ac{LDA} algorithm to reduce the data into two dimensions, and visualized part of the reduced-dimension data in Fig.~\ref{fig:visualization-chiller} described earlier. We can observe a general trend in the visualization: data points will deviate further away from the normal cluster when the corresponding fault develops into a higher \ac{SL}.

\paragraph{ASHRAE RP-1312 AHU Faults Dataset (``AHU dataset'')}\label{sec:rp-1312}
Another important component of a building \ac{HVAC} system is the \ac{AHU} whose functionality is to regulate and circulate air to the indoor zones in a building. Our study included 25 commonly encountered \ac{AHU} faults (eleven in spring, eight occur in summer, and six in winter). By treating data from each season as an independent dataset, we will then have three sub-datasets, namely \textit{AHU-spring}, \textit{AHU-summer} and \textit{AHU-winter}, for our experimental study. We adopted the features selected by Li~et~al.'s previous work~\cite{li2017optimal}.

\paragraph{Power System Faults Dataset (``power dataset'')}\label{sec:power}
To further validate the proposed approach, we also examined its performance on a fault dataset from another typical \ac{CPS}---an electric power system. This dataset contains rich dynamic characteristics and it models the high-order complexity of the power system under \acp{HIF}. The benchmark system that generates the \acp{HIF} has a variety of system configurations under different distributed energy resource technologies such as synchronous machines and inverter-interfaced renewable generators. Based on this benchmark system, we have created three sub-datasets for 1) faults occurred in three locations 2) faults resulting from six impedance values, and 3) faults of four different types (single-line-to-ground faults, line-to-line faults, line-to-line-to-ground faults, and three-phase faults); details of the dataset can be found in~\cite{cui2019feature}. We will refer to the three sub-datasets respectively as \textit{power-loc}, \textit{power-res}, and \textit{power-ft}. 

\subsection{Dataset Partitioning}\label{sec:dataset-partitioning}
To study the generalization performance of different cross-validation methods, we performed a series of experiments on each dataset. For each dataset consisting of $K$ subgroups, we repeated the experiment for $K$ times, each time leaving out a different subgroup as the \ac{o.o.d.} test set. The \ac{i.d.} test set is then partitioned out of the rest $K^\text{i.d.}=K-1$ subsets. The remaining data will constitute the development set.

\section{Experimental Details}\label{sec:experiment}
\subsection{Experiment Setup}
We conducted the experiments on all three datasets described above in Sec.~\ref{sec:dataset}. \ac{DT} and \ac{NN} models were used as base learners in our experimental study, and then combined them together into Bagging ensembles~\cite{breiman1996bagging}. We built Bagging ensembles of two different sizes $5$ and $10$, and used the single learner case as the baseline. For each experiment, we excluded one subgroup from the whole dataset and use it as the \ac{o.o.d.} test data, as described earlier in Sec.~\ref{sec:dataset-partitioning}. To induce diversity, we swept a wide range of hyperparameters settings, and selected the top 5 best-performing sets of hyperparameters with which the models obtained the highest test scores. More details of our experimental setup and implementation can be found in the released code.

%   Before testing, Gridsearch cross validation method was used to validate the models and the final models will be selected depending on the cross validation scores. Selected models will be tested on the \ac{o.o.d.} test set and evaluated by the \ac{FNR}. % In our experiment, we used two different ``final model selection'' methods: averaging cross validated models and refitting the model with all development set. Detailed experiment settings can be found in the supplementary material. \yingshui{should we mention the dataset here?}

\subsection{Comparing Final Model Selection Methods: \textsc{refit-all} vs. \textsc{combine}}
\begin{figure}[tb]
  \centering
  \begin{subfigure}[t]{0.23\linewidth}
    \centering
    \includegraphics[height=2.6cm]{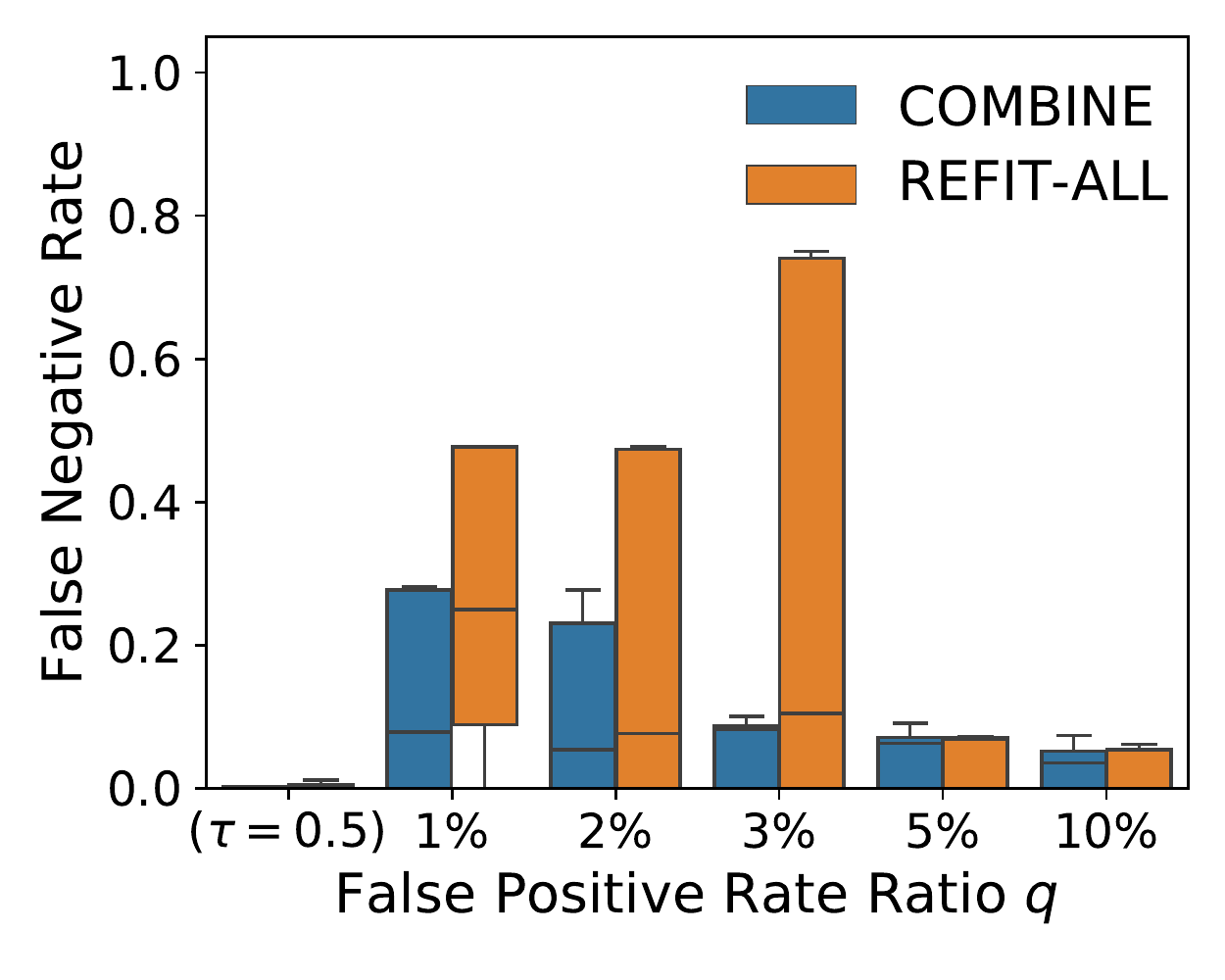}
    \caption{RF: chiller (FT-RL)}
  \end{subfigure}
  \begin{subfigure}[t]{0.23\linewidth}
    \centering
    \includegraphics[height=2.6cm]{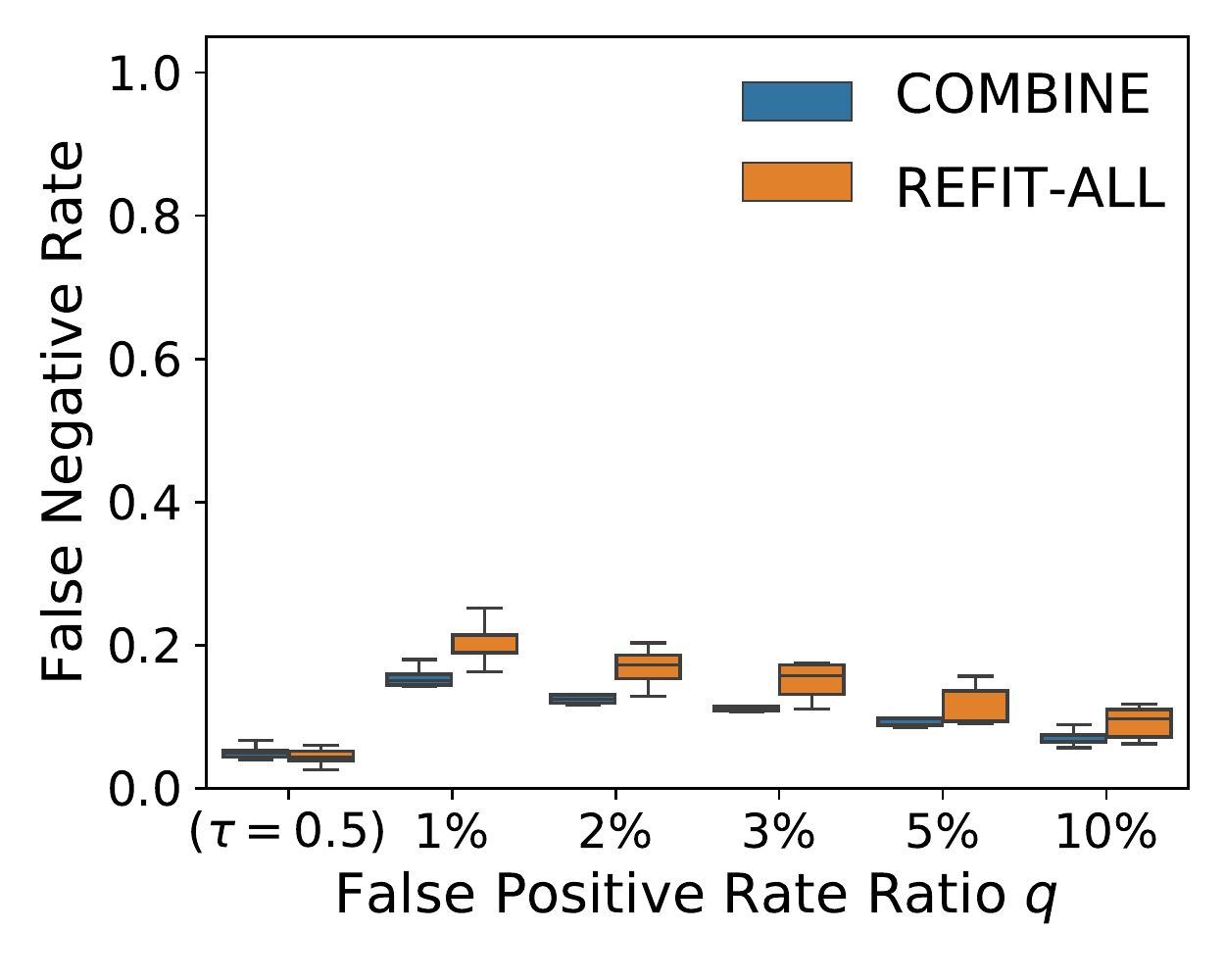}
    \caption{NN: chiller (FT-RL)}
  \end{subfigure}
  \begin{subfigure}[t]{0.23\linewidth}
    \centering
    \includegraphics[height=2.6cm]{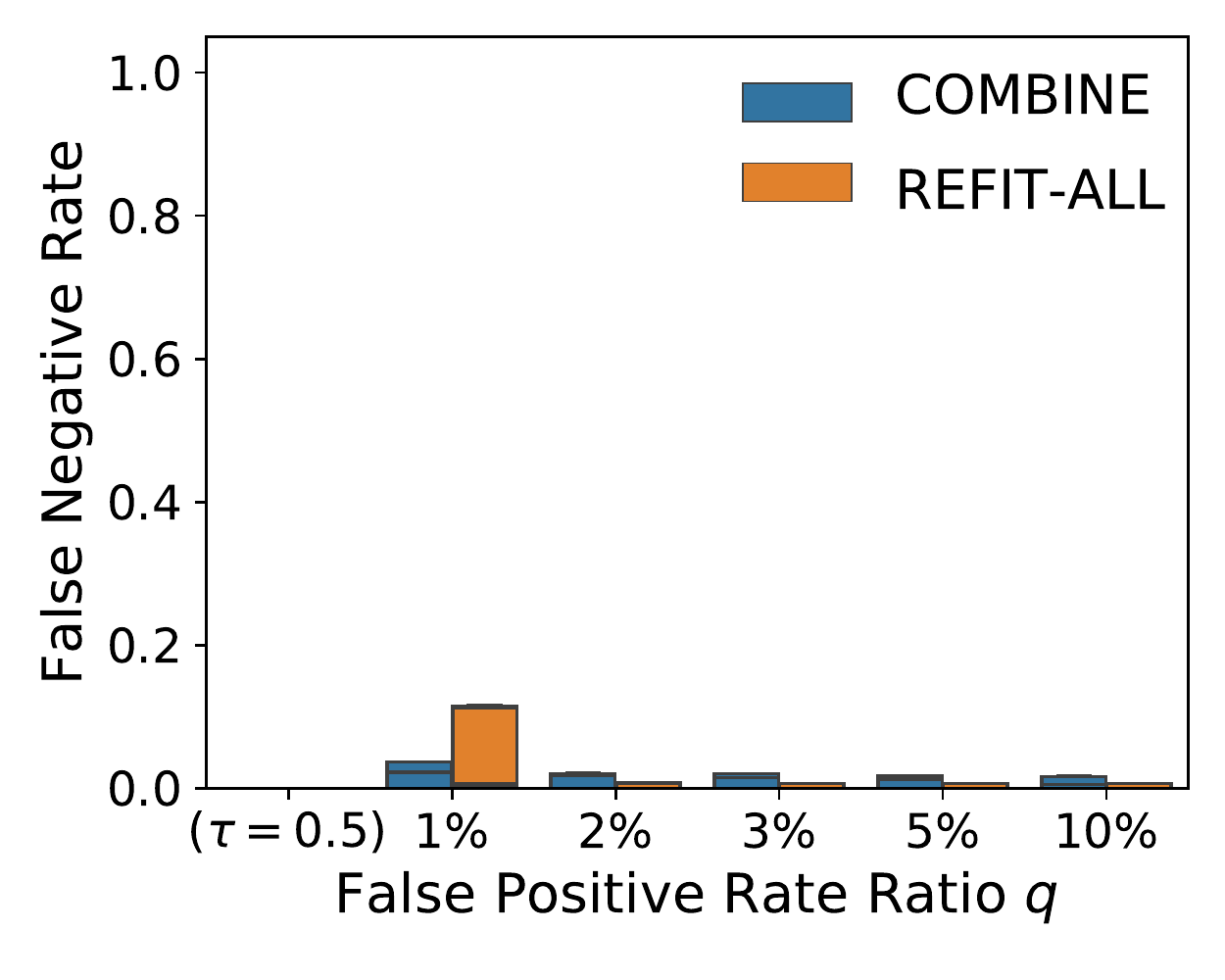}
    \caption{RF: chiller (FT-CF)}
  \end{subfigure}
  \begin{subfigure}[t]{0.23\linewidth}
    \centering
    \includegraphics[height=2.6cm]{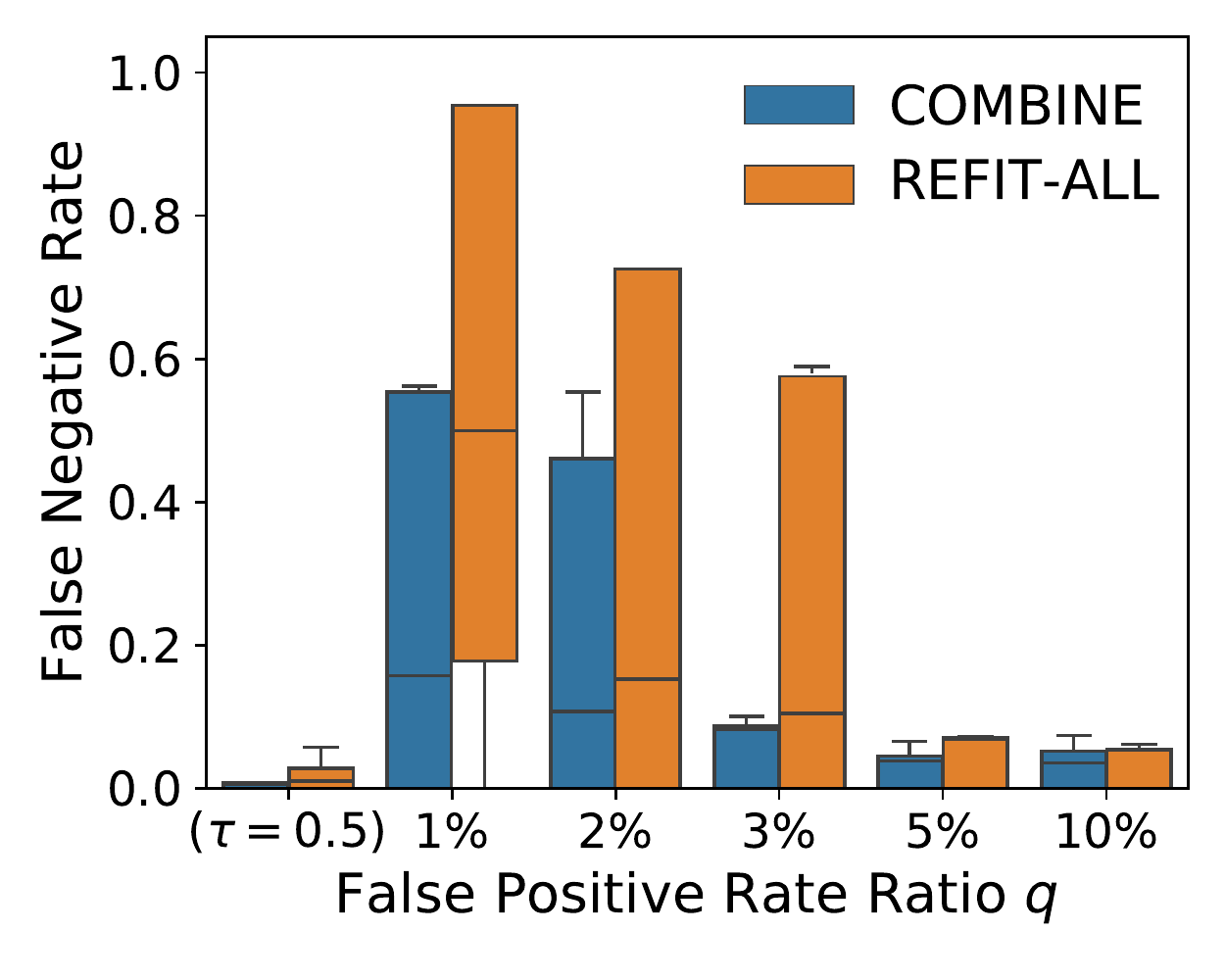}
    \caption{NN: chiller (FT-CF)}
  \end{subfigure}  

  \begin{subfigure}[t]{0.23\linewidth}
    \centering
    \includegraphics[height=2.6cm]{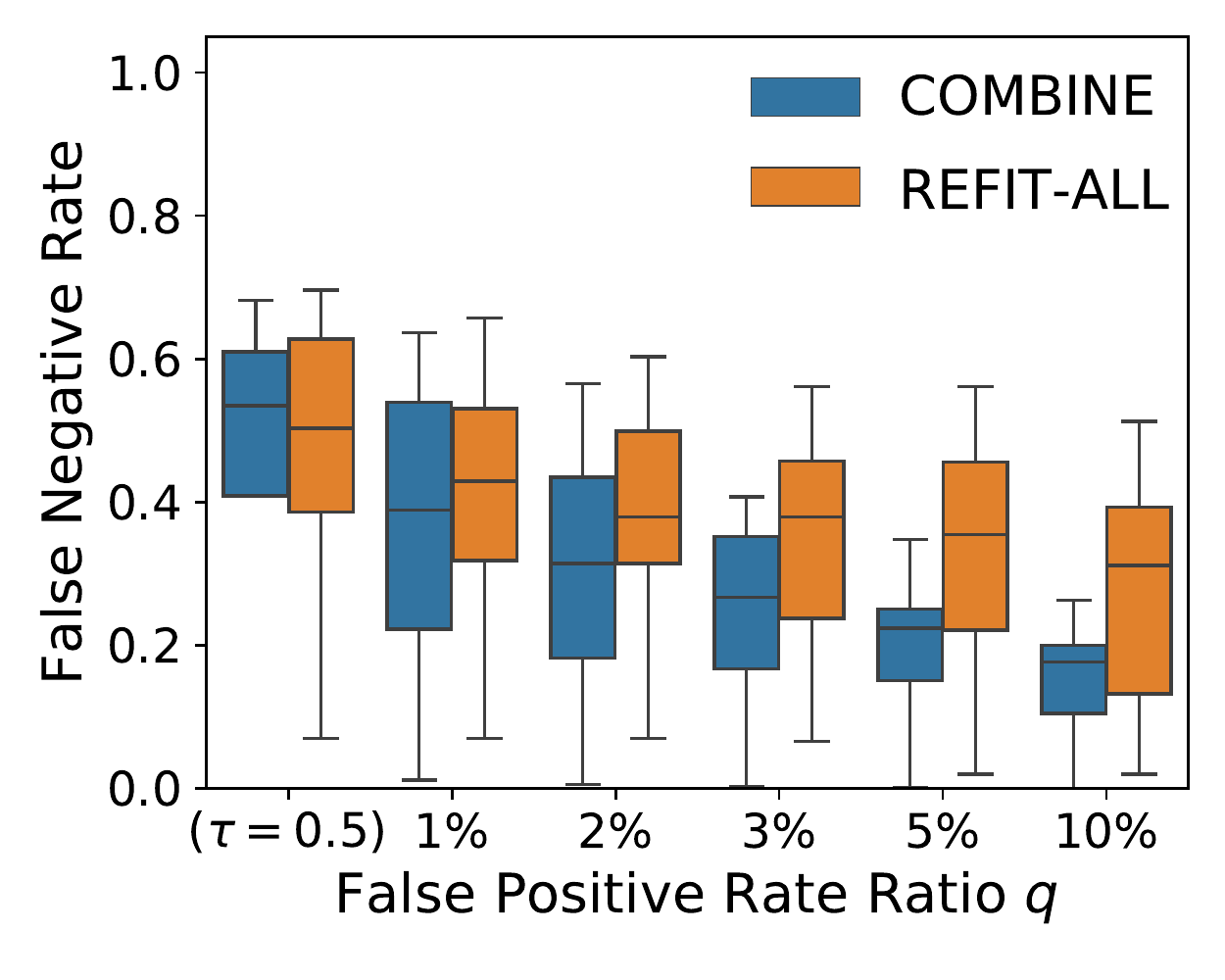}
    \caption{RF: AHU (SP-FT-8)}
  \end{subfigure}
  \begin{subfigure}[t]{0.23\linewidth}
    \centering
    \includegraphics[height=2.6cm]{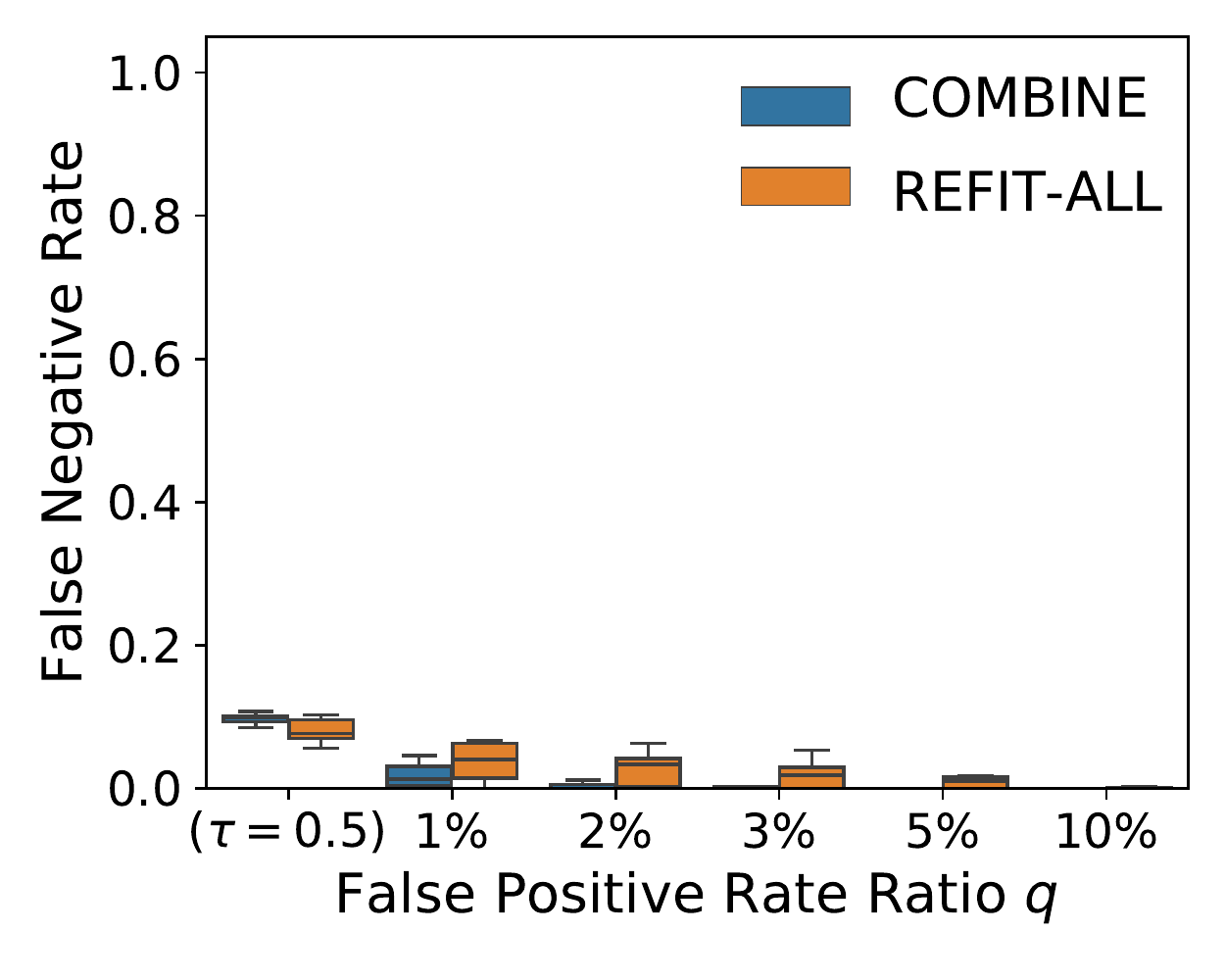}
    \caption{NN: AHU (SP-FT-8)}
  \end{subfigure}
  \begin{subfigure}[t]{0.23\linewidth}
    \centering
    \includegraphics[height=2.6cm]{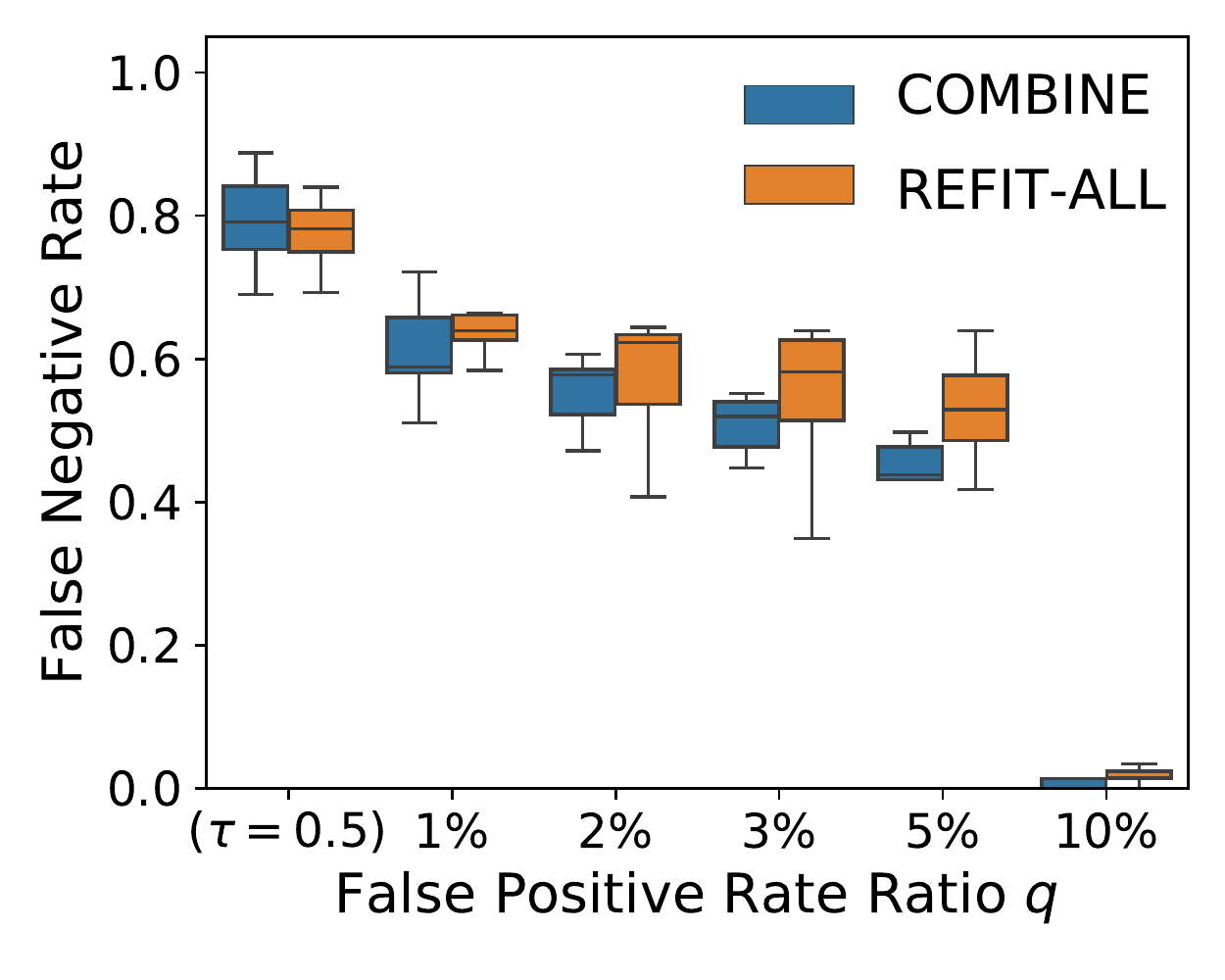}
    \caption{RF: power (FT-4)}
  \end{subfigure}
  \begin{subfigure}[t]{0.23\linewidth}
    \centering
    \includegraphics[height=2.6cm]{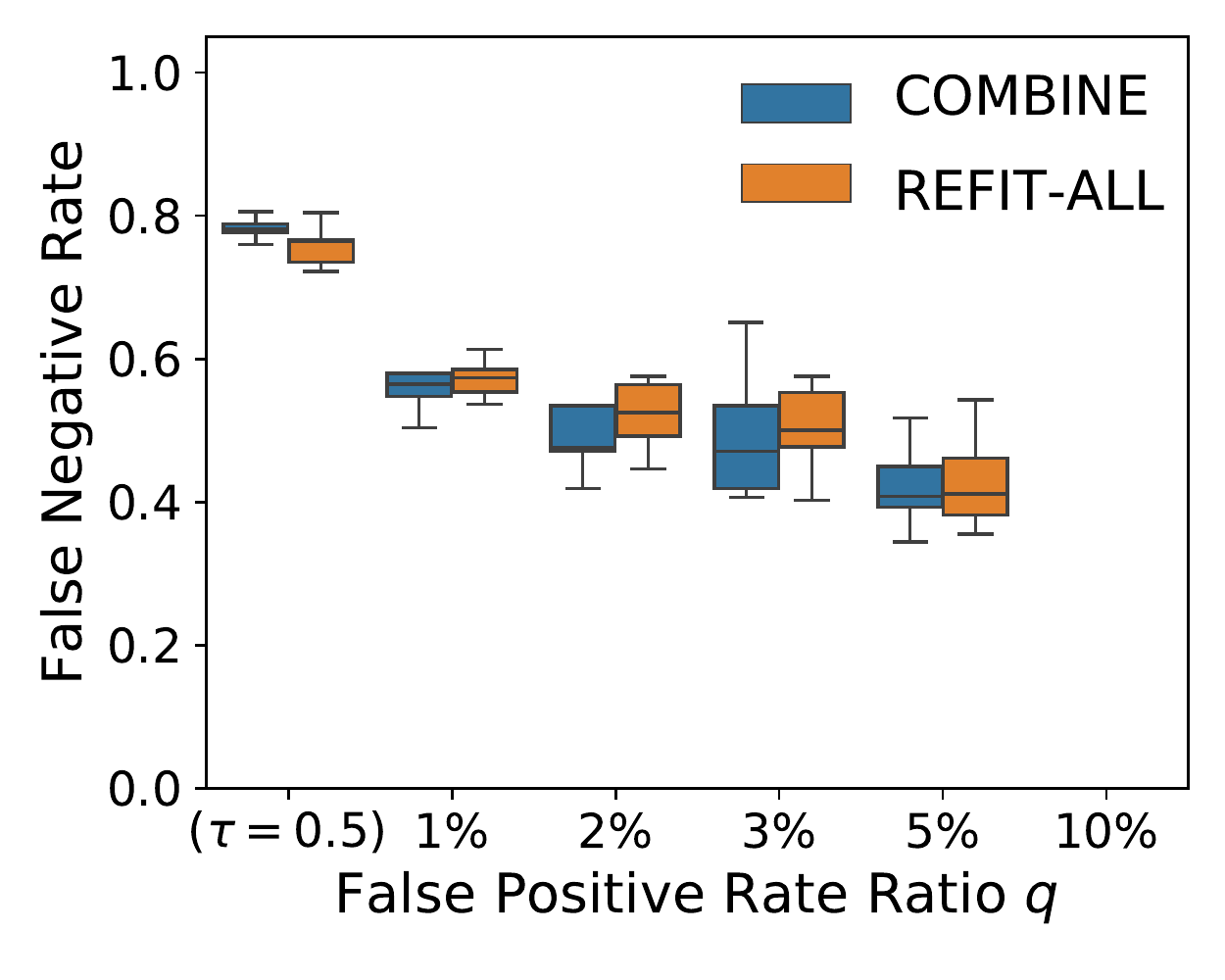}
    \caption{NN: power (FT-4)}
  \end{subfigure}
  
  \begin{subfigure}[t]{0.23\linewidth}
    \centering
    \includegraphics[height=2.6cm]{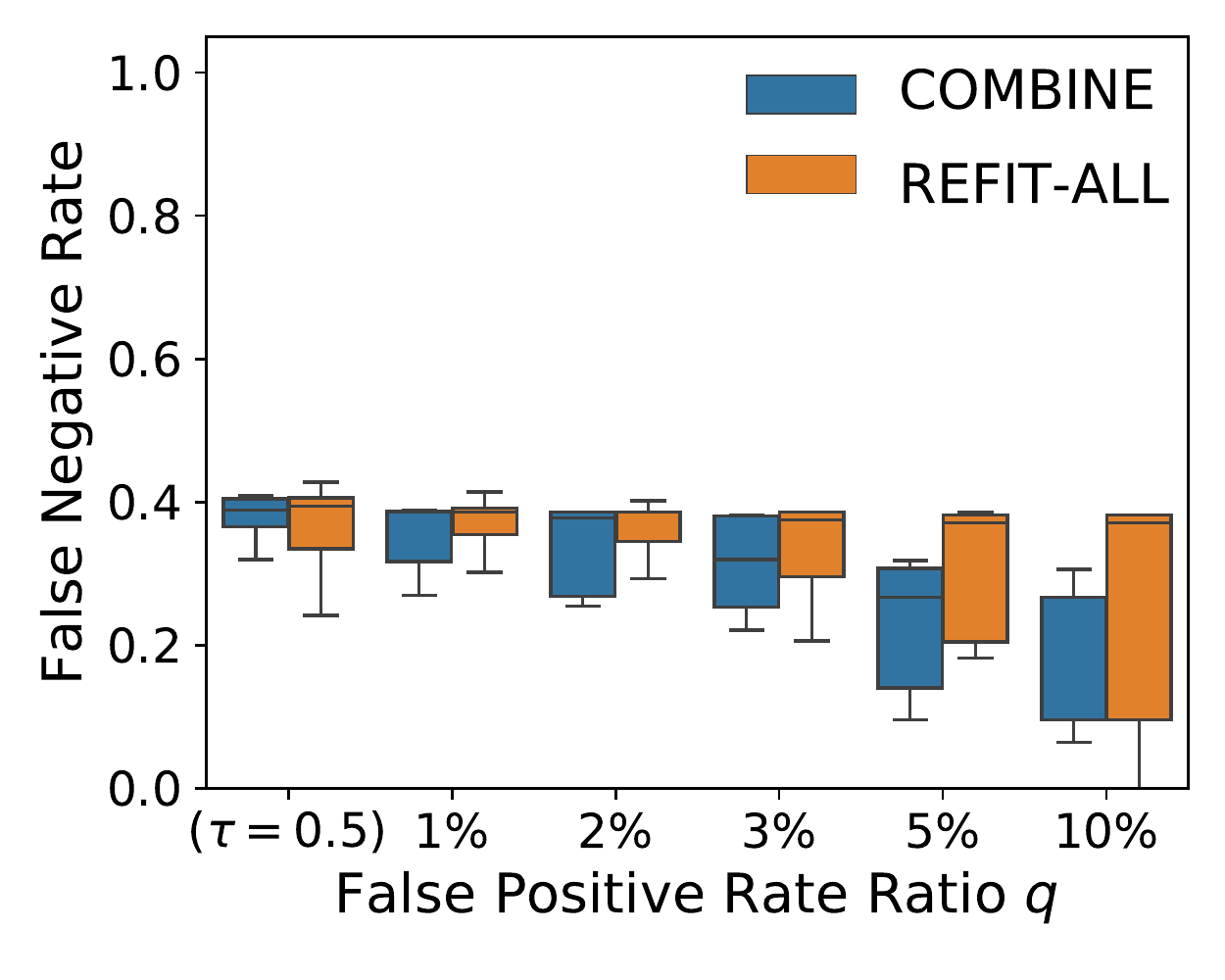}
    \caption{RF: AHU (SU-FT-4)}
  \end{subfigure}
  \begin{subfigure}[t]{0.23\linewidth}
    \centering
    \includegraphics[height=2.6cm]{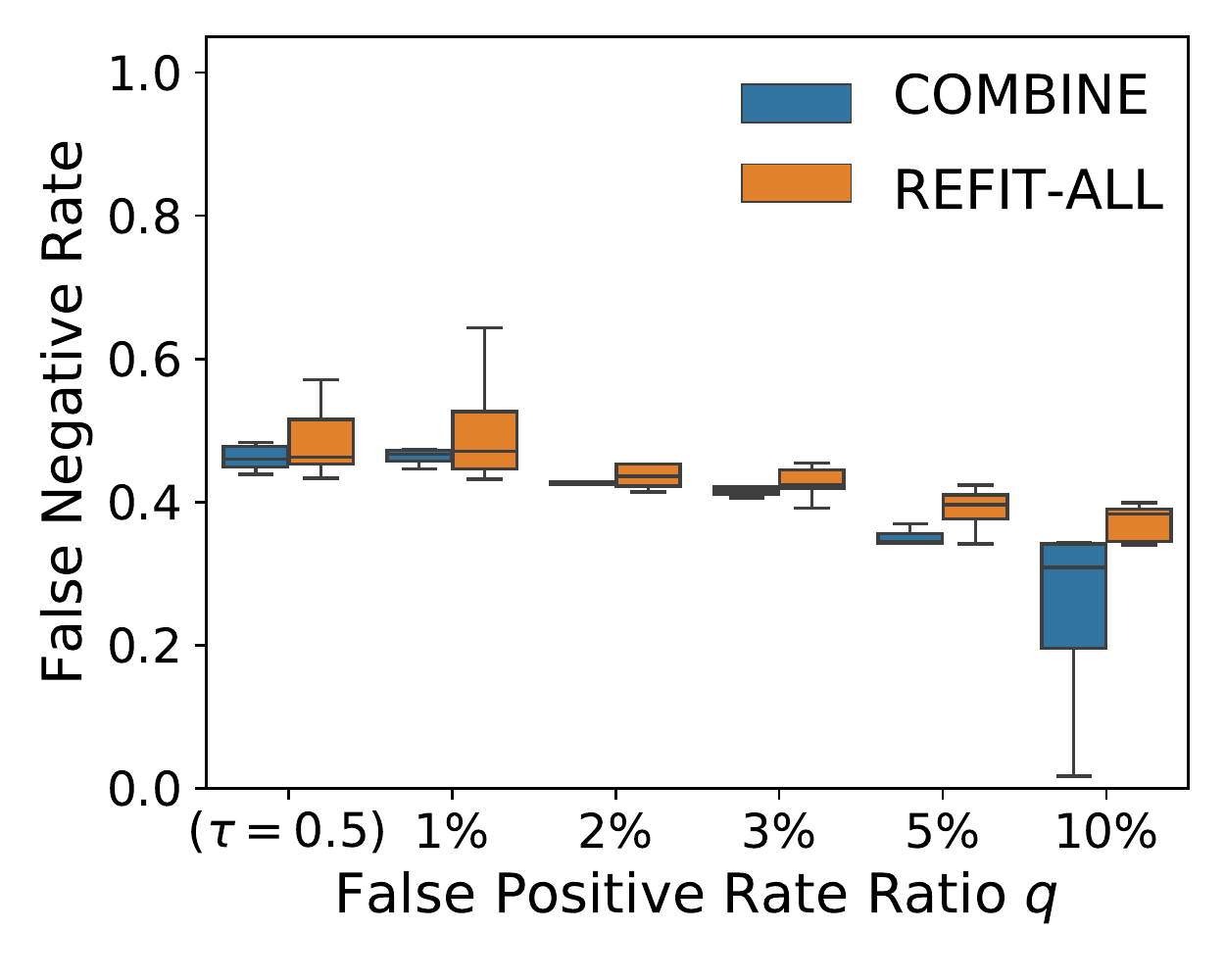}
    \caption{NN: AHU (SU-FT-4)}
  \end{subfigure}
  \begin{subfigure}[t]{0.23\linewidth}
    \centering
    \includegraphics[height=2.6cm]{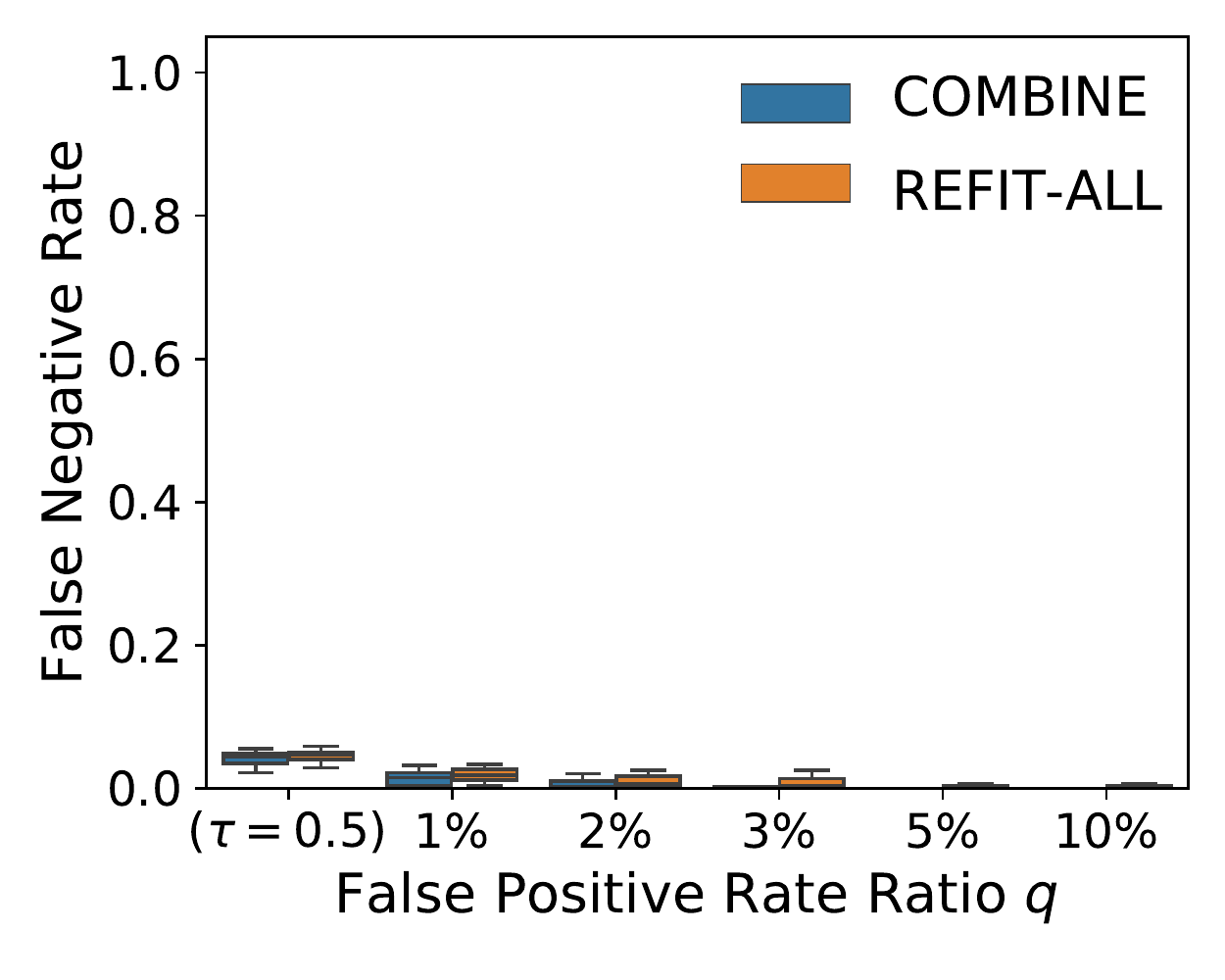}
    \caption{RF: power (LOC-2)}
  \end{subfigure}
  \begin{subfigure}[t]{0.23\linewidth}
    \centering
    \includegraphics[height=2.6cm]{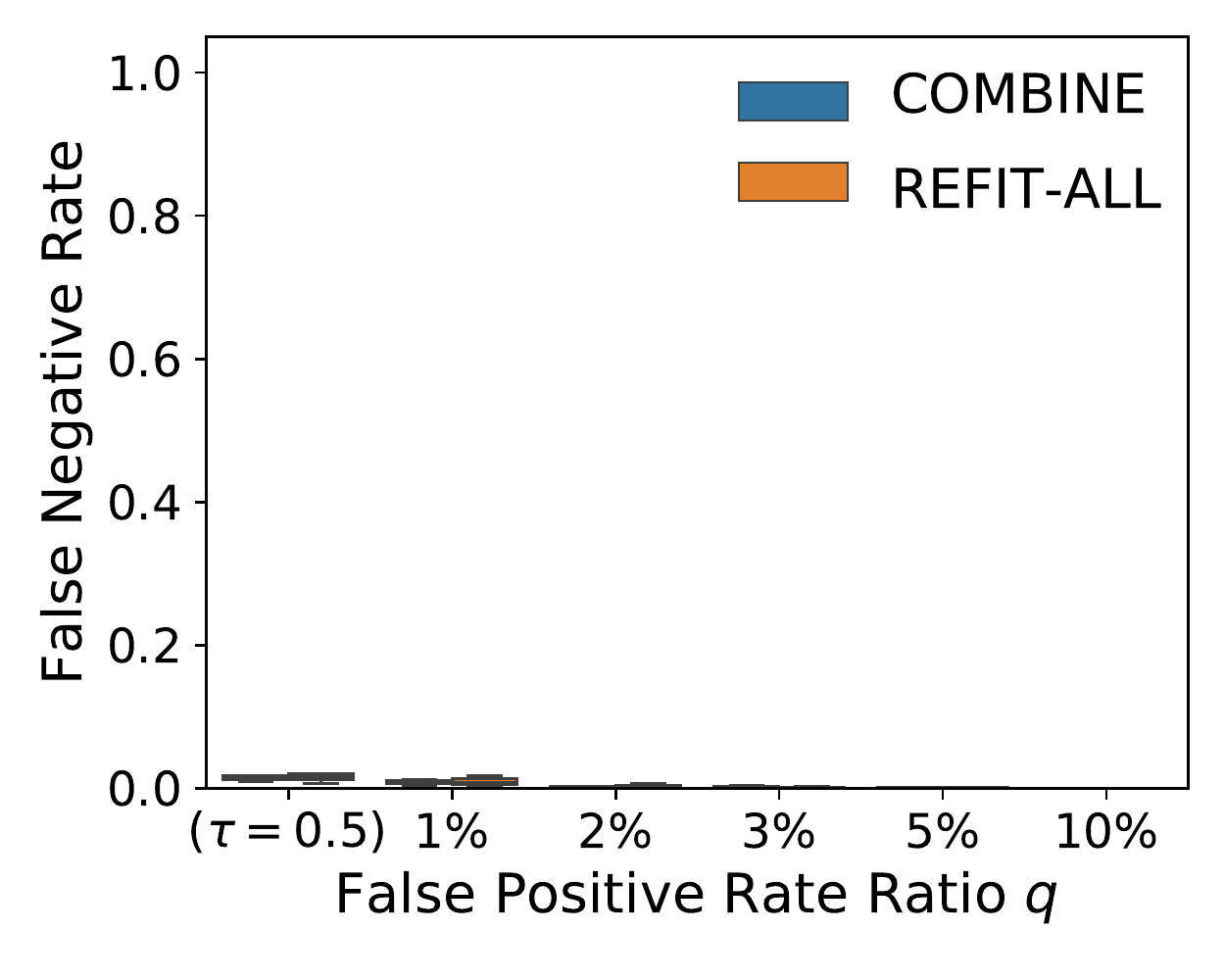}
    \caption{NN: power (LOC-2)}
  \end{subfigure}
  
  \begin{subfigure}[t]{0.23\linewidth}
    \centering
    \includegraphics[height=2.6cm]{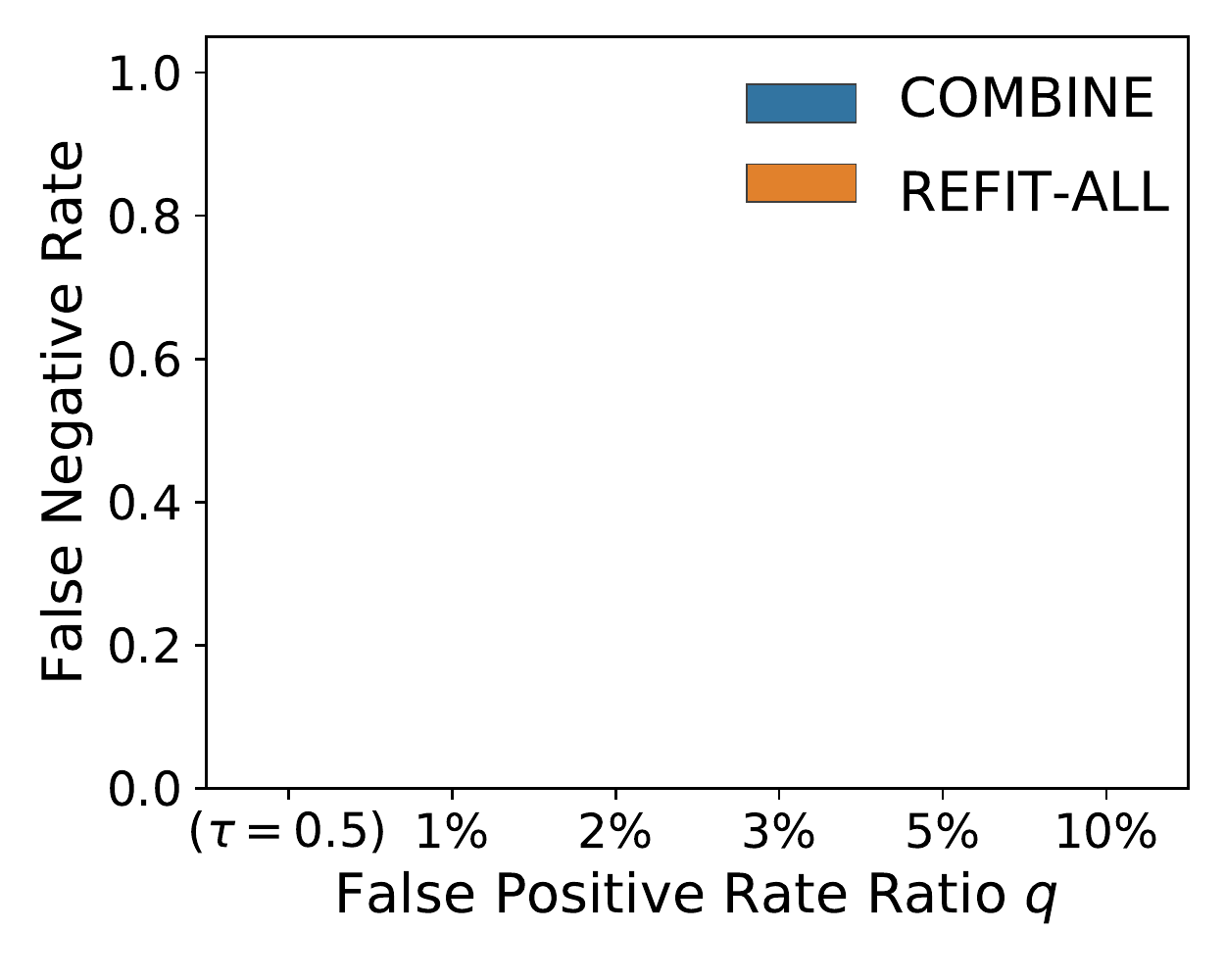}
    \caption{RF: AHU (WT-FT-2)}
  \end{subfigure}
  \begin{subfigure}[t]{0.23\linewidth}
    \centering
    \includegraphics[height=2.6cm]{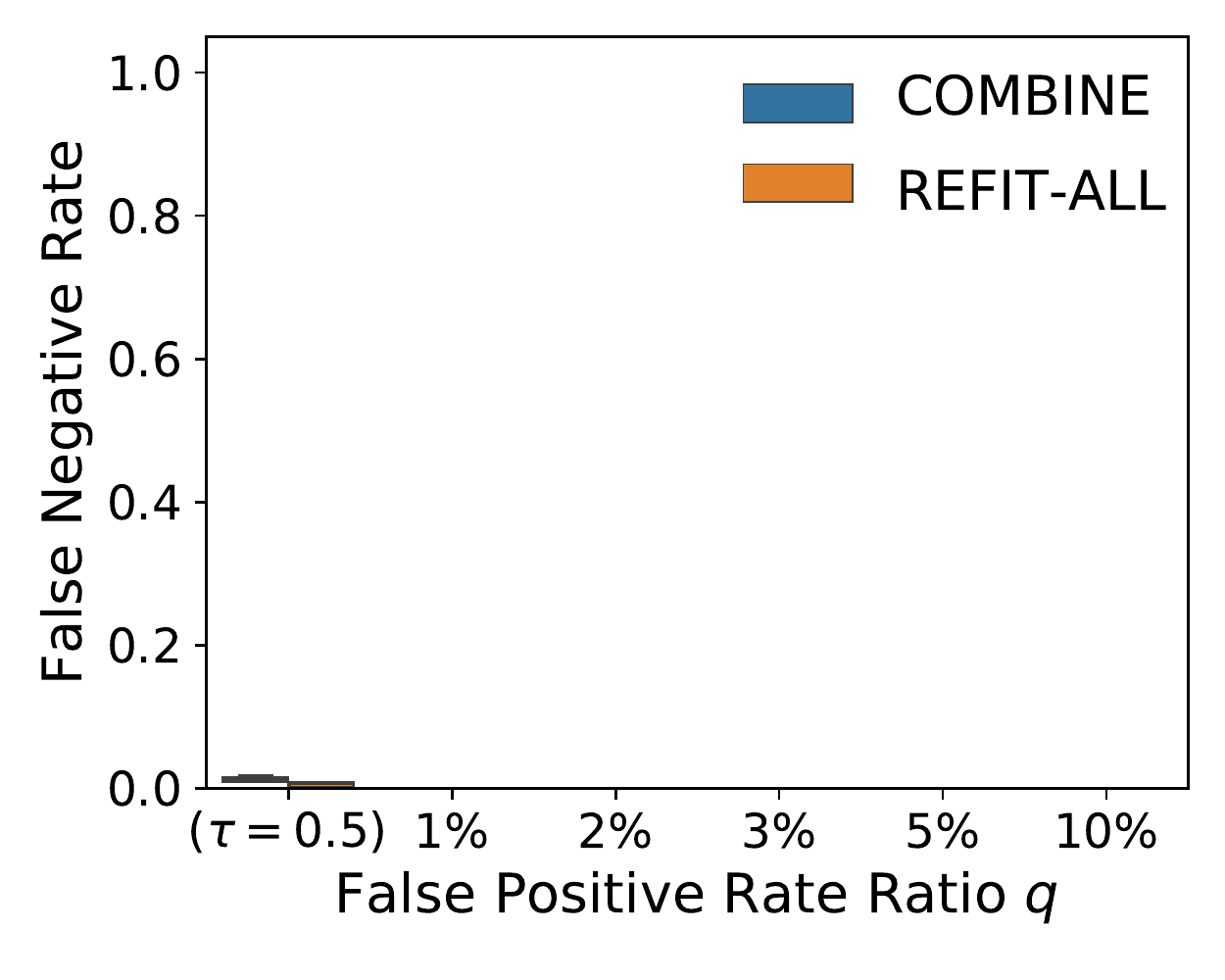}
    \caption{NN: AHU (WT-FT-2)}
  \end{subfigure}
  \begin{subfigure}[t]{0.23\linewidth}
    \centering
    \includegraphics[height=2.6cm]{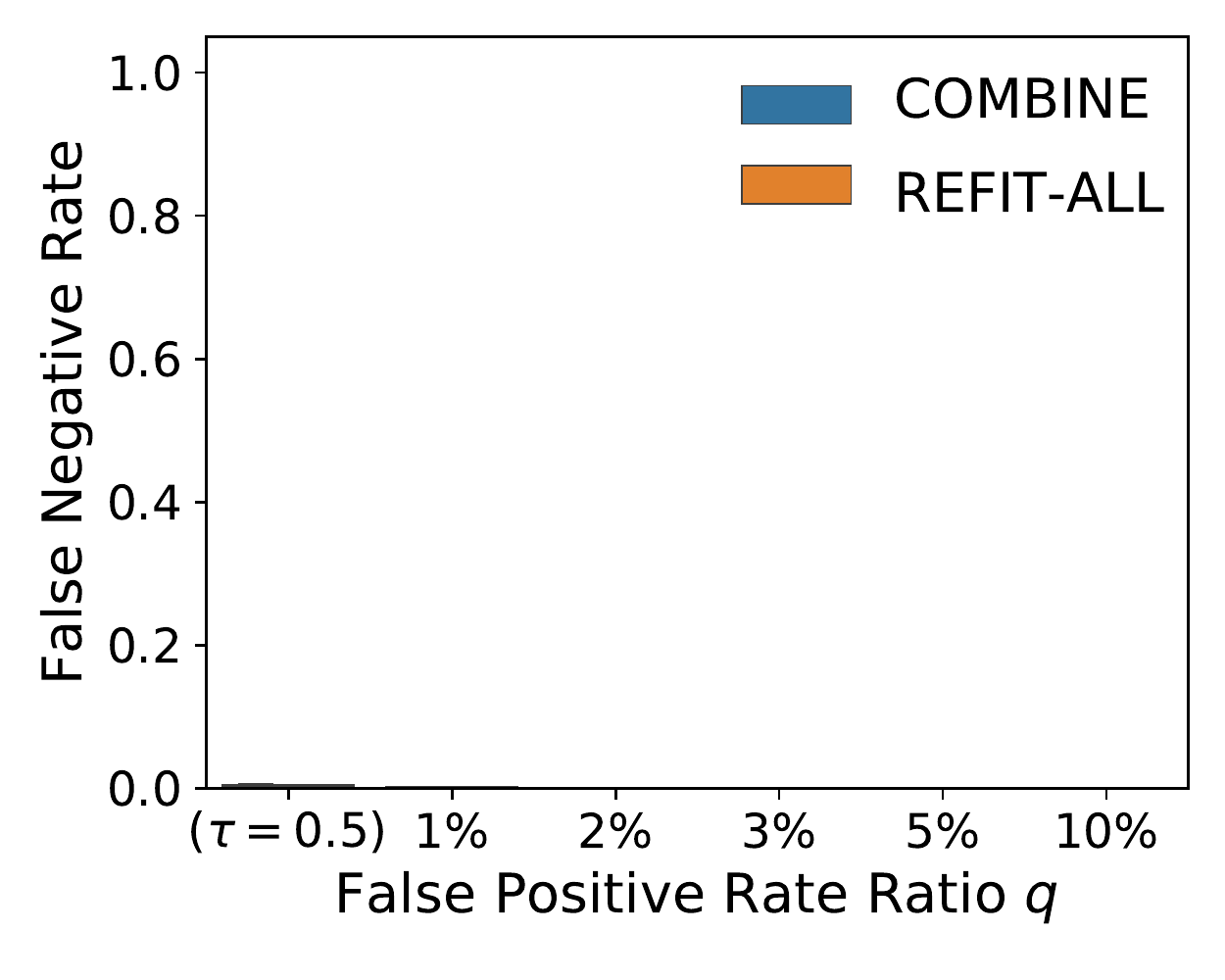}
    \caption{RF: power (RES-2)}
  \end{subfigure}
  \begin{subfigure}[t]{0.23\linewidth}
    \centering
    \includegraphics[height=2.6cm]{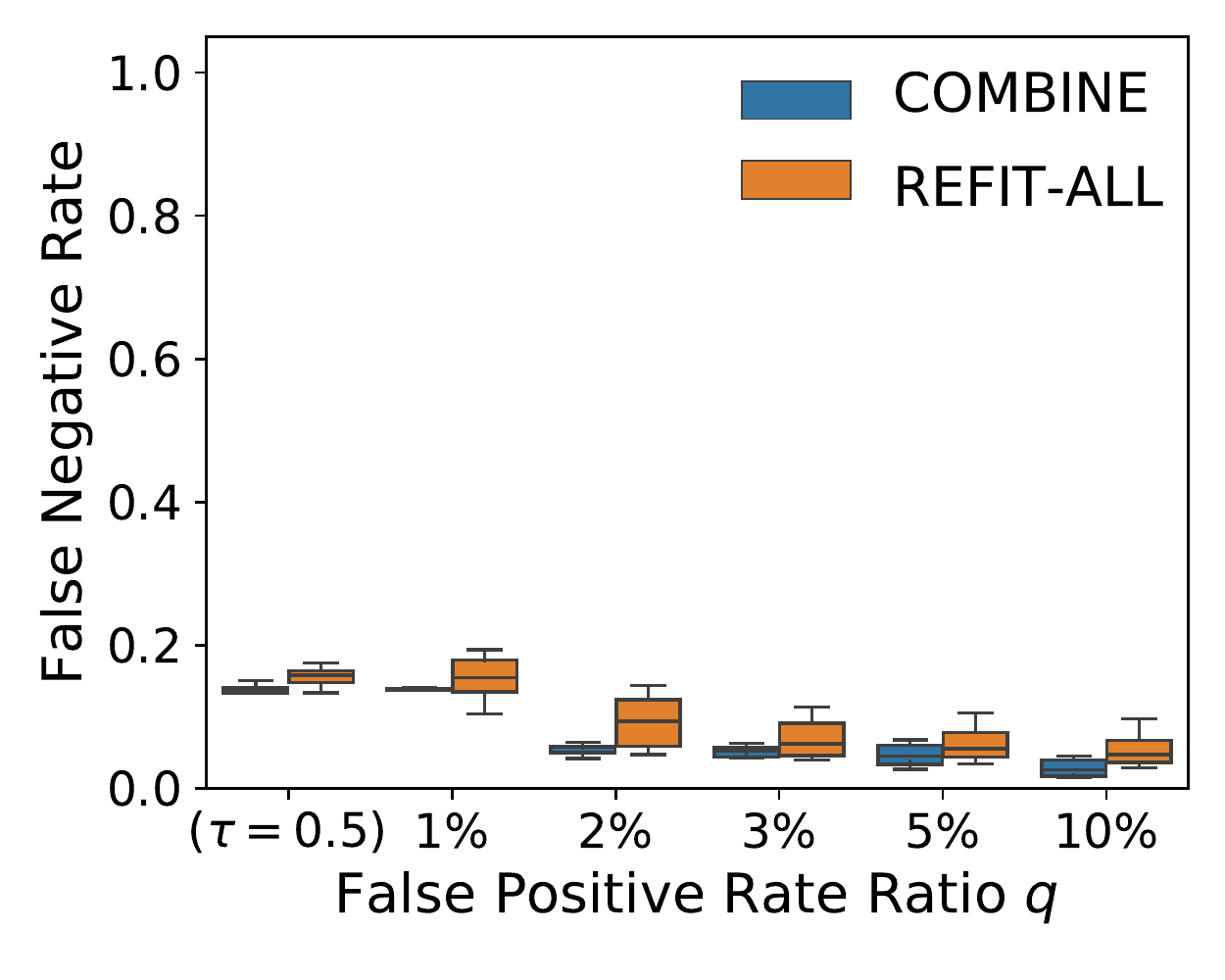}
    \caption{NN: power (RES-2)}
  \end{subfigure}  
  \caption{Performance comparison between the \textsc{refit-all} and the \textsc{combine} methods in terms of their \ac{FNR} on different datasets are presented: 1) the chiller dataset, 2) the \acs{AHU} dataset, and 3) the power dataset. The excluded subgroup that is used as the \ac{o.o.d.} test set and \ac{SACV} is used as the cross-validation method.}
  \label{fig:all-vs-bag}
\end{figure}

We first compare the two ``final model selection'' methods, \textsc{refit-all} and \textsc{combine}, described in Sec.~\ref{sec:final-model}, by examining their performance differences on the three datasets (including all of their sub-datasets). Both give similar performance on \ac{i.d.} data, and we further assess their performance in terms of the \ac{FNR} on the \ac{o.o.d.} data under 1) different configurations of $q$ (i.e. the predefined \ac{FPR} level on the development set): $1\%$, $2\%$, $3\%$, $5\%$, $10\%$ and also 2) under $\tau=0.5$. For the three AHU sub-datasets, we only noticed significant performance differences when SP-FT-8, SU-FT-4, WT-FT-4 were used as \ac{o.o.d.} data, and \textsc{combine} performed much better than \textsc{refit-all}. When the rest were used as the \ac{o.o.d.} data, both \textsc{refit-all} and \textsc{combine} gave very low \ac{FNR}. We observed similar phenomena with the power dataset and the chiller dataset. For the power dataset, we also achieved very low \ac{FNR} under both \textsc{refit-all} and \textsc{combine} for every data subgroup was used as \ac{o.o.d.} data except for FT-4, LOC-2, RES-2. For the chiller dataset, performance difference was only significant when RL and CF were held out as \ac{o.o.d.} test set. Again, \textsc{combine} outperformed \textsc{refit-all}. In Fig.~\ref{fig:all-vs-bag}, we only displayed results for the above-mentioned cases where there was significant performance gap between \textsc{refit-all} and \textsc{combine}, and omitted the rest. The low \ac{FNR} in the omitted cases may be a result of the held-out subgroups not being enough ``out-of-distribution''; in other words, the held-out subgroup may resemble one or more of the \ac{i.d.} subgroups which leads to high detection performance. In our upcoming analysis, we will omit these cases as well, and focus on the challenging cases where the held-out test set presents real \ac{o.o.d.} challenges to fault detection models.

To sum up, it is clear that the \textsc{combine} method has lower \ac{FNR} compared with the \textsc{refit-all}, indicating that the \textsc{combine} has a better performance in improving the models' generalization ability. Therefore, in our next experiments, we will only display results from \textsc{combine}.

% The generalization performance of our proposed \ac{SACV} method were evaluated in terms of  We examined the \ac{FNR} for \ac{o.o.d.} data 
% We conducted the experiment . For the AHU dataset and the power dataset, there is one class of fault data having great distribution difference with the others. We therefore use them as our \ac{o.o.d.} test set. And for the chiller dataset, we chose two \ac{o.o.d.} classes because there are two classes being quite different from the rest. 
% First, two ``final model selection'' methods are compared and the experiment results can be seen in Fig.~\ref{fig:all-vs-bag}. It 

%in order to assess the performance of our proposed \ac{SACV} in improving generalization, 
\subsection{Comparing Validation Strategies: {SACV} vs. $k$-fold vs. Holdout}
\begin{figure}[tb]
  \centering
  \begin{subfigure}[t]{0.23\linewidth}
    \centering
    \includegraphics[height=2.6cm]{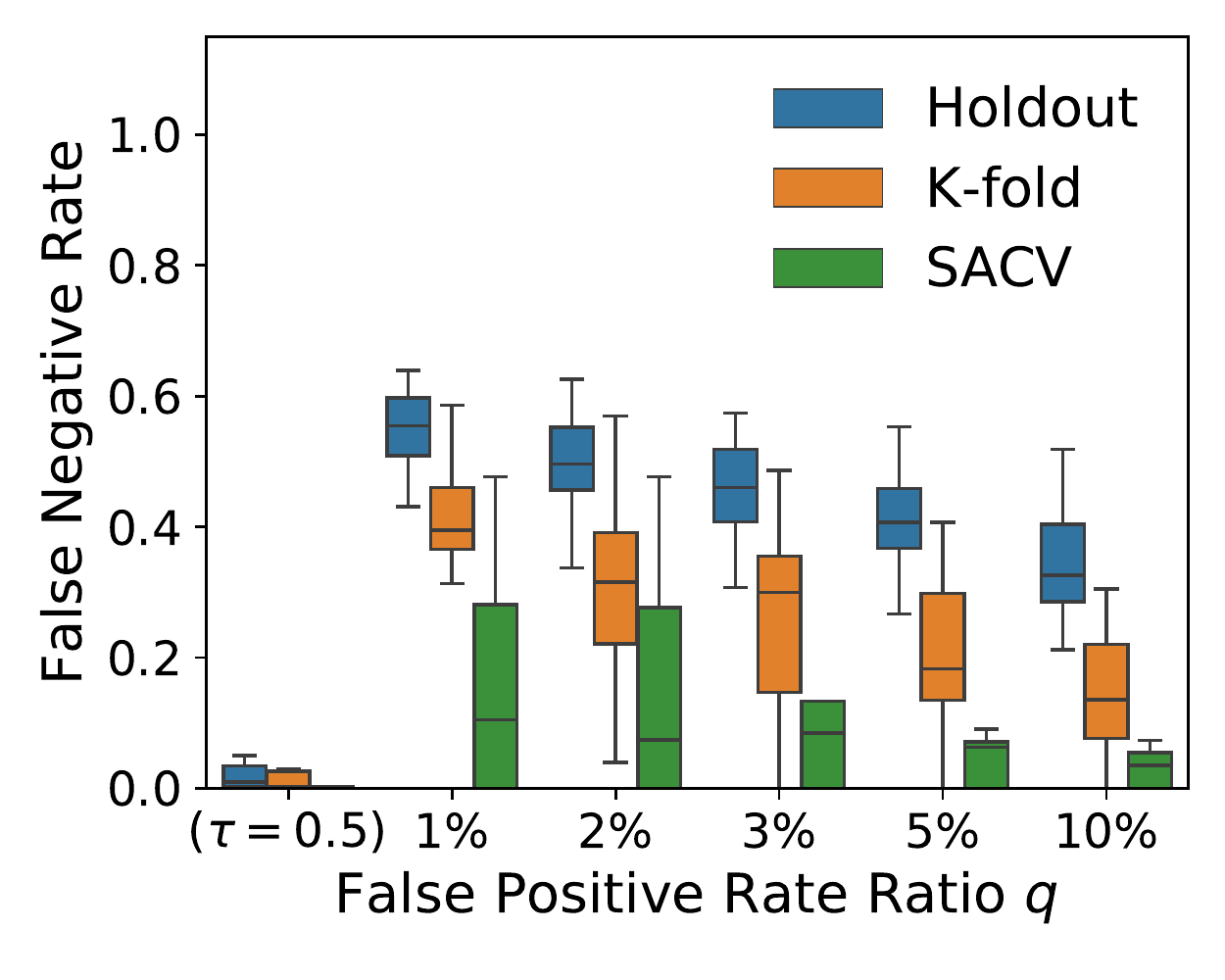}
    \caption{RF: chiller (FT-RL)}
  \end{subfigure}
  \begin{subfigure}[t]{0.23\linewidth}
    \centering
    \includegraphics[height=2.6cm]{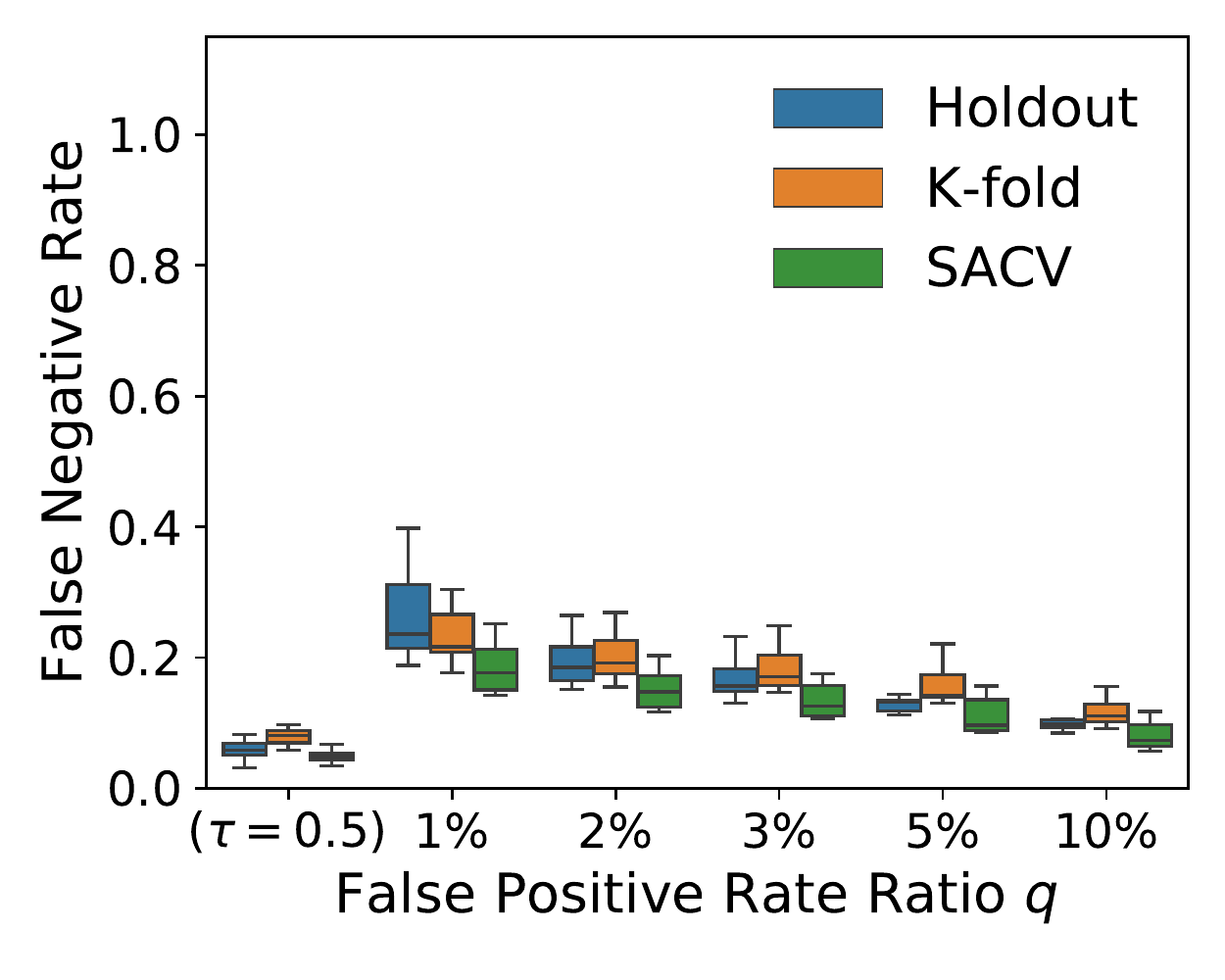}
    \caption{NN: chiller (FT-RL)}
  \end{subfigure}
  \begin{subfigure}[t]{0.23\linewidth}
    \centering
    \includegraphics[height=2.6cm]{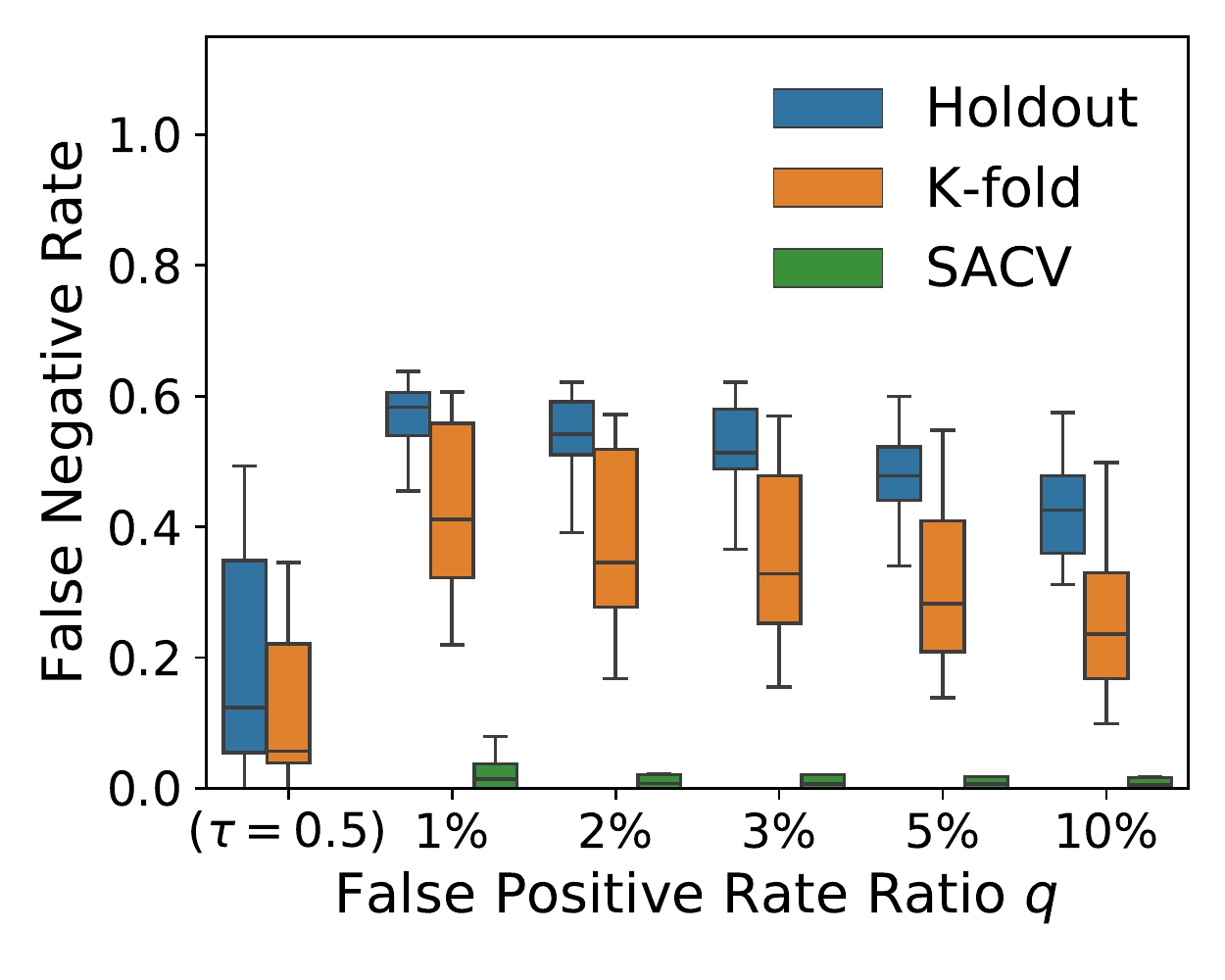}
    \caption{RF: chiller (FT-CF)}
  \end{subfigure}
  \begin{subfigure}[t]{0.23\linewidth}
    \centering
    \includegraphics[height=2.6cm]{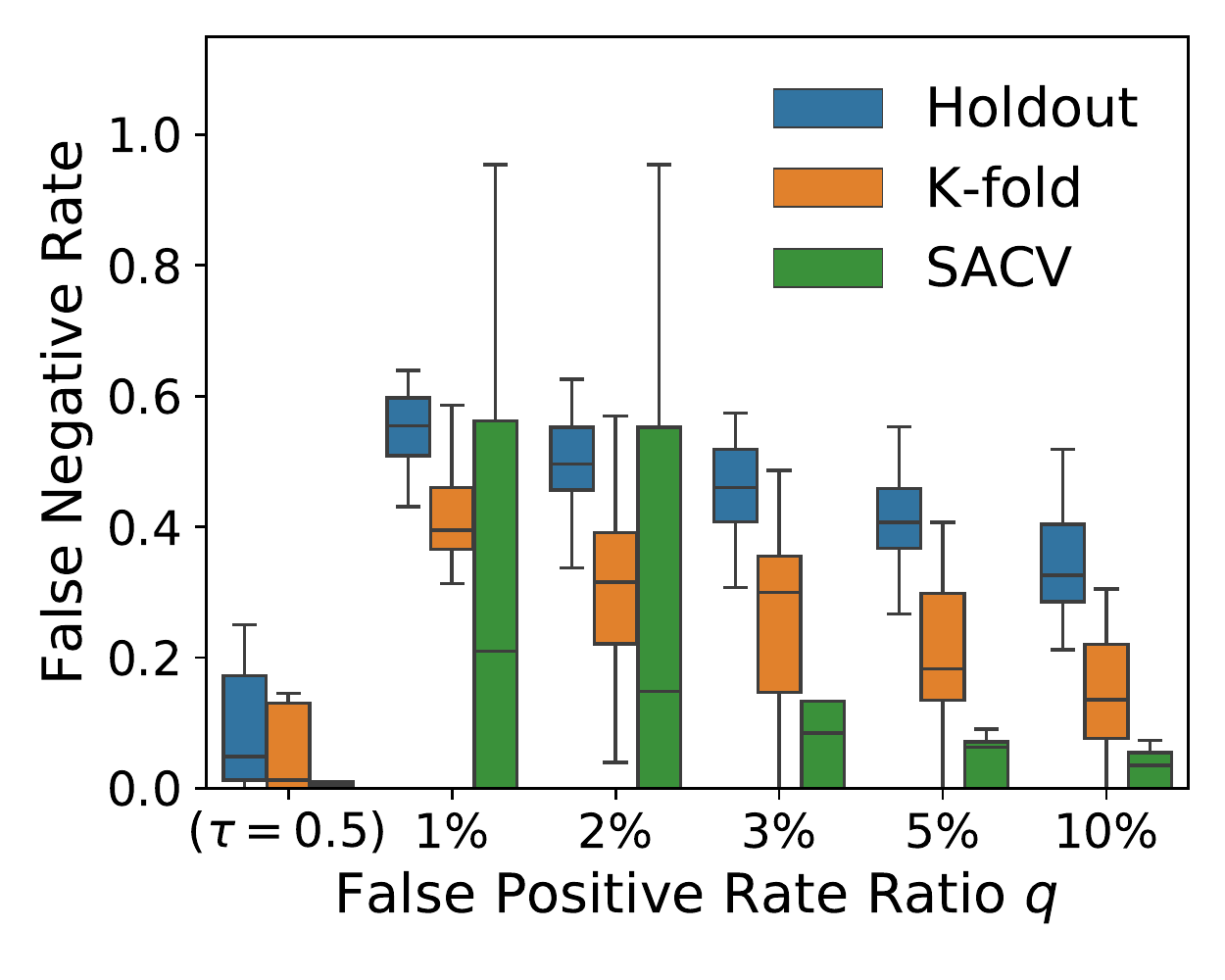}
    \caption{NN: chiller (FT-CF)}
  \end{subfigure}  

  \begin{subfigure}[t]{0.23\linewidth}
    \centering
    \includegraphics[height=2.6cm]{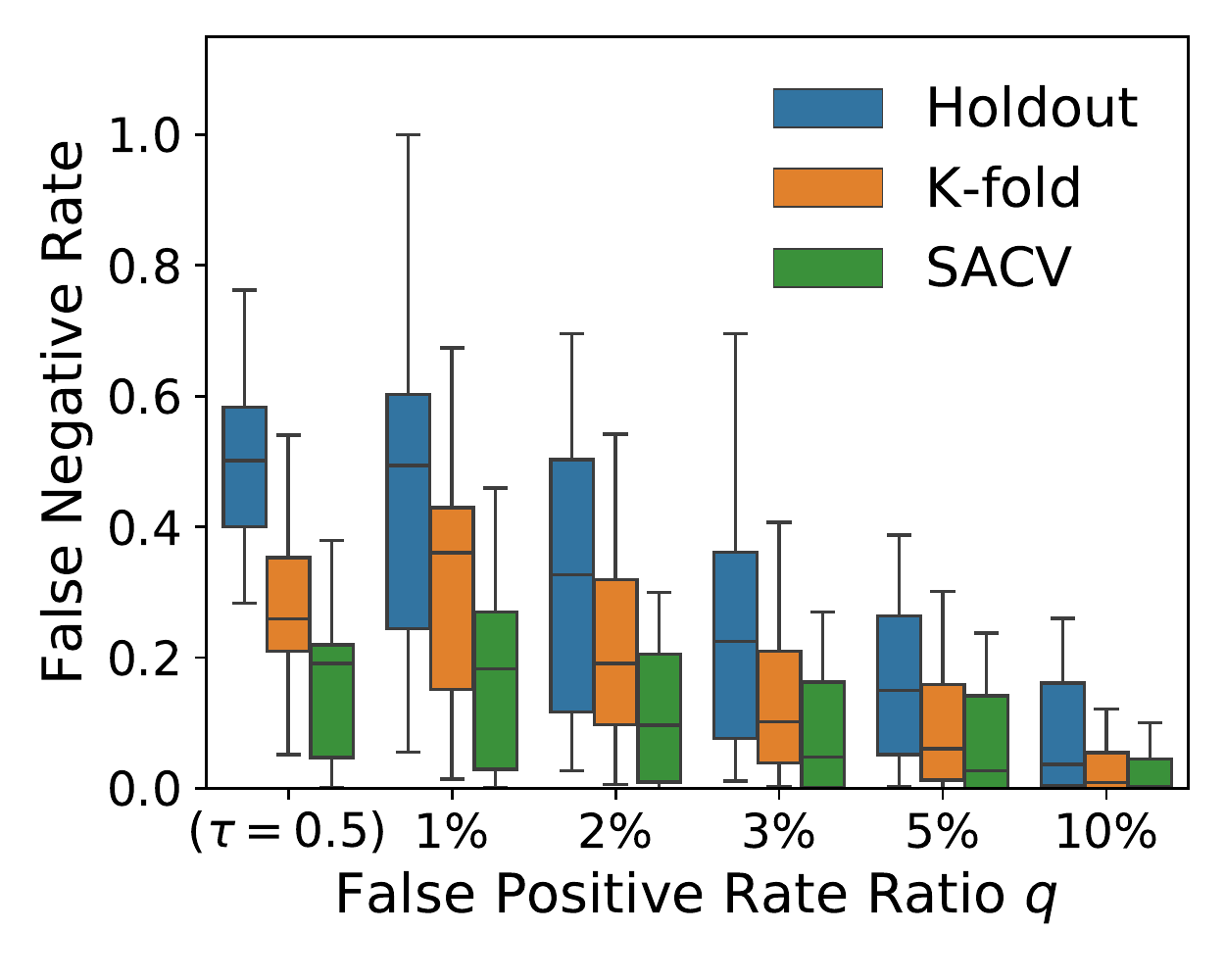}
    \caption{RF: AHU (SP-FT-8)}
  \end{subfigure}
  \begin{subfigure}[t]{0.23\linewidth}
    \centering
    \includegraphics[height=2.6cm]{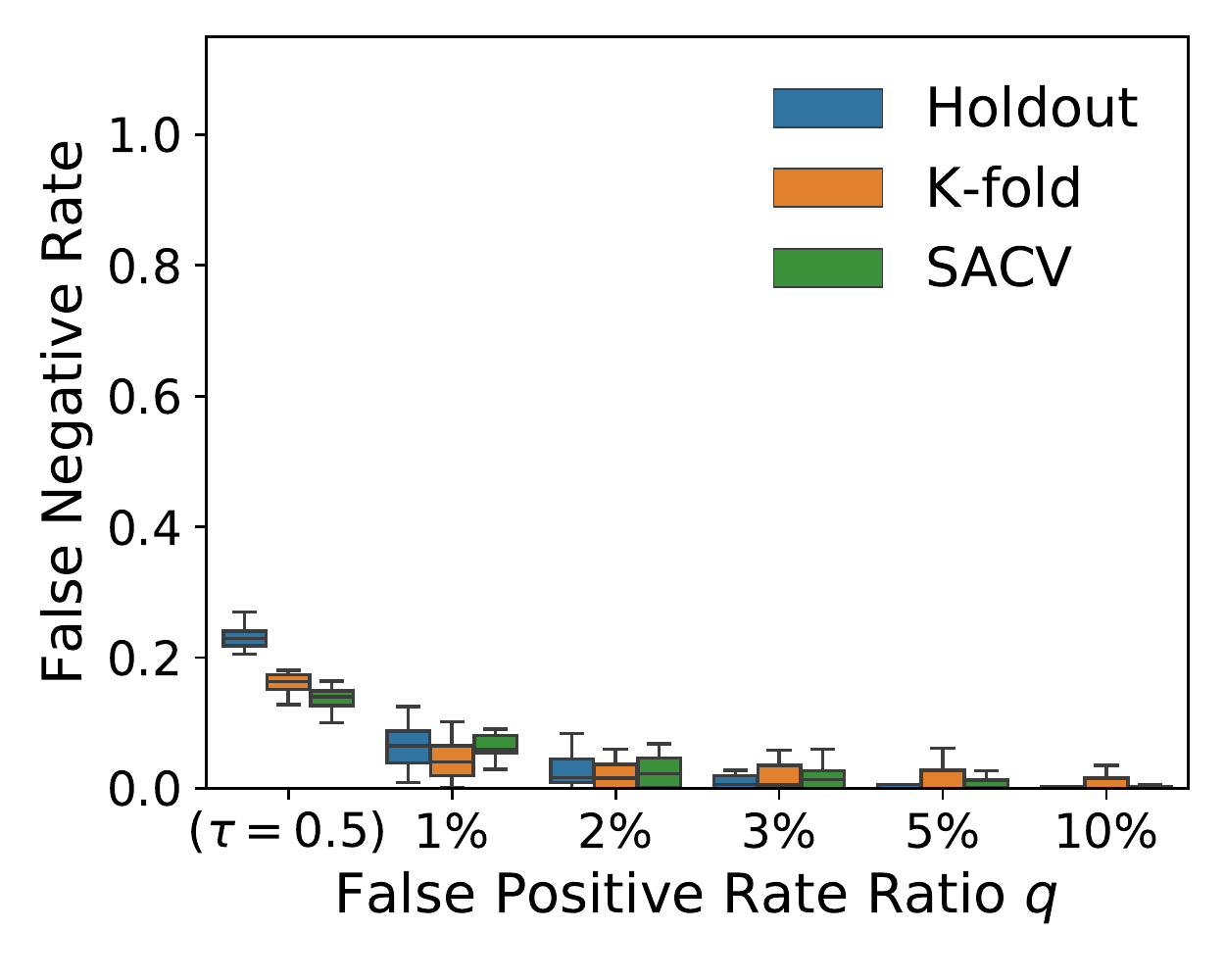}
    \caption{NN: AHU (SP-FT-8)}
  \end{subfigure}
  \begin{subfigure}[t]{0.23\linewidth}
    \centering
    \includegraphics[height=2.6cm]{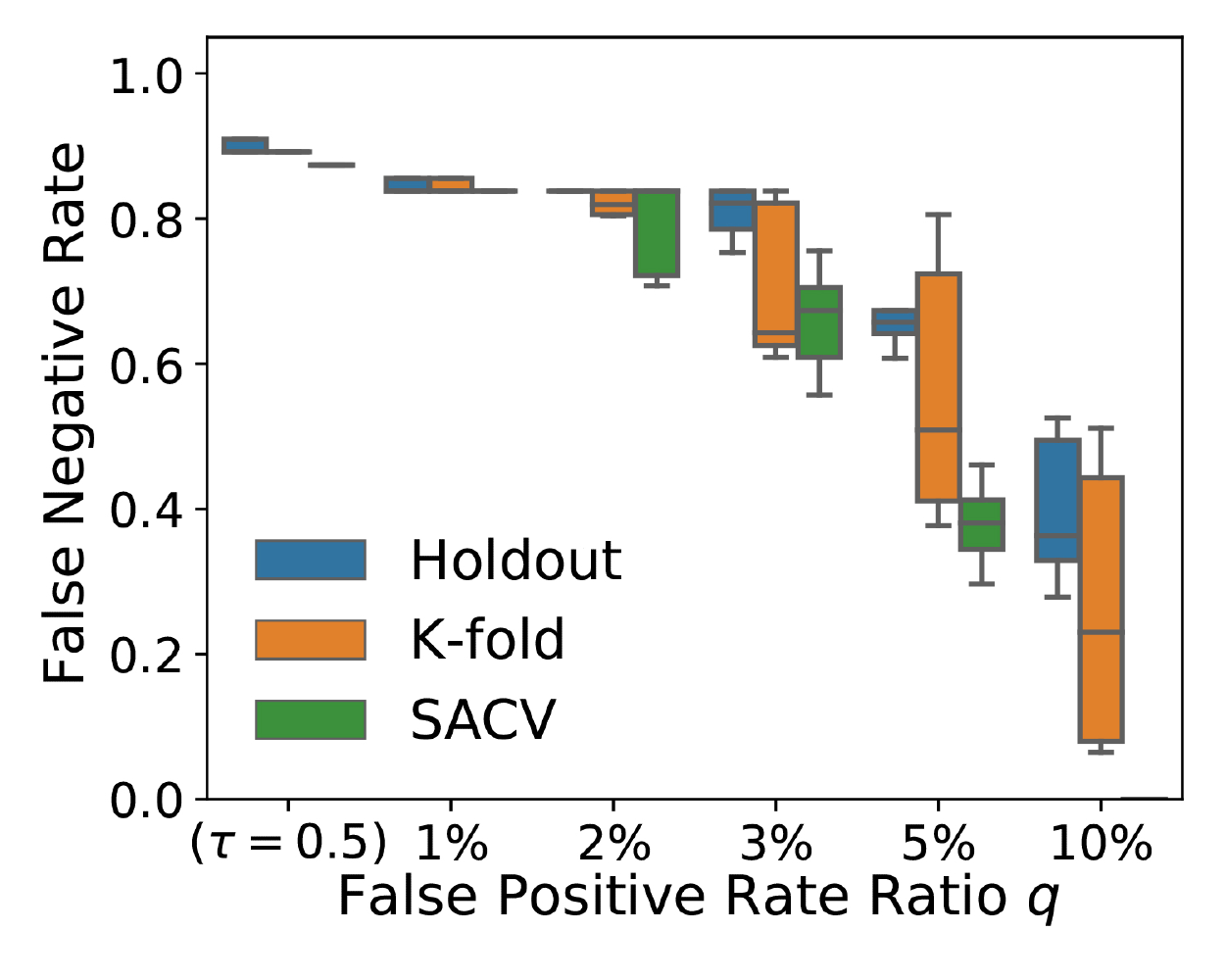}
    \caption{RF: power (FT-4)}
  \end{subfigure}
  \begin{subfigure}[t]{0.23\linewidth}
    \centering
    \includegraphics[height=2.6cm]{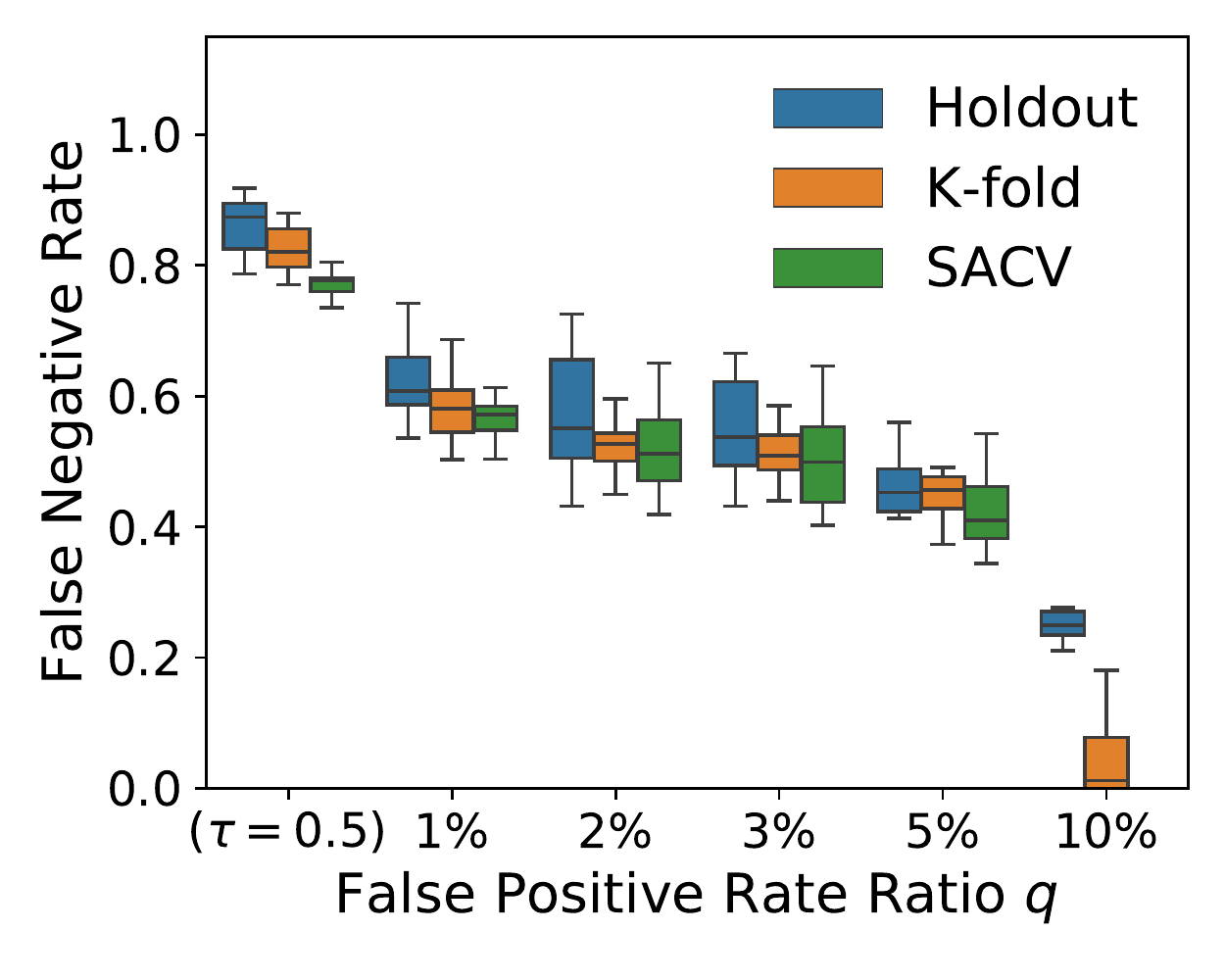}
    \caption{NN: power (FT-4)}
  \end{subfigure}
  
  \begin{subfigure}[t]{0.23\linewidth}
    \centering
    \includegraphics[height=2.6cm]{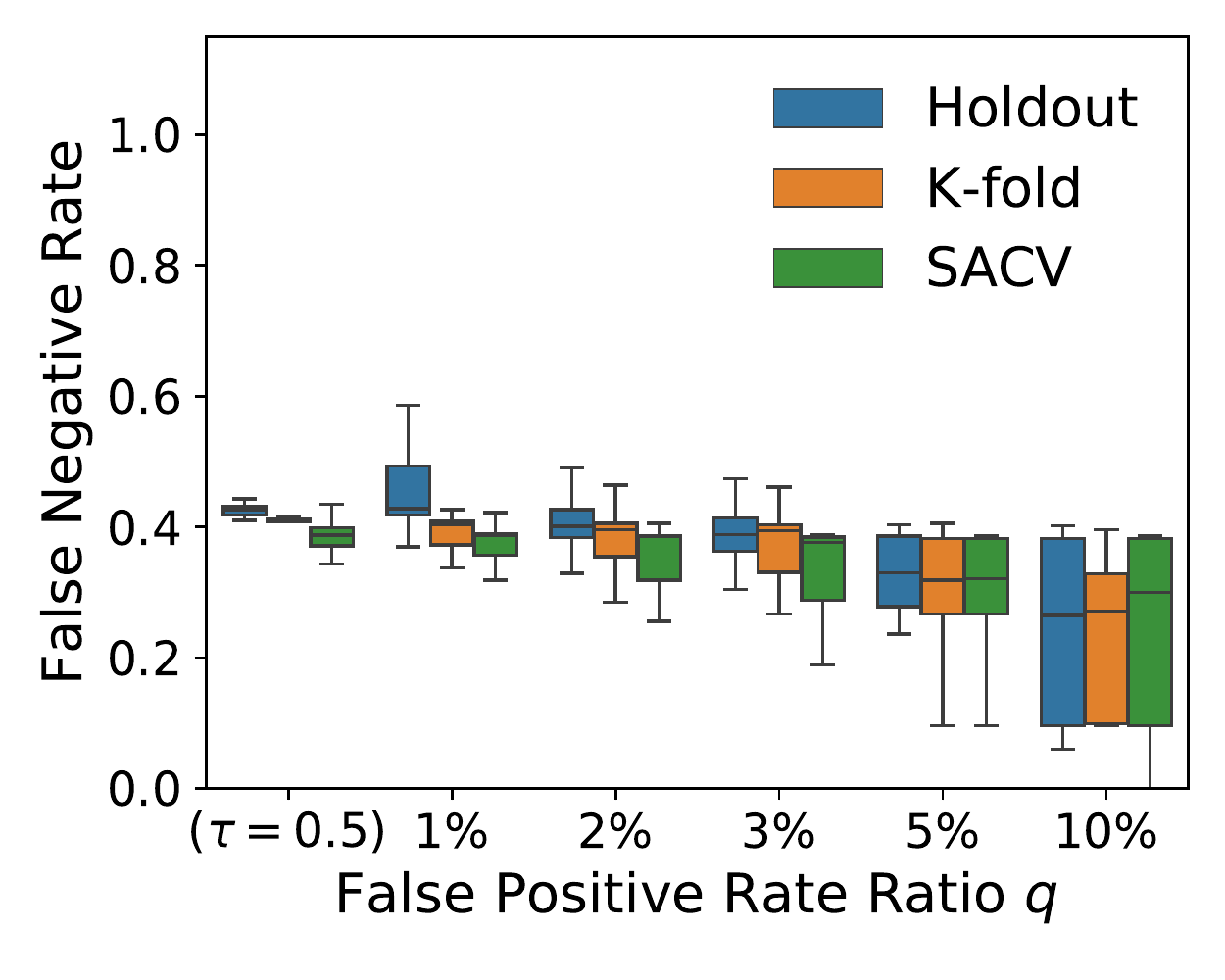}\caption{RF: AHU (SU-FT-4)}
  \end{subfigure}
  \begin{subfigure}[t]{0.23\linewidth}
    \centering
    \includegraphics[height=2.6cm]{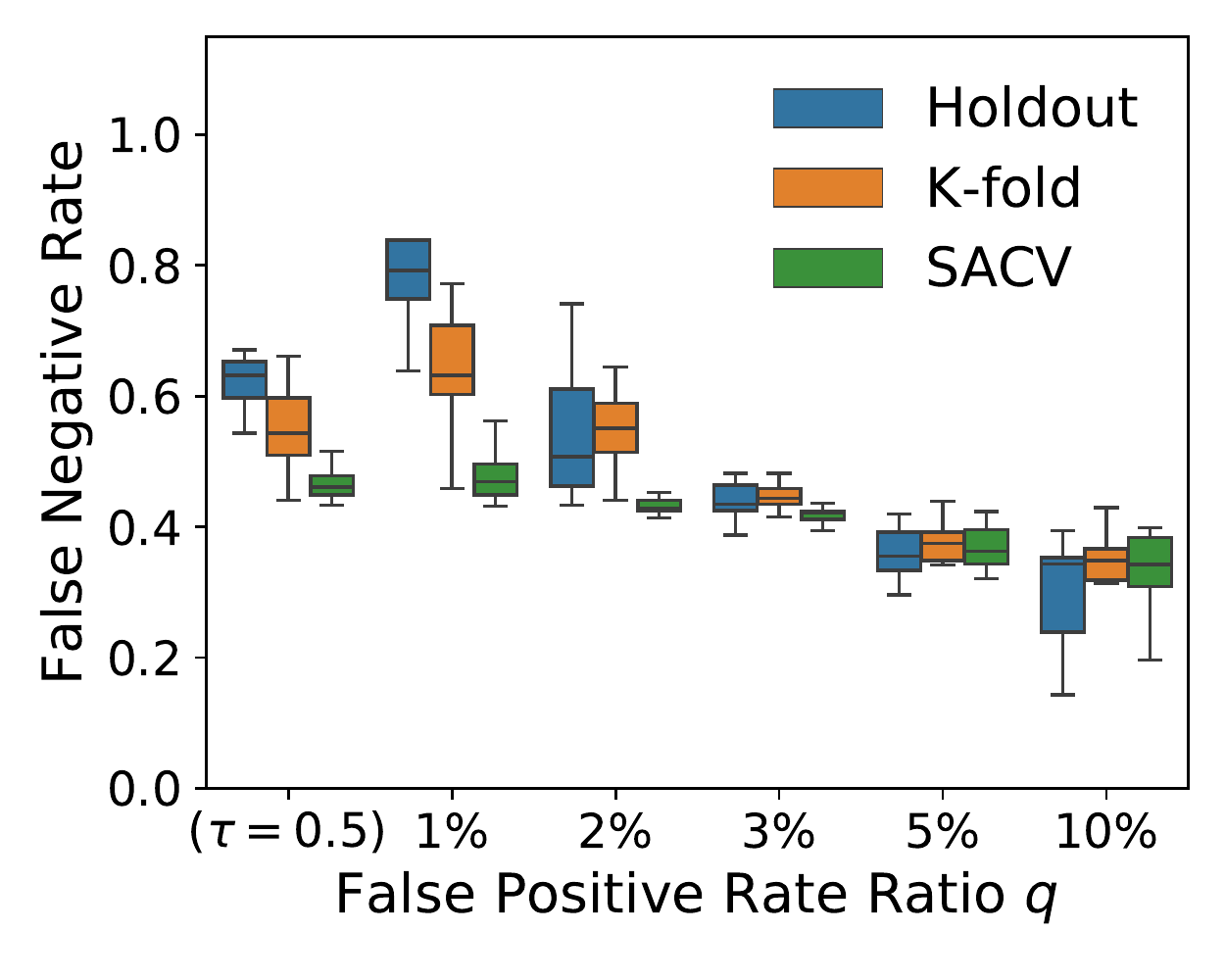}\caption{NN: AHU (SU-FT-4)}
  \end{subfigure}
  \begin{subfigure}[t]{0.23\linewidth}
    \centering
    \includegraphics[height=2.6cm]{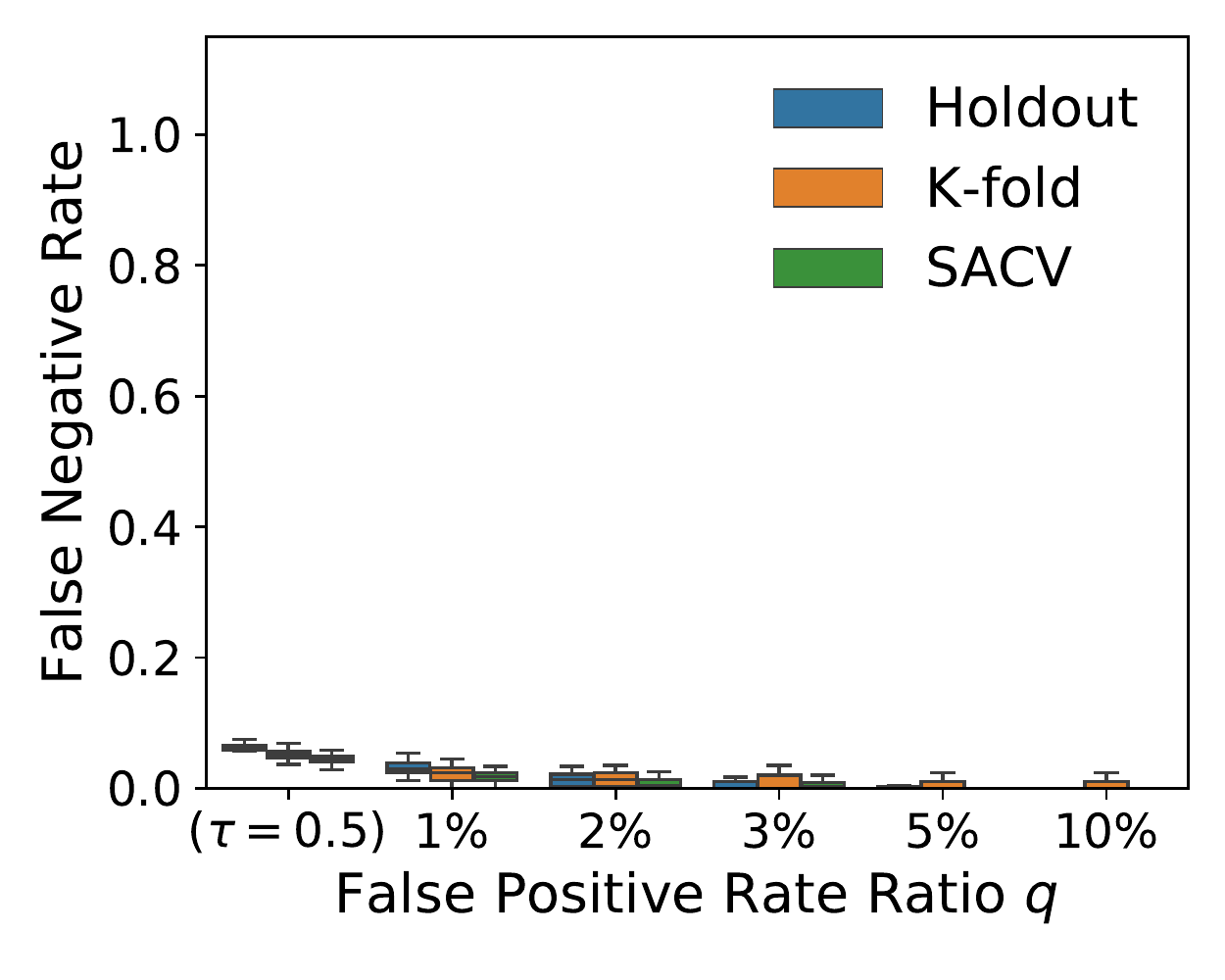}\caption{RF: power (LOC-2)}
  \end{subfigure}
  \begin{subfigure}[t]{0.23\linewidth}
    \centering
    \includegraphics[height=2.6cm]{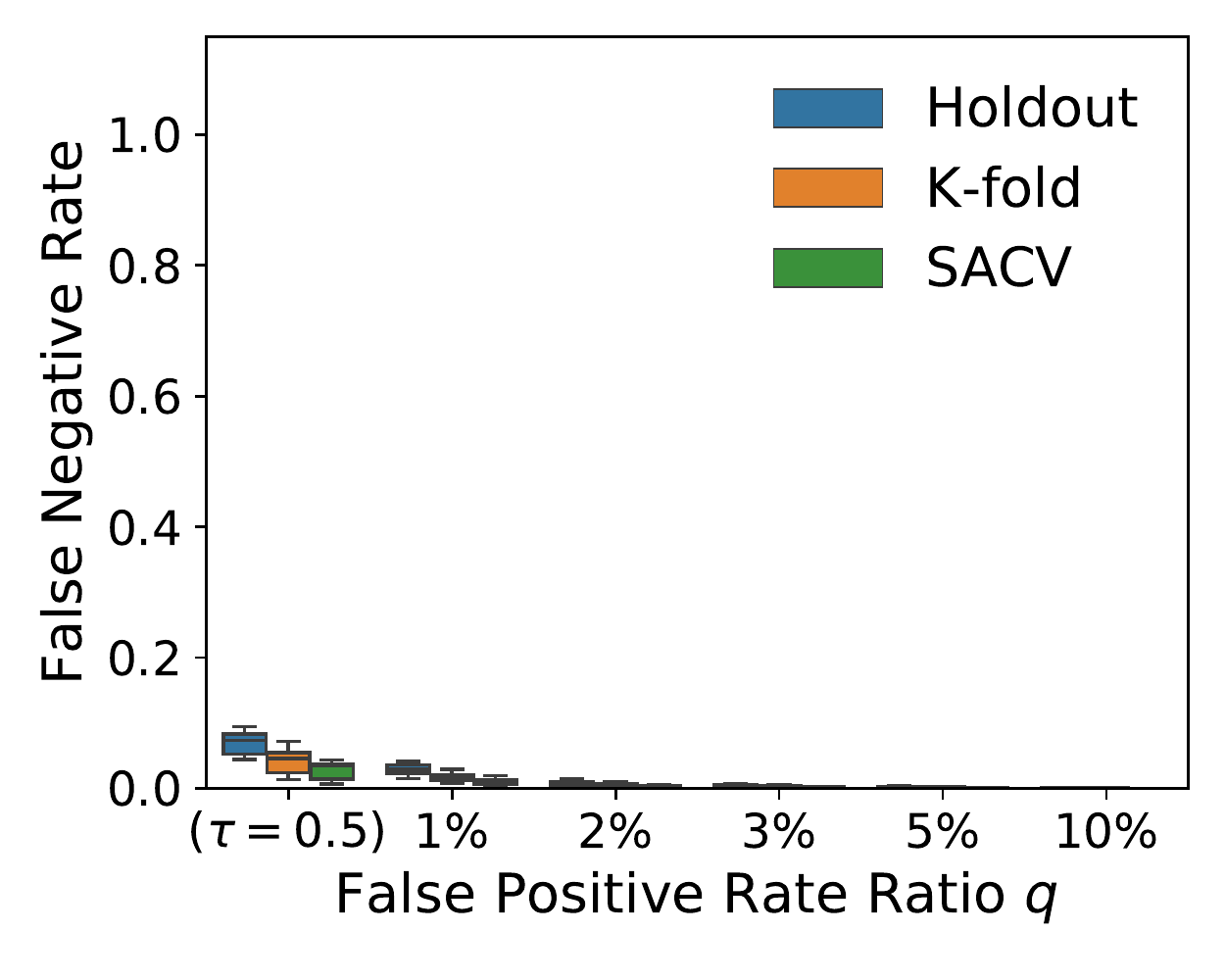}
    \caption{NN: power (LOC-2)}
  \end{subfigure}
  
  \begin{subfigure}[t]{0.23\linewidth}
    \centering
    \includegraphics[height=2.6cm]{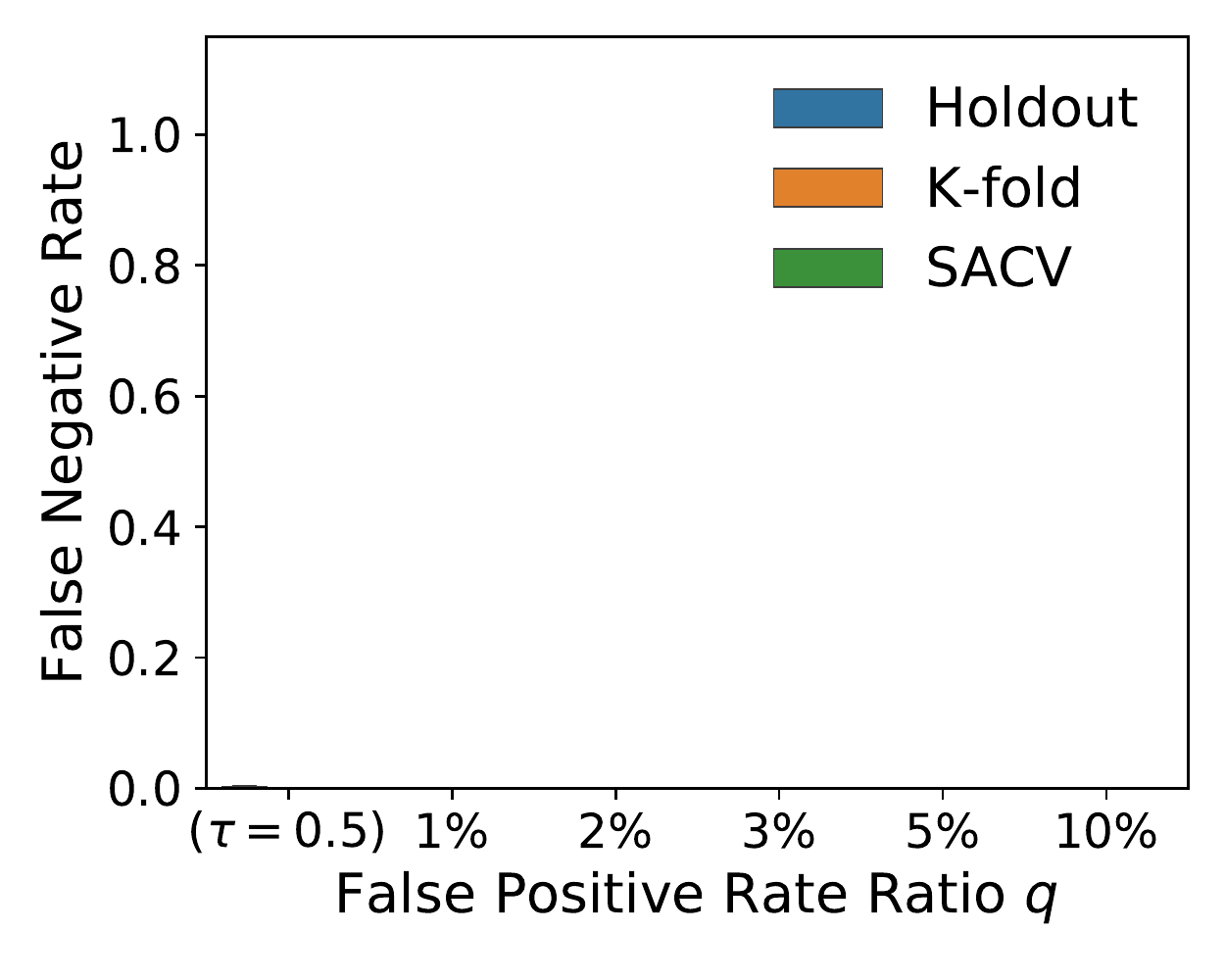}
    \caption{RF: AHU (WT-FT-2)}
  \end{subfigure}
  \begin{subfigure}[t]{0.23\linewidth}
    \centering
    \includegraphics[height=2.6cm]{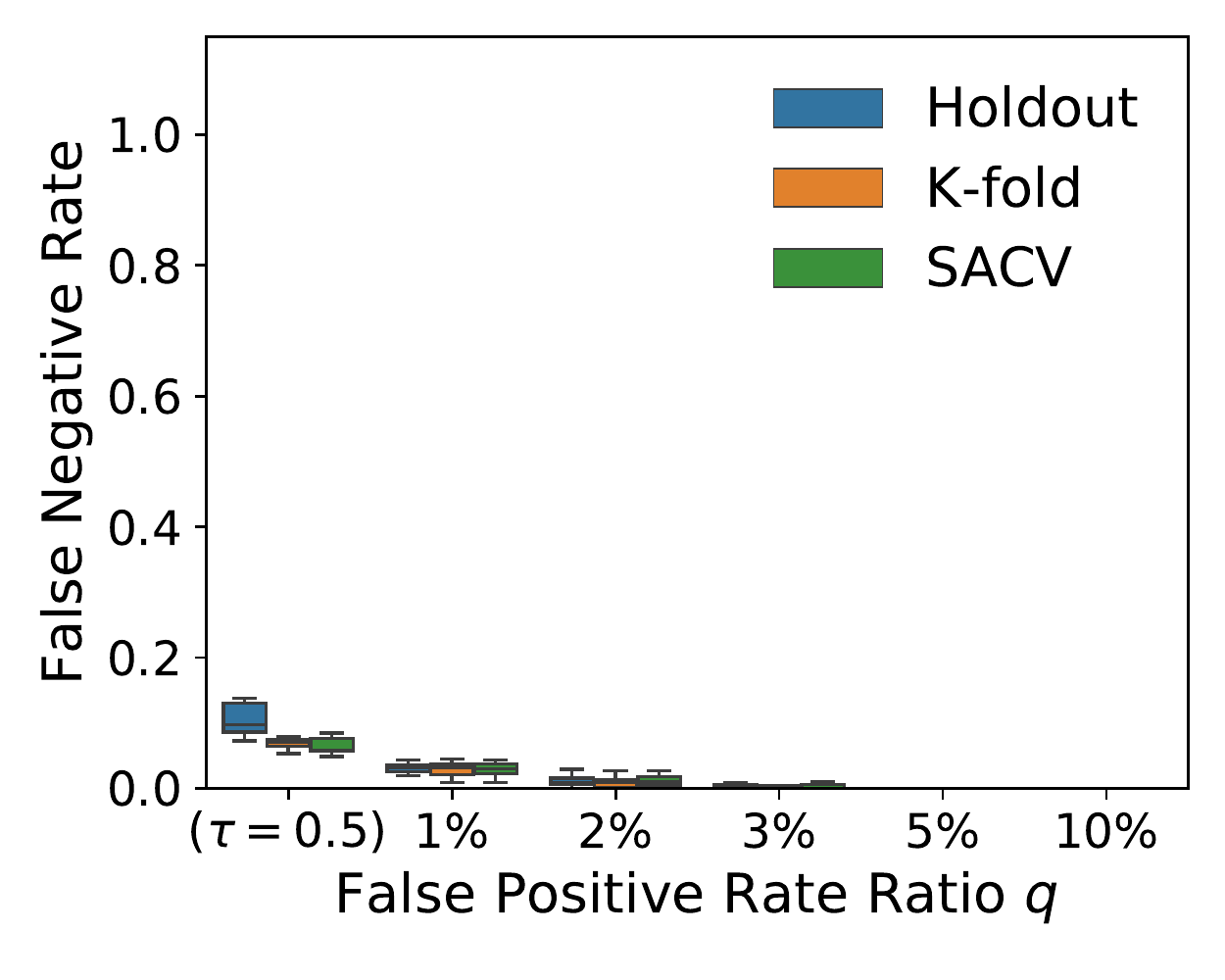}
    \caption{NN: AHU (WT-FT-2)}
  \end{subfigure}
  \begin{subfigure}[t]{0.23\linewidth}
    \centering
    \includegraphics[height=2.6cm]{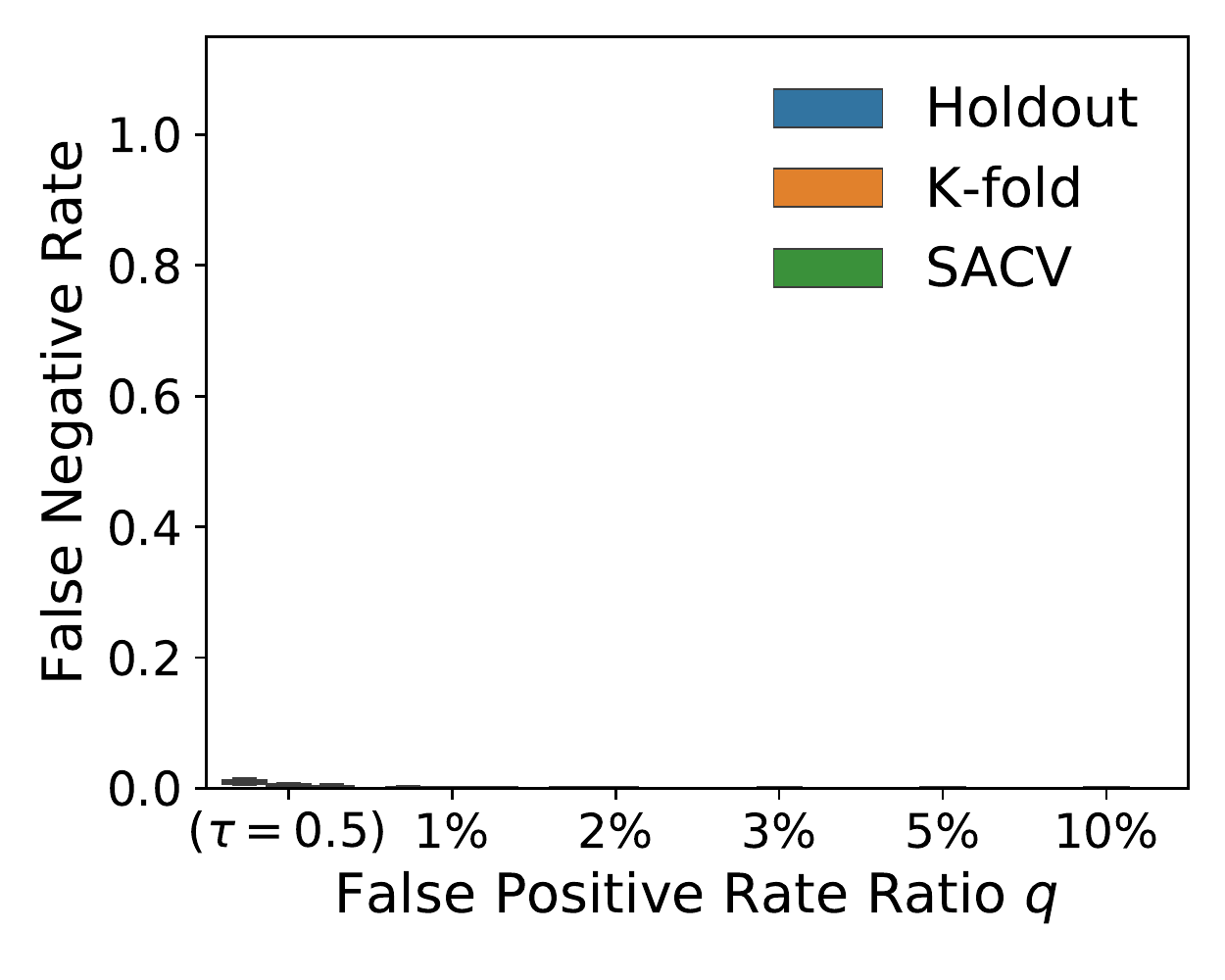}
    \caption{RF: power (RES-2)}
  \end{subfigure}
  \begin{subfigure}[t]{0.23\linewidth}
    \centering
    \includegraphics[height=2.6cm]{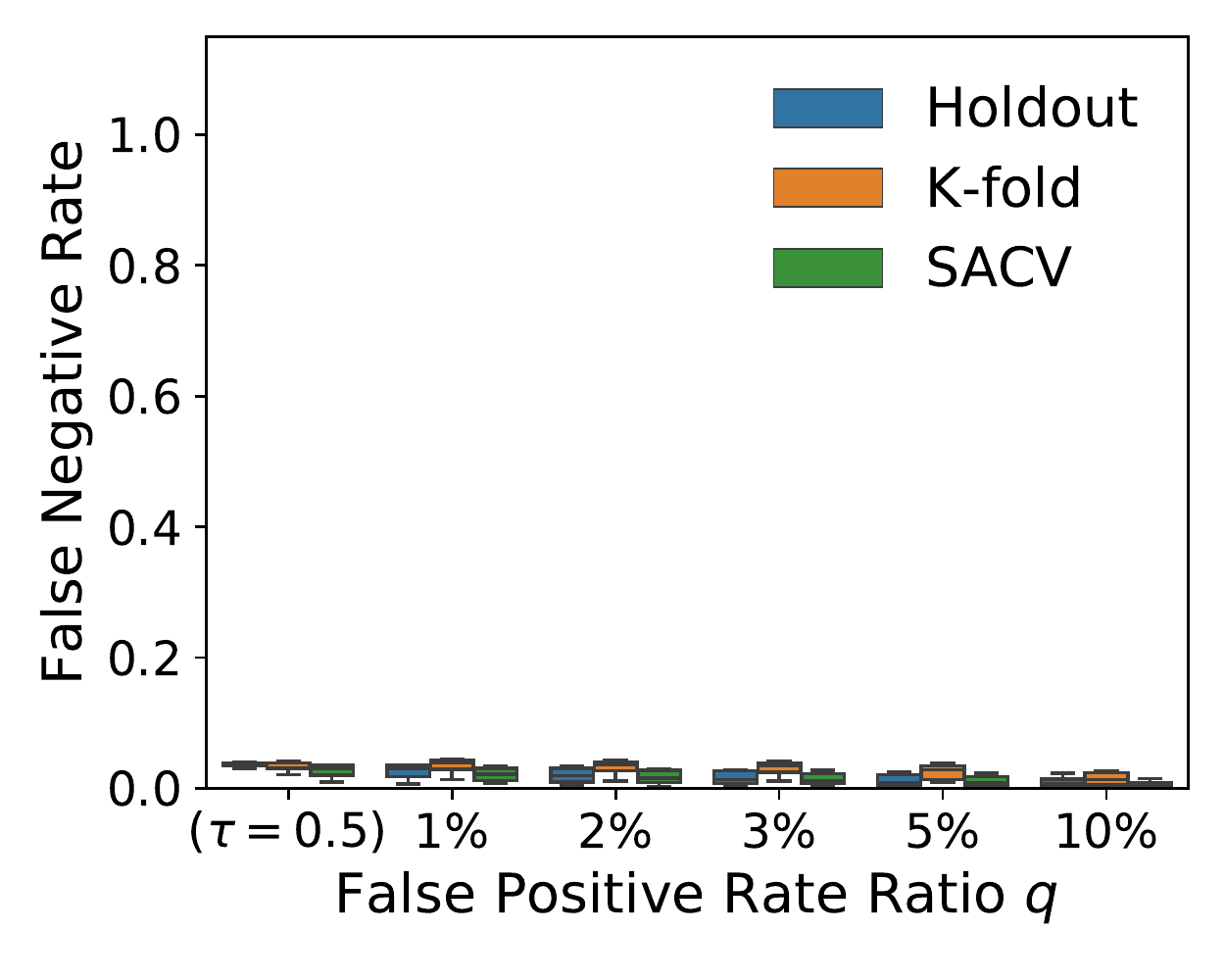}
    \caption{NN: power (RES-2)}
  \end{subfigure}  
  \caption{The \ac{FNR} given by different (cross-)validation methods: 1) holdout, 2) $k$-fold, and 3) \ac{SACV}. Results from \ac{RF} and \ac{NN} on different datasets are presented: 1) the chiller dataset, 2) the AHU dataset and 3) the power dataset.}
  \label{fig:FNR-results}
\end{figure}

Next, we evaluated the ensemble methods' performance on the \ac{o.o.d.} data when different validation strategies are used. As in the previous experiment, we examined the \acp{FNR} across different configurations of $\tau$ (by directly setting $\tau=0.5$ or varying $q$). For comparison, we used the holdout validation and the $k$-fold cross-validation as our baselines. The number of splits used in $k$-fold cross-validation is set to be equal to the number of classes of the development set, i.e. $k = K^\text{i.d.}$. We visualized the results from same subgroups as introduced in the previous analysis. The results can be found in Fig.~\ref{fig:FNR-results}. 

Comparing the three validation strategies, we can clearly see in Fig.~\ref{fig:FNR-results} that \ac{SACV} achieved significant improvement in \ac{FNR} over the other two validation strategies, indicating that \ac{SACV} is indeed effective in improving the models' generalization performance. In Fig.~\ref{fig:FNR-results}, we only showed the results for a selected number of cases where baseline methods performed poorly on the held out \ac{o.o.d.} data, and omitted the rest since the baseline \acp{FNR} for these omitted cases are already close to zero. In addition, we can also see from the results that the \acp{FNR} decrease with the increment of fixed q.

\subsection{Comparing Uncertainty Metrics: \textsc{mean} vs. \textsc{var}}
Finally, we compared the different metrics used for uncertainty estimating including: 1) \textsc{mean}, 2) \textsc{var}, described in Sec.~\ref{sec:uncertainty-estimation}. We evaluated our models' generalization performance by calculating the number of remaining false negatives after applying uncertainty estimation, assuming all of the identified false negatives can be corrected perfectly by human experts. 

% In addition, we also used the results given by \testsc{baseline} as the lower bound of performance, which means doing thresholding without utilizing any uncertainty metrics.

\begin{figure}[tb]
  \centering
  \begin{subfigure}[t]{0.23\linewidth}
    \centering
    \includegraphics[height=2.6cm]{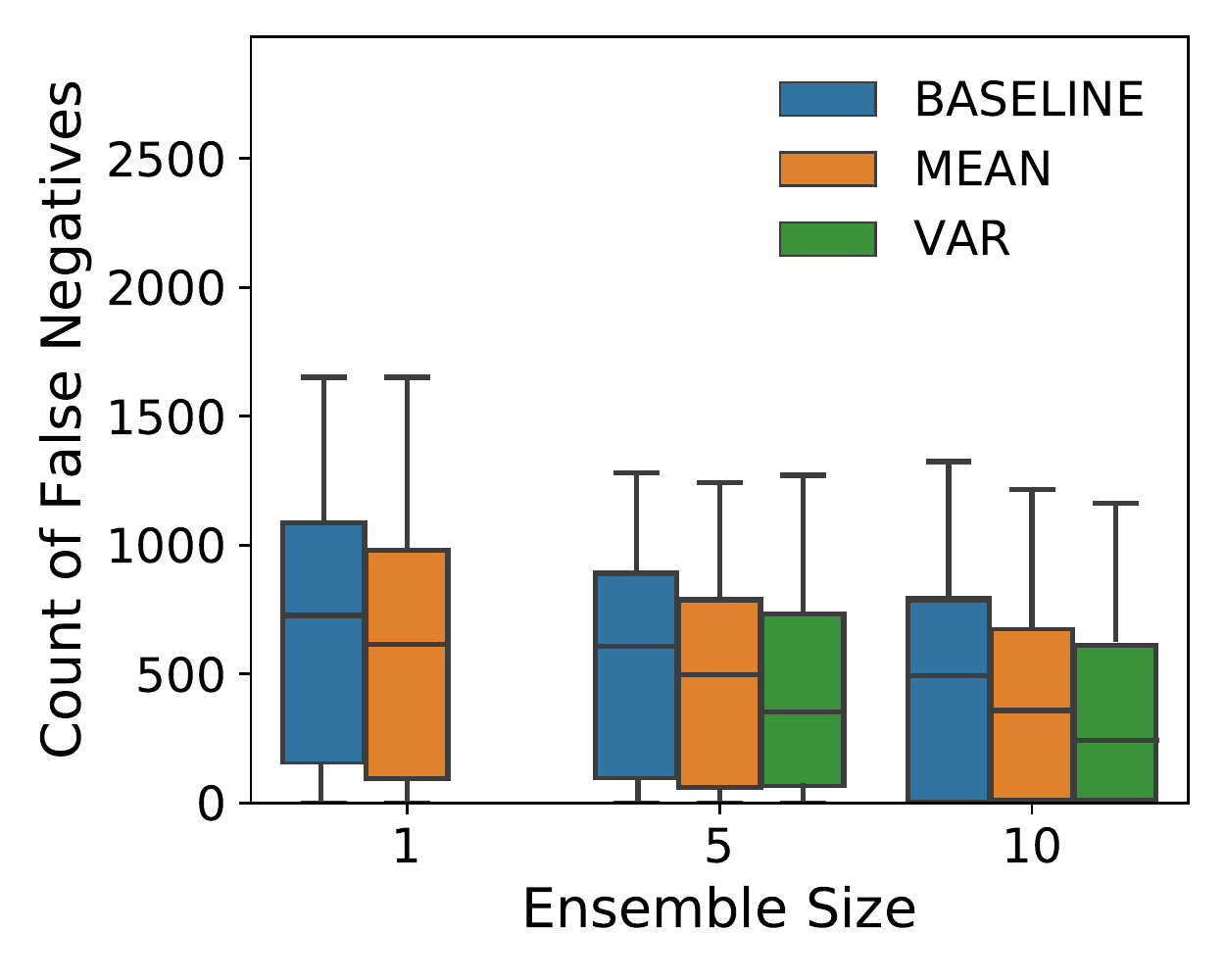}
    \caption{RF: chiller (FT-RL)}
  \end{subfigure}
  \begin{subfigure}[t]{0.23\linewidth}
    \centering
    \includegraphics[height=2.6cm]{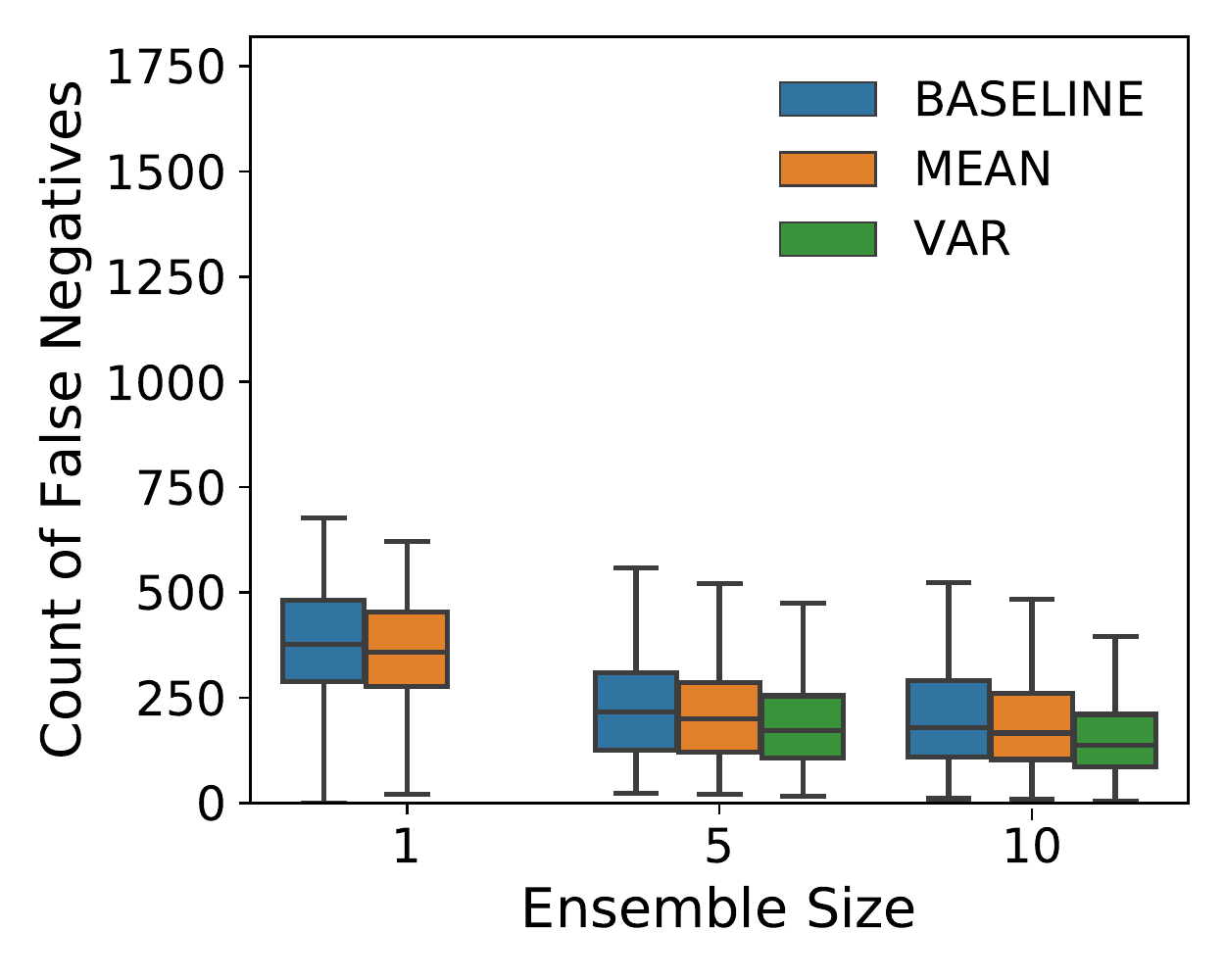}
    \caption{NN: chiller (FT-RL)}
  \end{subfigure}
  \begin{subfigure}[t]{0.23\linewidth}
    \centering
    \includegraphics[height=2.6cm]{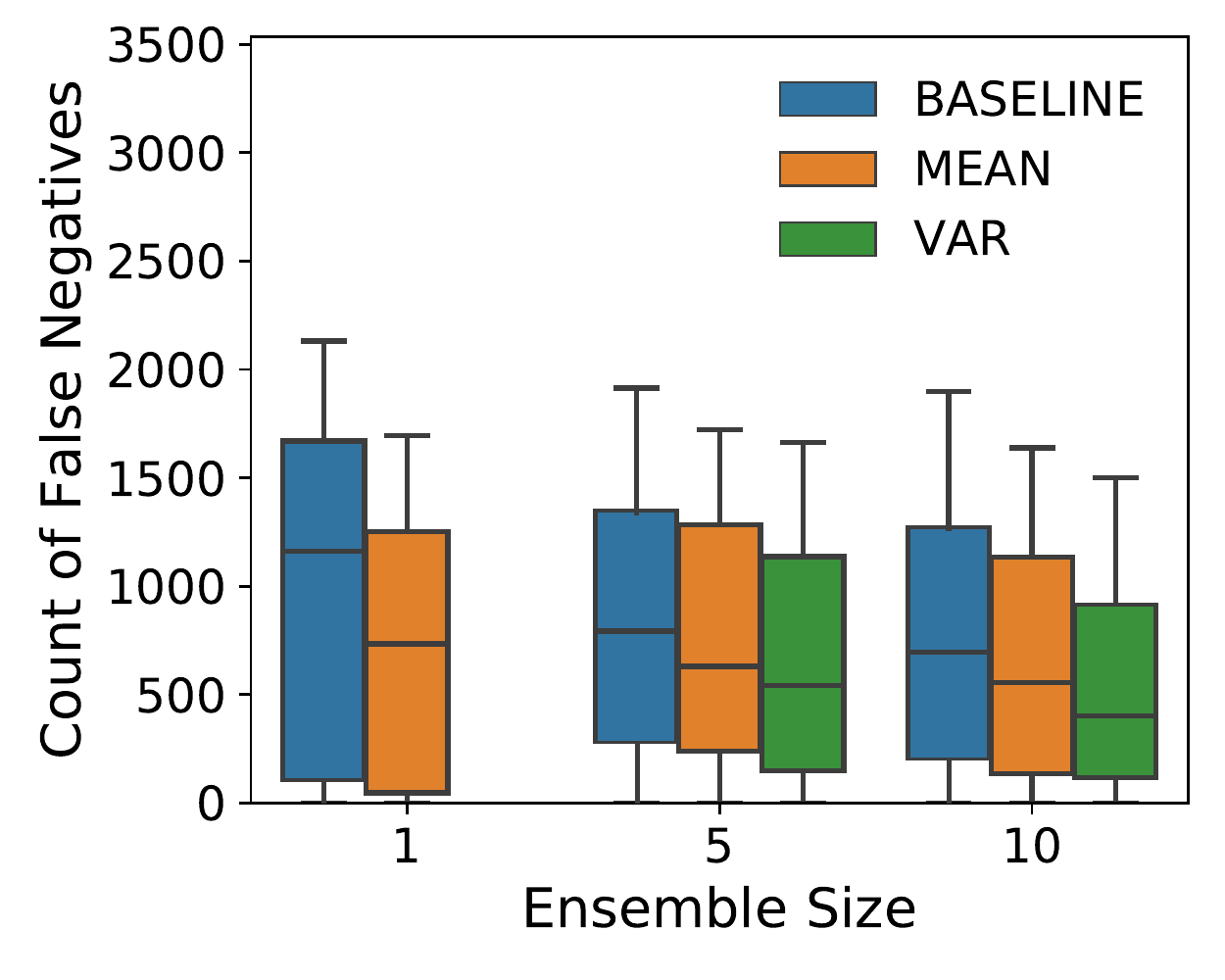}
    \caption{RF: chiller (FT-CF)}
  \end{subfigure}
  \begin{subfigure}[t]{0.23\linewidth}
    \centering
    \includegraphics[height=2.6cm]{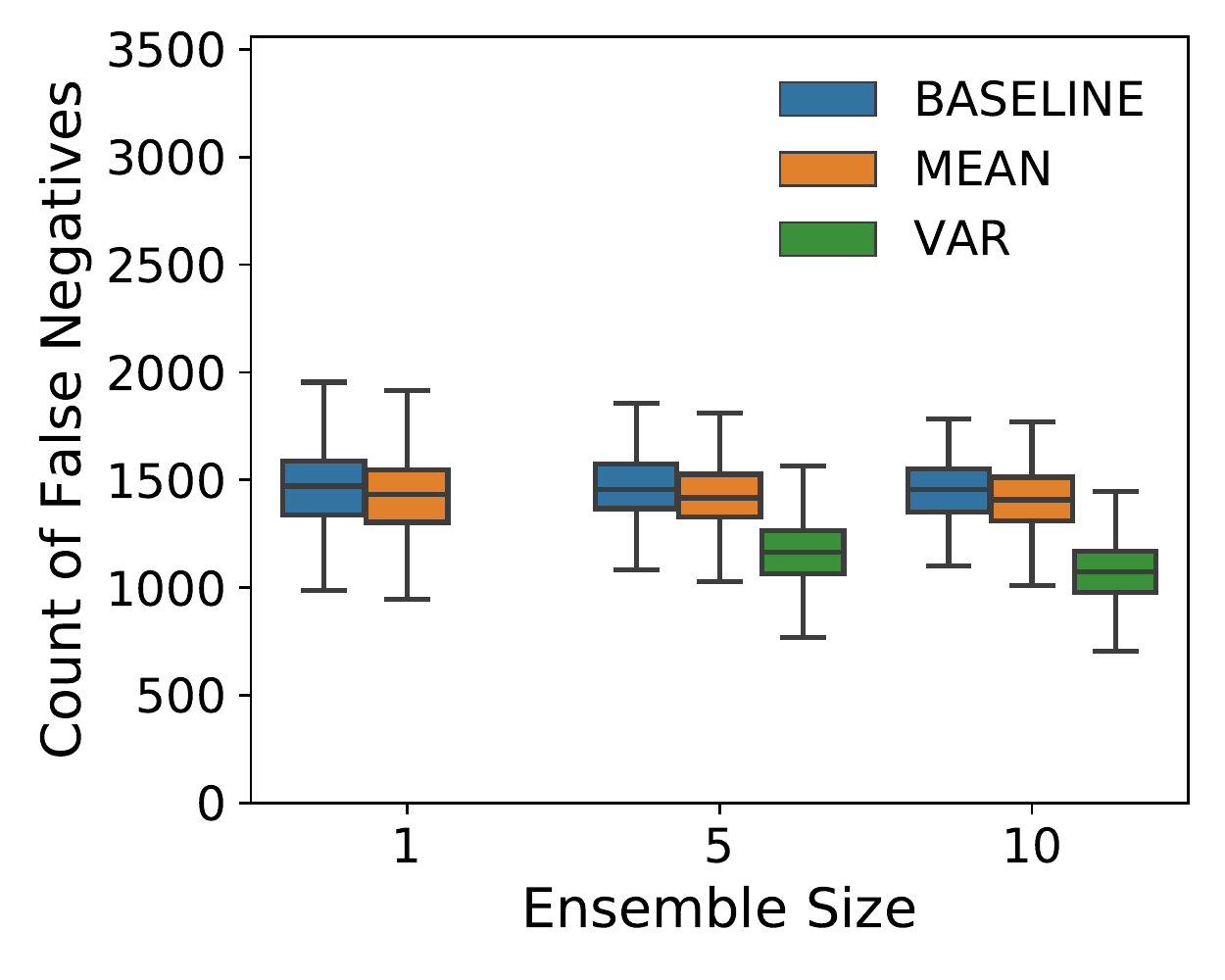}
    \caption{NN: chiller (FT-CF)}
  \end{subfigure}  

  \begin{subfigure}[t]{0.23\linewidth}
    \centering
    \includegraphics[height=2.6cm]{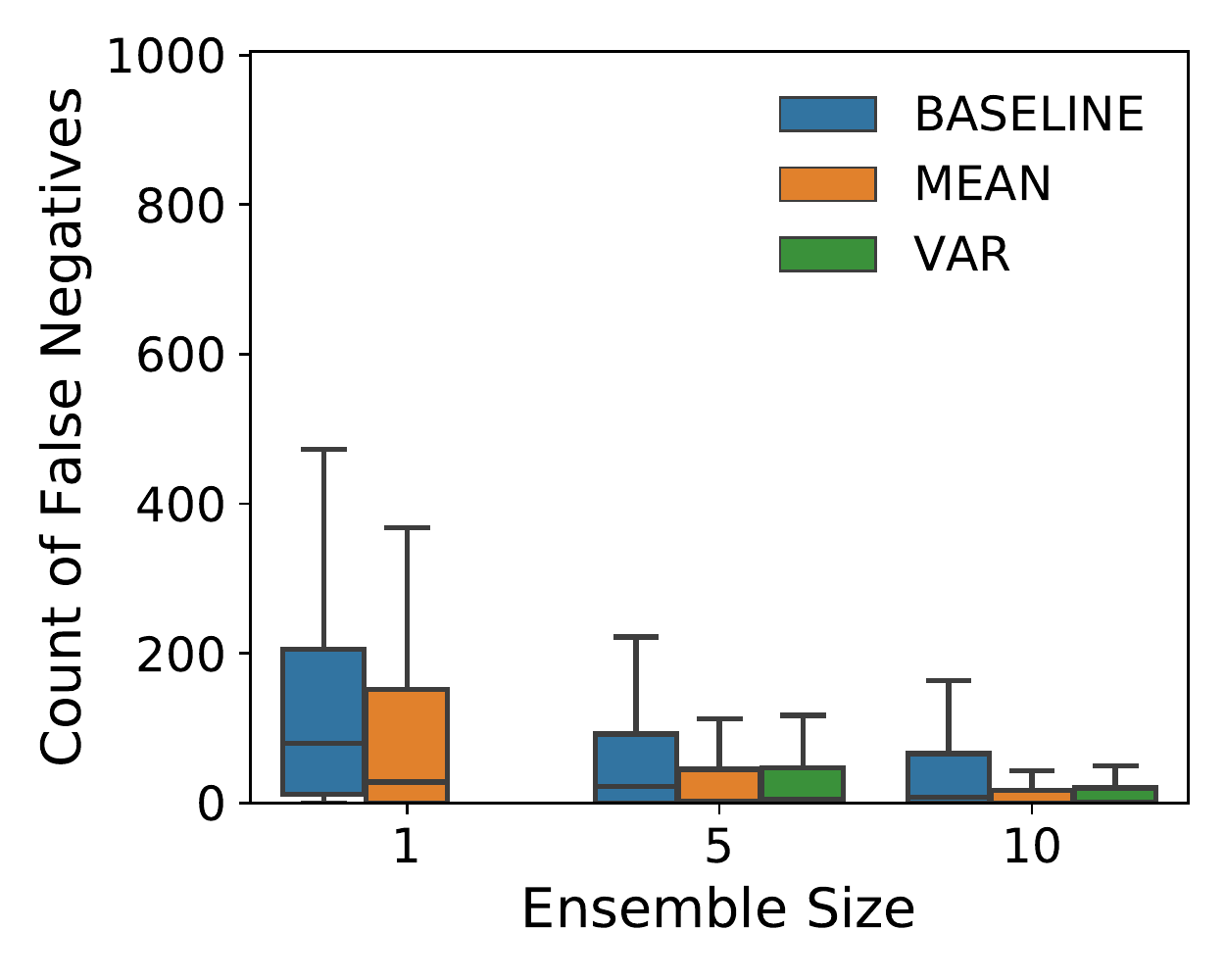}
    \caption{RF: AHU (SP-FT-8)}
  \end{subfigure}
  \begin{subfigure}[t]{0.23\linewidth}
    \centering
    \includegraphics[height=2.6cm]{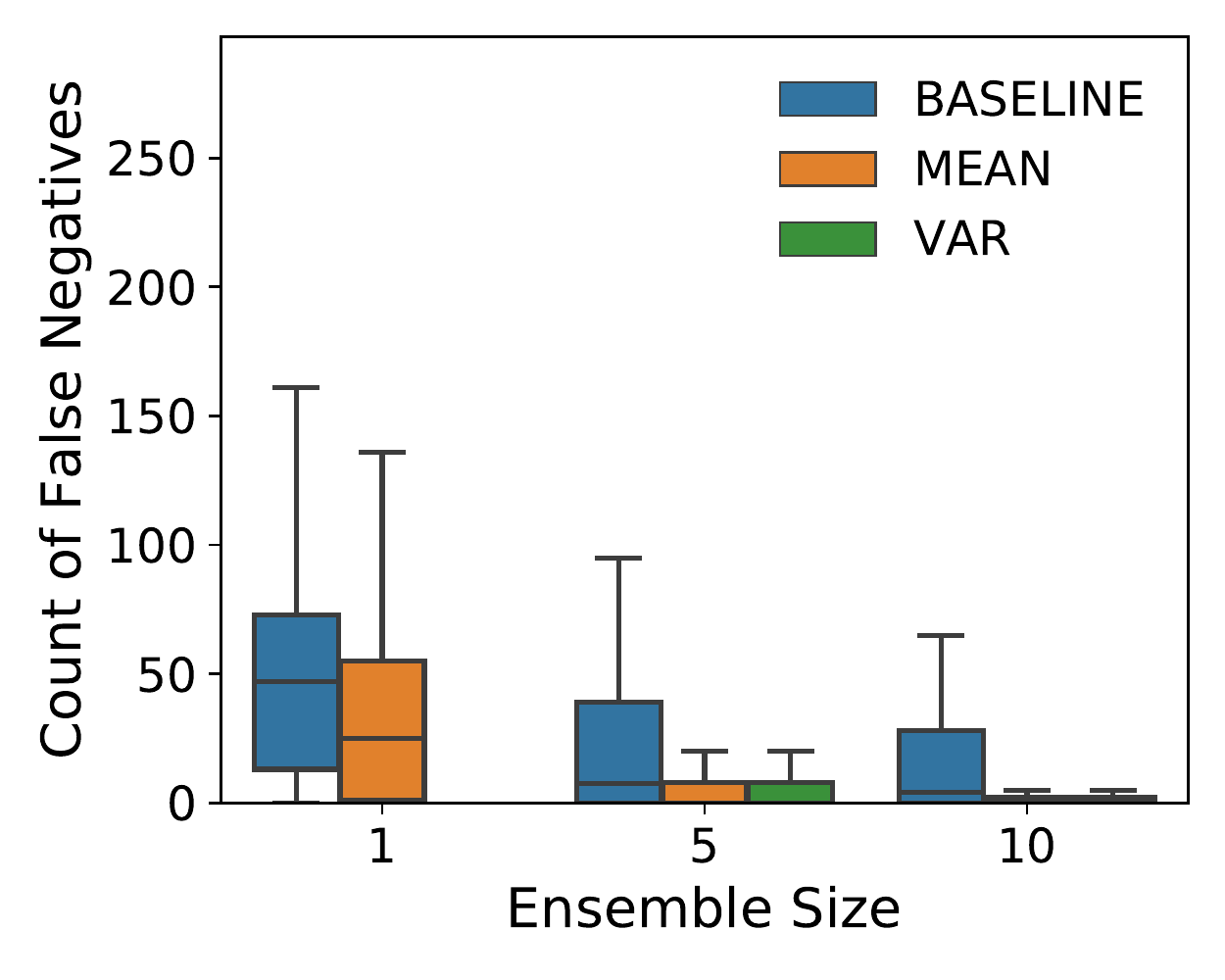}
    \caption{NN: AHU (SP-FT-8)}
  \end{subfigure}
  \begin{subfigure}[t]{0.23\linewidth}
    \centering
    \includegraphics[height=2.6cm]{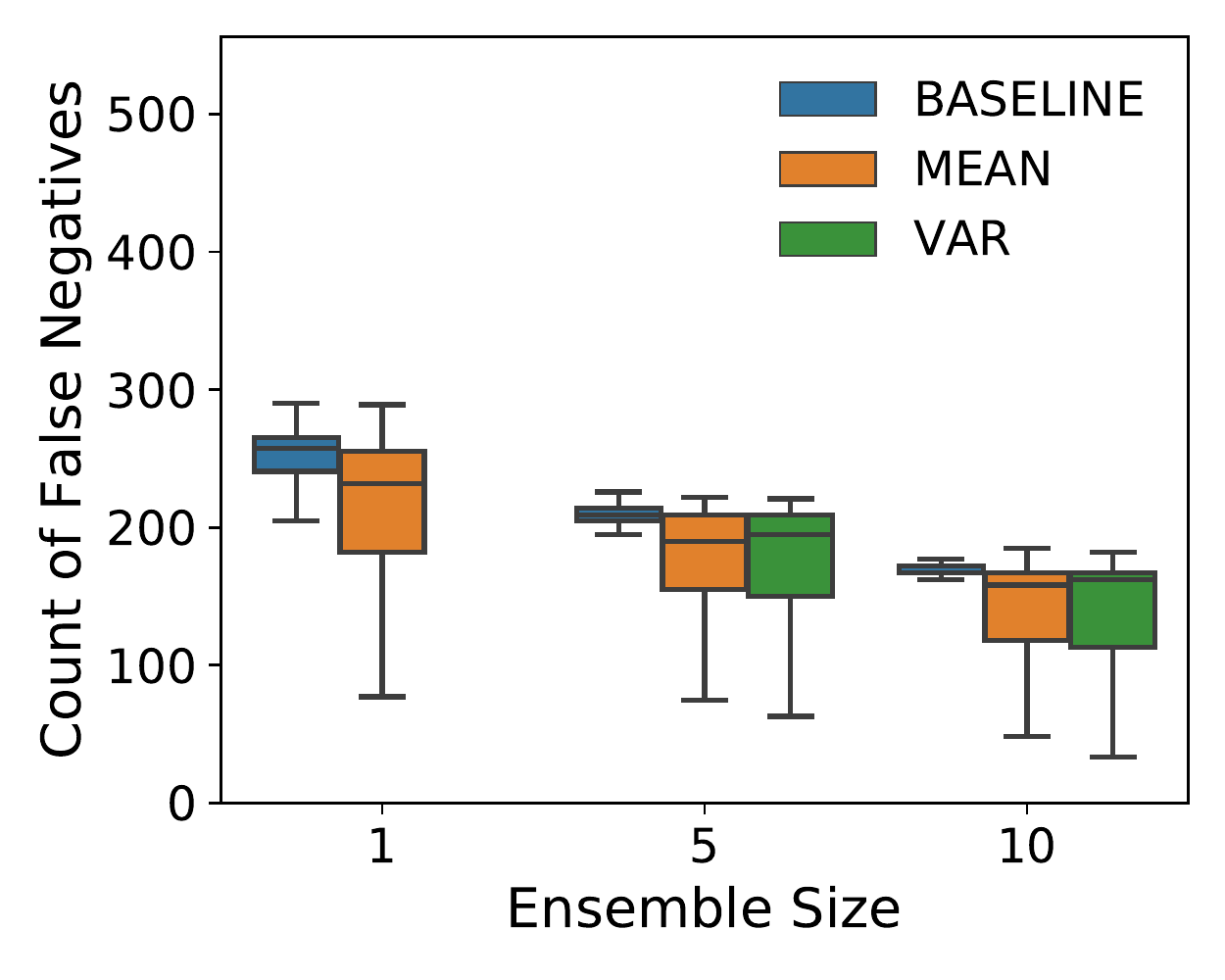}
    \caption{RF: power (FT-4)}
  \end{subfigure}
  \begin{subfigure}[t]{0.23\linewidth}
    \centering
    \includegraphics[height=2.6cm]{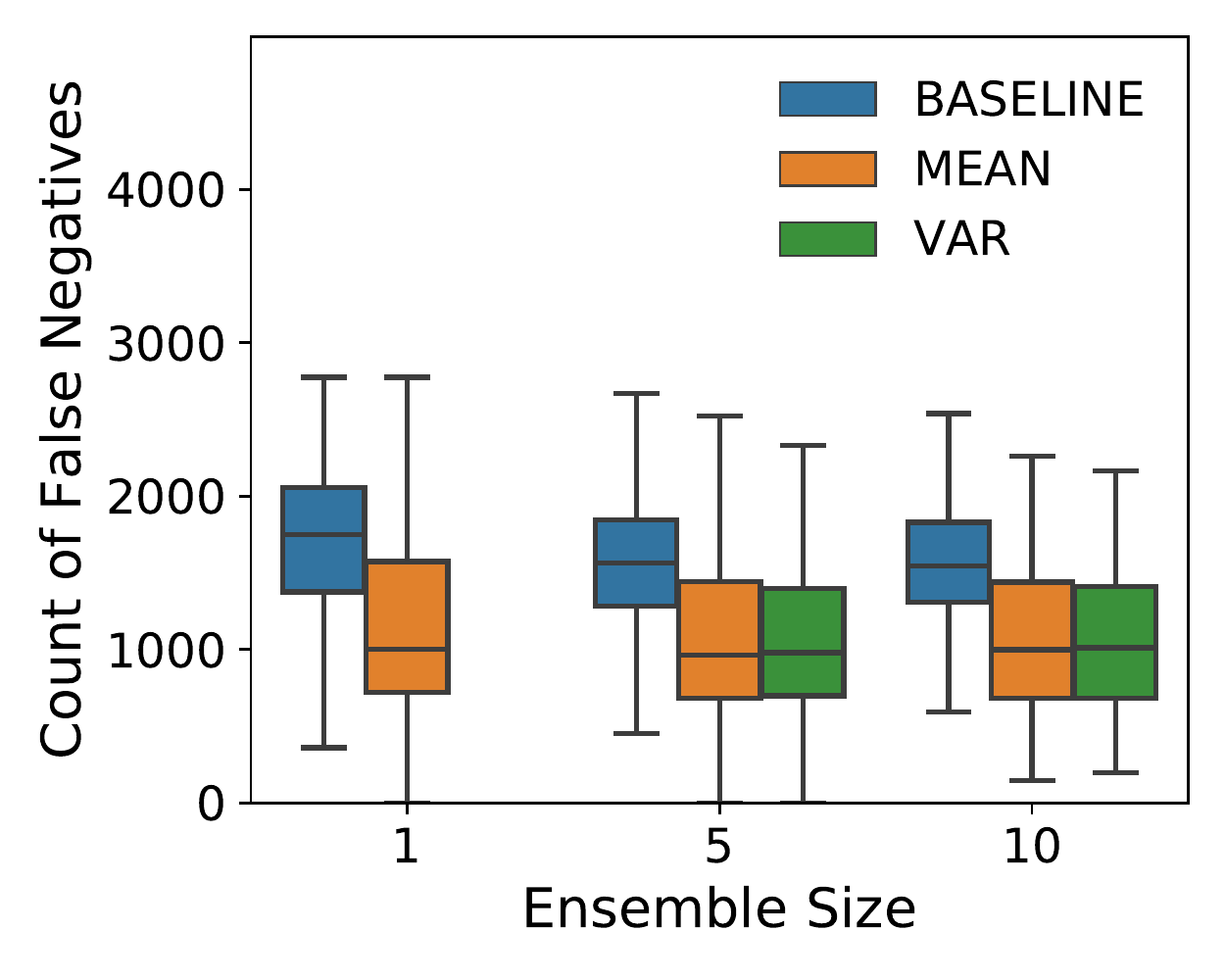}
    \caption{NN: power (FT-4)}
  \end{subfigure}
  
  \begin{subfigure}[t]{0.23\linewidth}
    \centering
    \includegraphics[height=2.6cm]{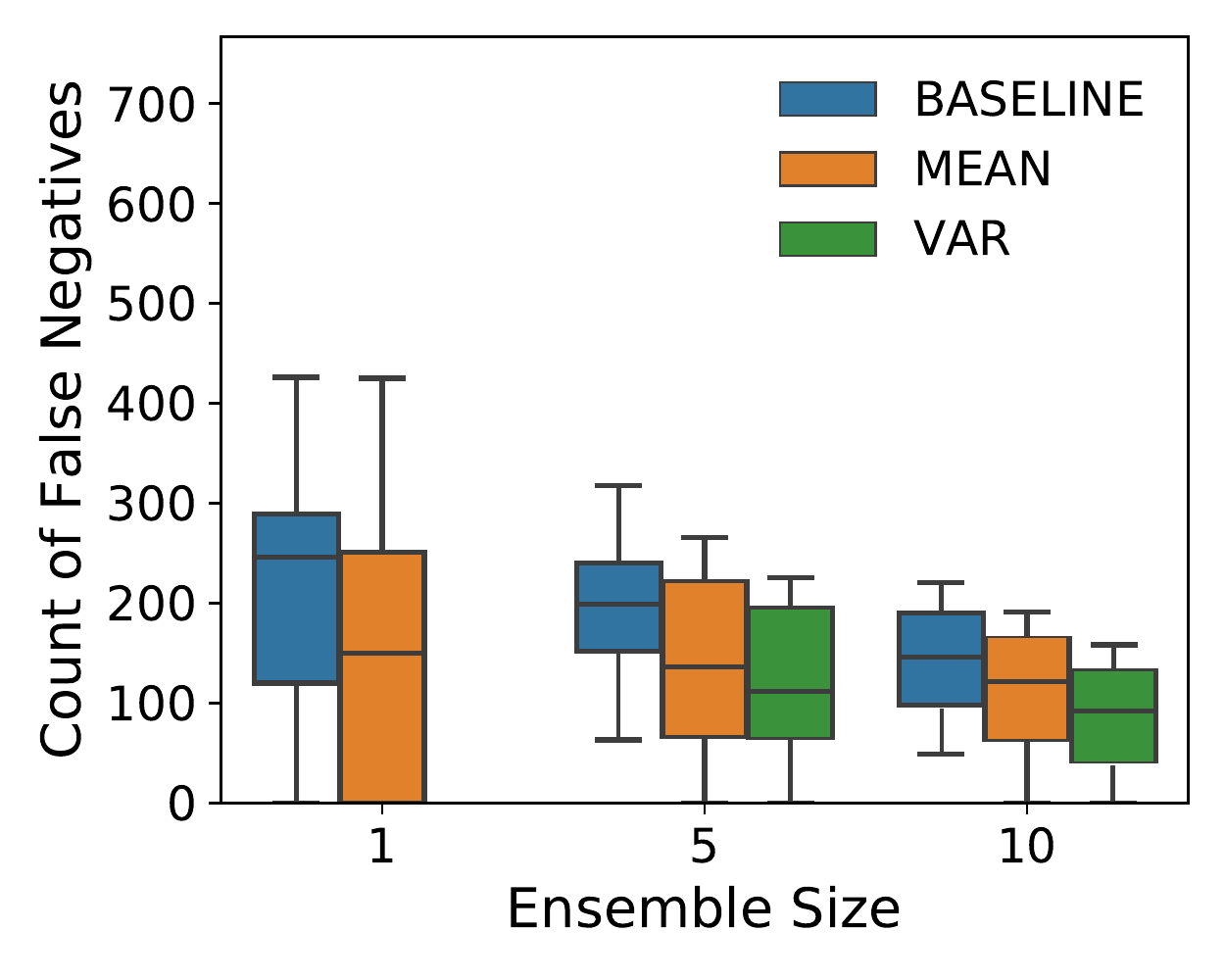}
    \caption{RF: AHU (SU-FT-4)}
  \end{subfigure}
  \begin{subfigure}[t]{0.23\linewidth}
    \centering
    \includegraphics[height=2.6cm]{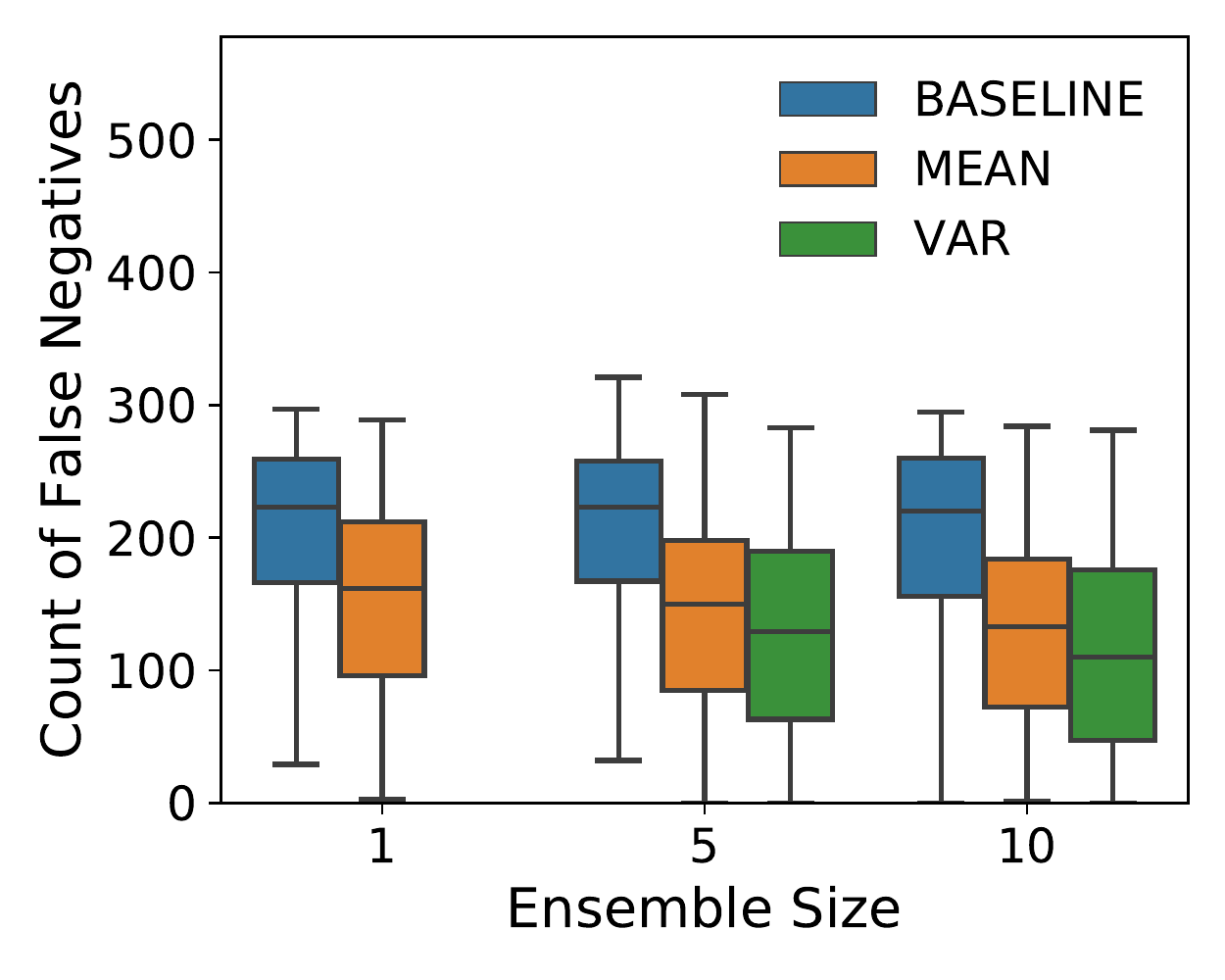}
    \caption{NN: AHU (SU-FT-4)}
  \end{subfigure}
  \begin{subfigure}[t]{0.23\linewidth}
    \centering
    \includegraphics[height=2.6cm]{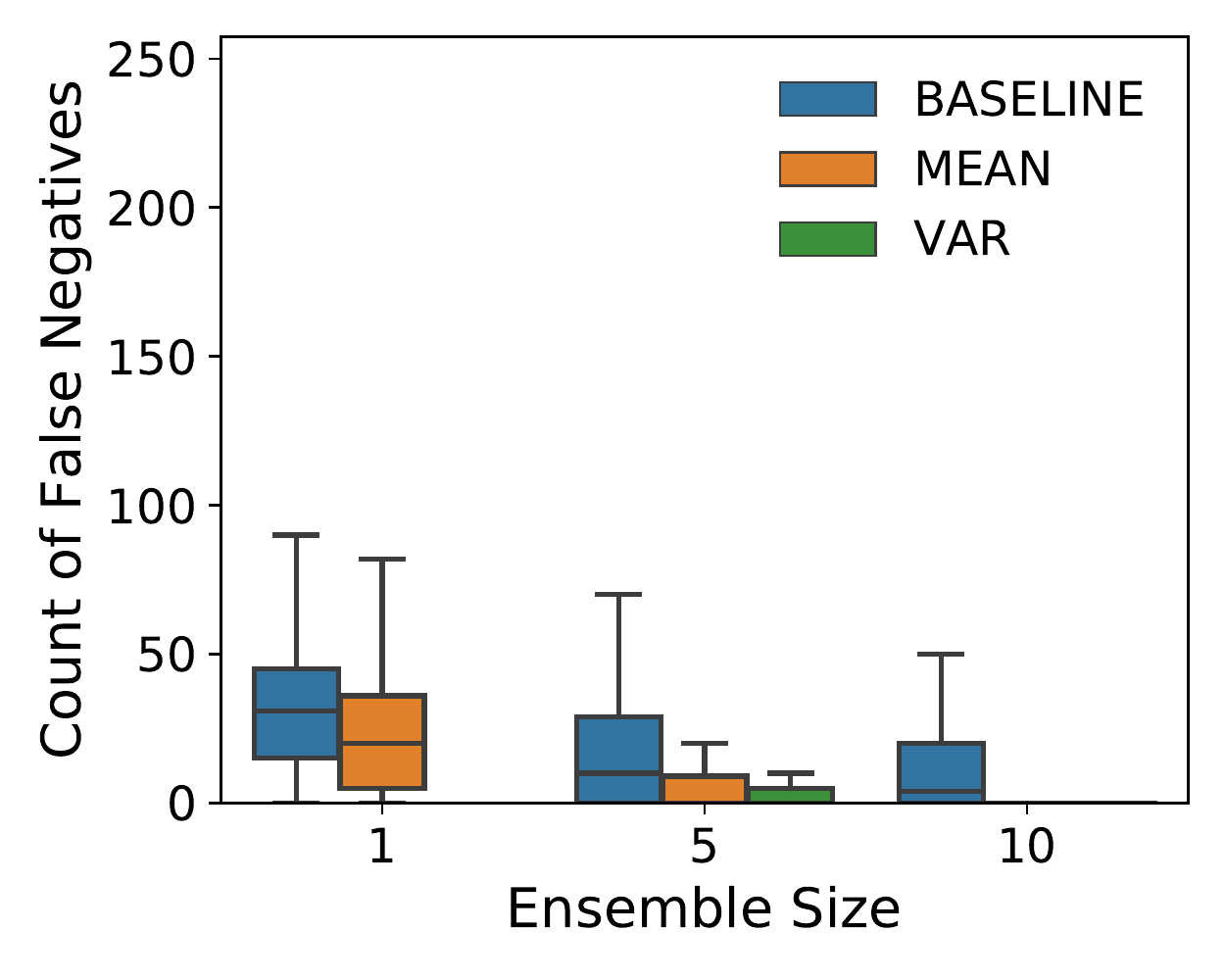}
    \caption{RF: power (LOC-4)}
  \end{subfigure}
  \begin{subfigure}[t]{0.23\linewidth}
    \centering
    \includegraphics[height=2.6cm]{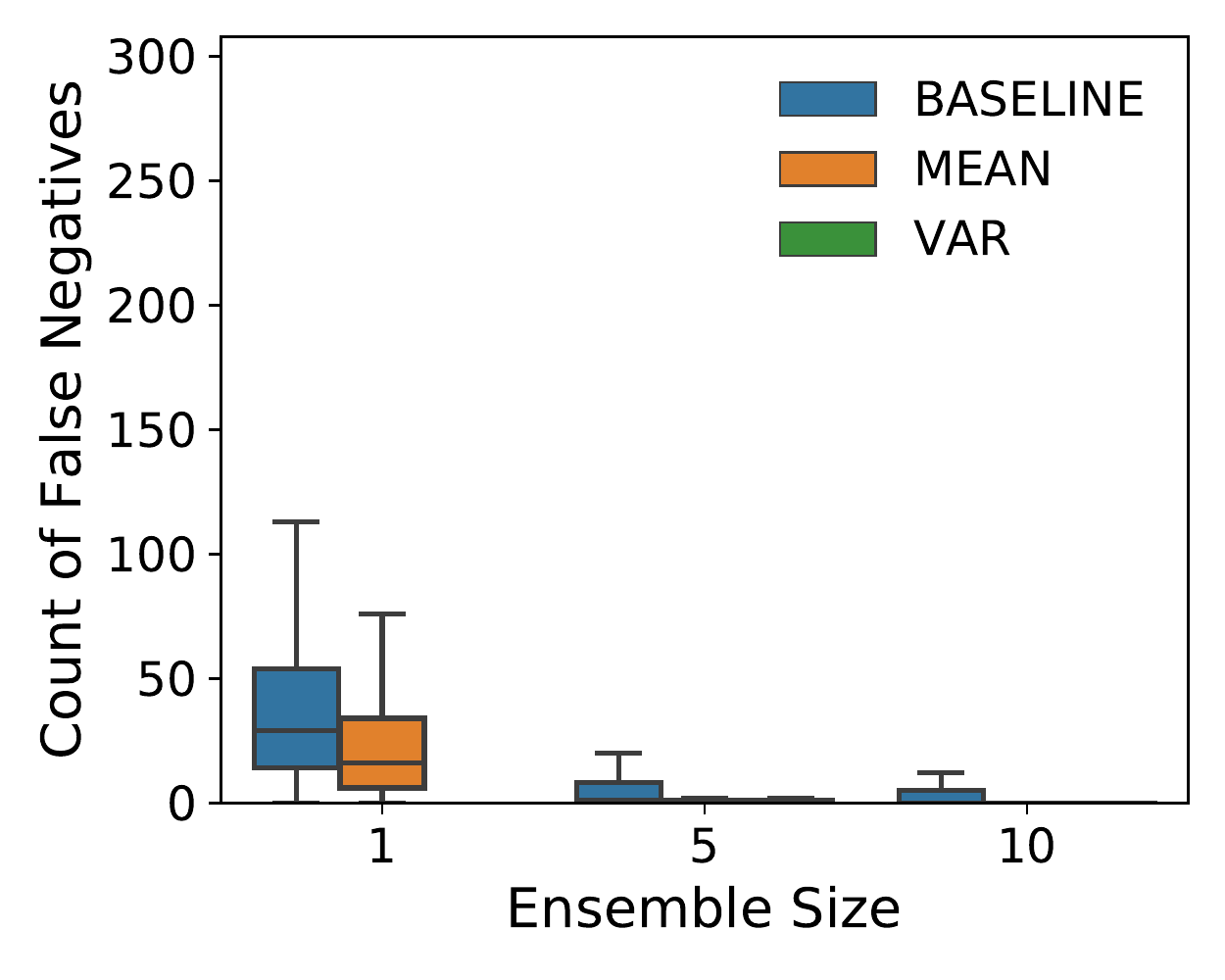}
    \caption{NN: power (LOC-4)}
  \end{subfigure}
  
  \begin{subfigure}[t]{0.23\linewidth}
    \centering
    \includegraphics[height=2.6cm]{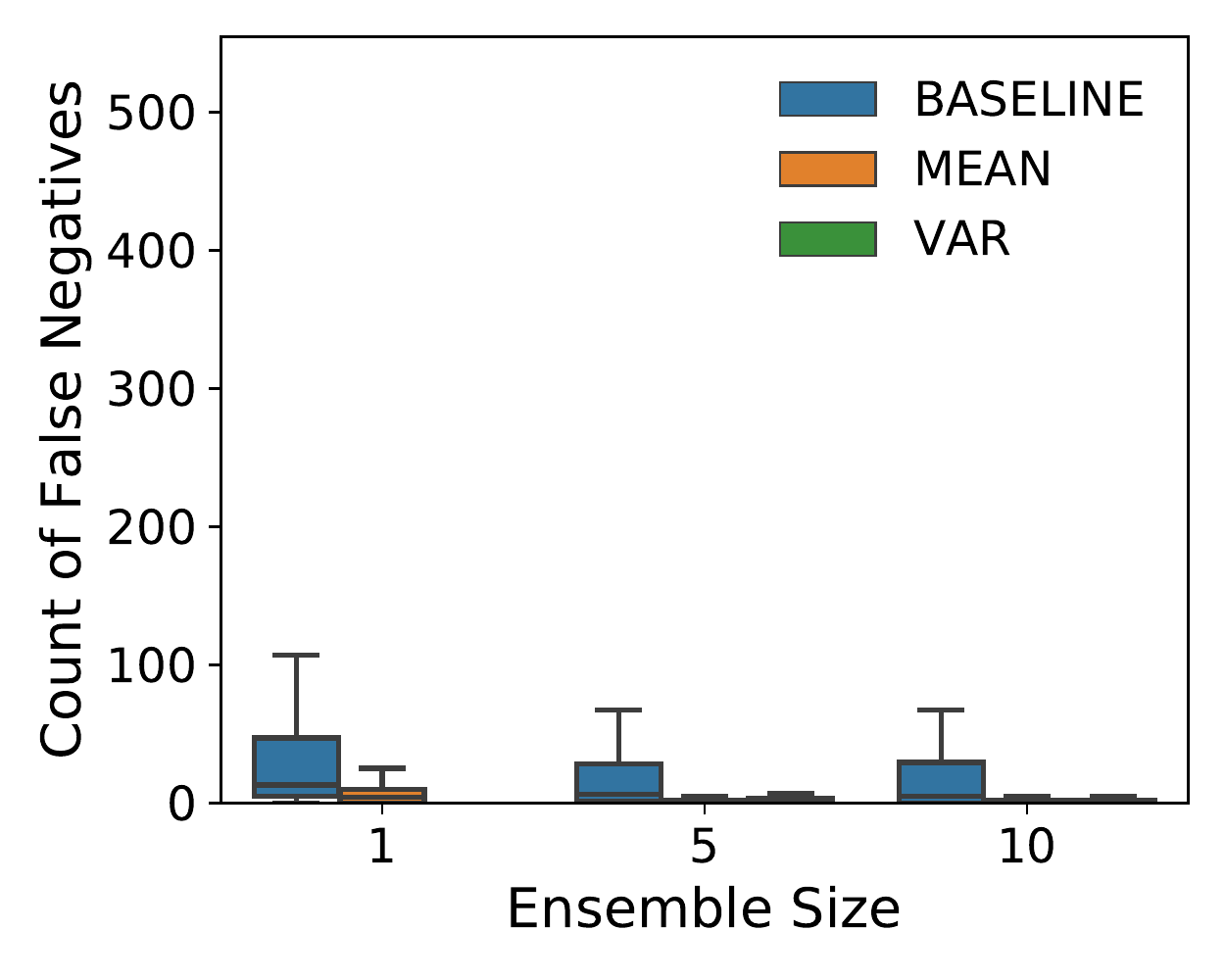}
    \caption{RF: AHU (WT-FT-4)}
  \end{subfigure}
  \begin{subfigure}[t]{0.23\linewidth}
    \centering
    \includegraphics[height=2.6cm]{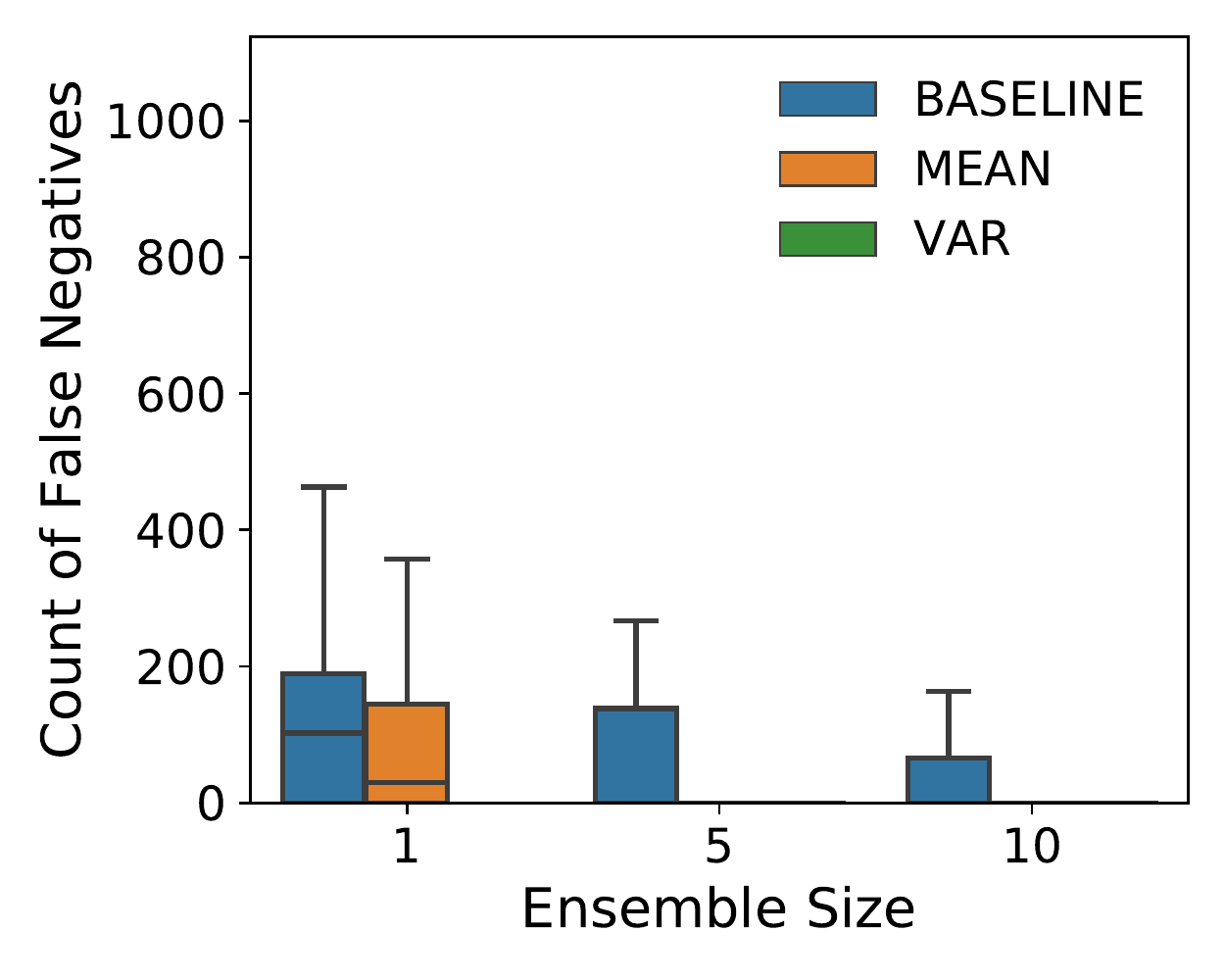}
    \caption{NN: AHU (WT-FT-4)}
  \end{subfigure}
  \begin{subfigure}[t]{0.23\linewidth}
    \centering
    \includegraphics[height=2.6cm]{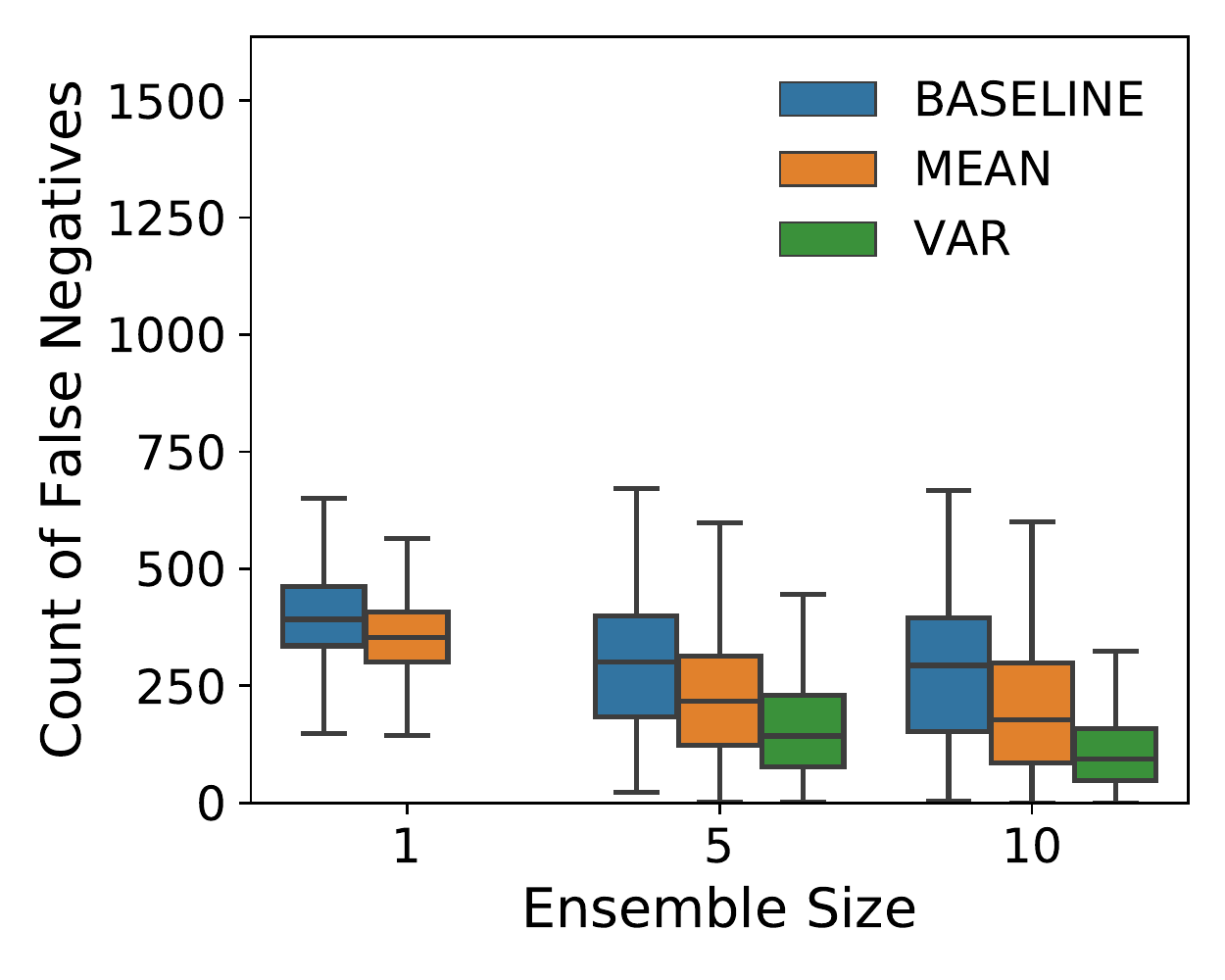}
    \caption{RF: power (RES-2)}
  \end{subfigure}
  \begin{subfigure}[t]{0.23\linewidth}
    \centering
    \includegraphics[height=2.6cm]{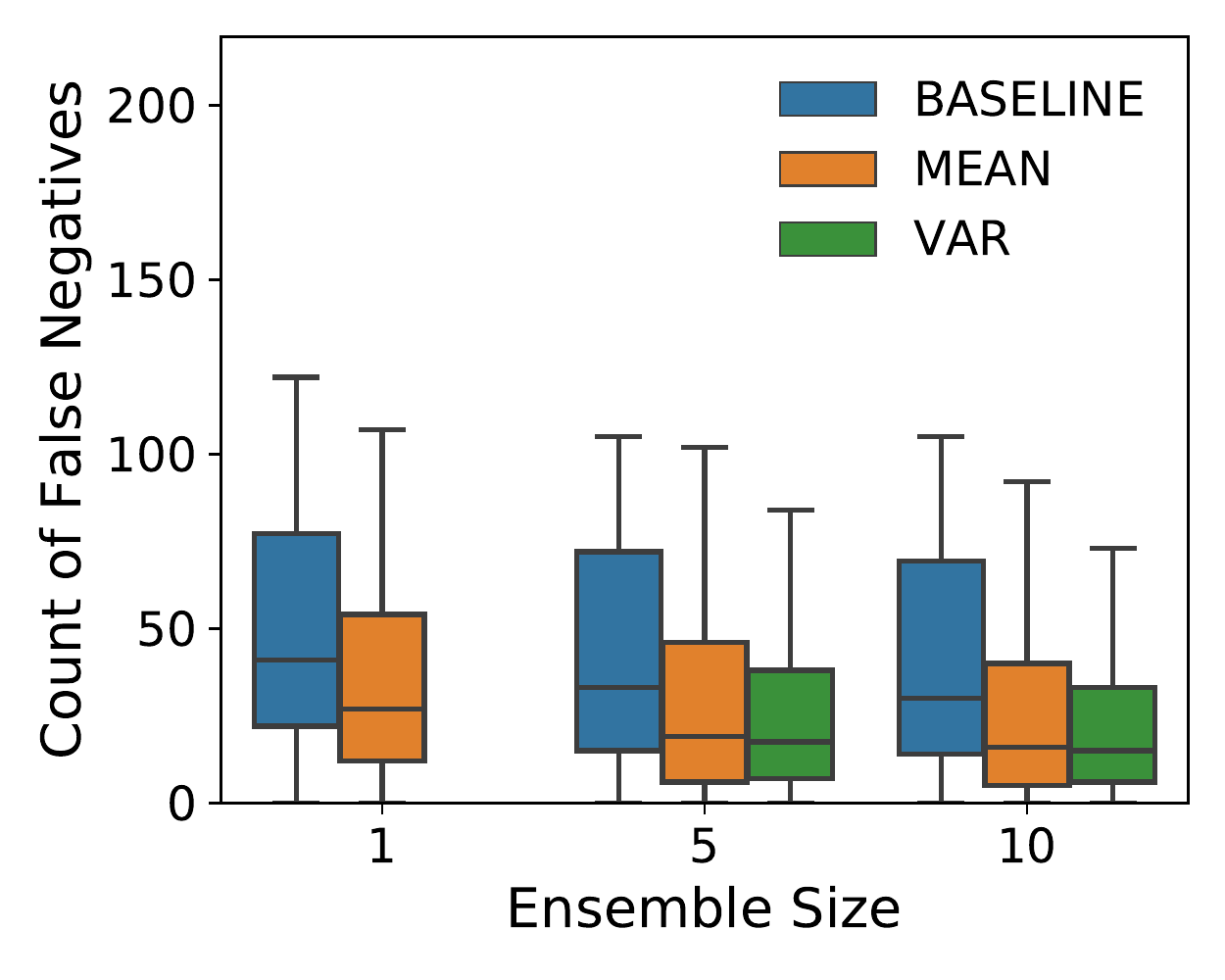}
    \caption{NN: power (RES-2)}
  \end{subfigure}  
  \caption{The count of remaining false negatives under different uncertainty metrics: 1) \textsc{baseline} ($\theta=0$, i.e. no uncertainty information is exploited), 2) \textsc{mean} and 3) \textsc{var}. The results from tree ensembles (\acp{RF}) and \ac{NN} ensembles on the three datasets are presented.}
  \label{fig:cnt-FN-results}
\end{figure}

The results given by \textsc{mean} and by \textsc{var}, as well as the performance baseline where no uncertainty estimation is applied (\textsc{baseline}), are displayed in Fig.~\ref{fig:cnt-FN-results}. As illustrated in the plots, it is clear that both \textsc{mean} and \textsc{var} metrics have decent improvement in identifying false negatives over \textsc{baseline}. Specifically, comparing \textsc{mean} and \textsc{var}, we also found that \textsc{var} outperformed \textsc{mean}, indicating that \textsc{var} excelled at estimating \ac{o.o.d.} data.

Another finding is the \ac{RF} has larger improvement as the ensemble size grows, compared to \ac{NN}. One possible reason for this is that single \ac{NN} classifiers have stronger classification abilities over single \acp{DT} classifiers.

The above results seem to contradict the conclusion from our previous work~\cite{Tan2020ExploitingUF,Jin2020incipient}, where we showed that \textsc{mean} is more preferable to \textsc{var} for identifying incipient anomalies (faults or diseases). It is worth mentioning that our focus in this paper is \ac{o.o.d.} fault data that are not included in the development set during training, rather than incipient faults. We illustrate the differences between the two scenarios in Fig.~\ref{fig:ensemble-illustration}, and how ensemble methods can help with fault detection in both scenarios. It will be interesting future work to understand why \textsc{var} excels at identifying \ac{o.o.d.} data.

% In addition, it is worth mentioning that the experiment results of our project seem to be different from those given by our previous paper~\cite{Tan2020ExploitingUF}. The reason for this is that the experiments of this paper have different concentration with  This paper focus on the \ac{o.o.d.} While our previous paper focus on detecting the incipient fault data which only has only milder severity levels than the fault data. There are no conflicts between these two works and both results make sense. \yingshui{need to be rephased to be more professional}

% % no need to reset, ok to use acronym in captions

\section{Related Work}\label{sec:related-work}

\subsection{Adversarial Validation}\label{sec:adversarial-validation}
A closely related technique that also deals with the domain shift phenomenon between training and test distributions is the \textit{adversarial validation} approach~\cite{pan2020adversarial} whose goal is to detect and address the difference between the training and test datasets. The idea of adversarial validation is to create an adversarial validation set as an \textit{proxy} of the test set for selecting robust models during training such that the resulting model can achieve satisfactory performance on the adversarial validation set (and hopefully on the test set as well). 

The creation of effective adversarial validation sets, however, will usually require prior information about the test data distribution. In some occasions, for example in Kaggle competitions, part of the test set data is made public at training time while the rest is used as a  ``private test set''. In adversarial validation approaches, a classifier is trained to distinguish the training and the (public) test set data, and then part of the training data (e.g., the difficult-to-classify ones) that resembles the test data can be held out as an adversarial validation set. Such approach is described as the ``validation data selection'' method in Pan~et~al.'s recent work~\cite{pan2020adversarial}, which also describes other types of adversarial validation methods; see details therein for further information. 

In fact, the ``public test set'' data mentioned above can also be considered as part of the development set (because the public test set data are available at training time), and thus not actually a ``real'' test set. In situations where little information about the test data distribution is available, adversarial validation will not be applicable. Our \ac{SACV} approach does not require prior information about the unseen test distribution. Instead, this approach rely only on the available development set data. By holding out one fault subgroup (stratum) at a time, we seek to build a model that is most robust against possible adversarial fault examples.

\subsection{Out-of-Distribution Data Detection}\label{sec:ood-detection}
In recent years, a number of research papers~\cite{lakshminarayanan2017simple,gal2016uncertainty} related to the detection of \ac{o.o.d.} data are seen in literature. Lakshminarayanan~et~al.~\cite{lakshminarayanan2017simple} proposed using \textit{random initialization} and \textit{random shuffling} of training examples to diversify base learners of the same network architecture. Gal and Ghahramani proposed using MC-dropout~\cite{gal2016uncertainty} to estimate a network's prediction uncertainty by using dropout not only at training time but also at test time. By repeatedly sampling a dropout model $\mathcal{M}$ using the same input for $T$ times, we can obtain an ensemble of prediction results with $T$ individual probability vectors. The dropout technique provides an inexpensive approximation to training and evaluating an ensemble of exponentially many similar yet different neural networks.

\subsection{Incipient Anomaly Detection}\label{sec:incipient-anomaly-detection}
Another application of decision uncertainty and ensemble methods is the detection of incipient anomalies (e.g., industrial machine faults and human diseases). Incipient anomalies~\cite{jin2019detecting} present milder symptoms compared to severe ones, and can be easily mistaken as the normal operating conditions due to their close resemblance to normal operating conditions. The lack of incipient anomaly examples in the training data can pose severe risks to anomaly detection methods that are built upon \ac{ML} techniques. To address this challenge, the authors of~\cite{jin2019detecting,jin2020ensemble,Tan2020ExploitingUF} propose to utilize the uncertainty information from ensemble learners to identify potential misclassified incipient anomalies, and show that the uncertainty-informed detection scheme gives improved results on incipient anomalies without sacrificing performance on non-incipient anomaly examples. 

\section{Conclusion}\label{sec:conclusion}
In this paper, we showed that domain shift in stratified data can undermine fault detection performance, especially when some subgroups (strata) appear in the test data distribution but not in the training distribution. We proposed an easy-to-use cross-validation method to mitigate the issue and demonstrated its efficacy on three representative \ac{CPS} datasets. Our proposed \ac{SACV} approach achieved significant performance improvement over traditional holdout and $k$-fold validation methods on \ac{o.o.d.} data, in the meantime without sacrificing its performance on \ac{i.d.} data. For future work, we plan to extend the proposed methodology to datasets of different modalities, such as image data.

%%
%% The next two lines define the bibliography style to be used, and
%% the bibliography file.
\bibliographystyle{ACM-Reference-Format}
\bibliography{refs}

% \begin{acks}
% This work is supported by the National Research Foundation of Singapore through a grant
% to the Berkeley Education Alliance for Research in Singapore (BEARS) for the Singapore-Berkeley Building Efficiency and Sustainability in the Tropics (SinBerBEST) program, and by the Defence Science \& Technology Agency (DSTA) of Singapore.
% \end{acks}

%%
%% If your work has an appendix, this is the place to put it.
\appendix
\clearpage
\acresetall

\section{RP-1043 Chiller Dataset}\label{sec:app-chiller}
\begin{figure}[b]
    \centering
    \includegraphics[width=0.85\linewidth]{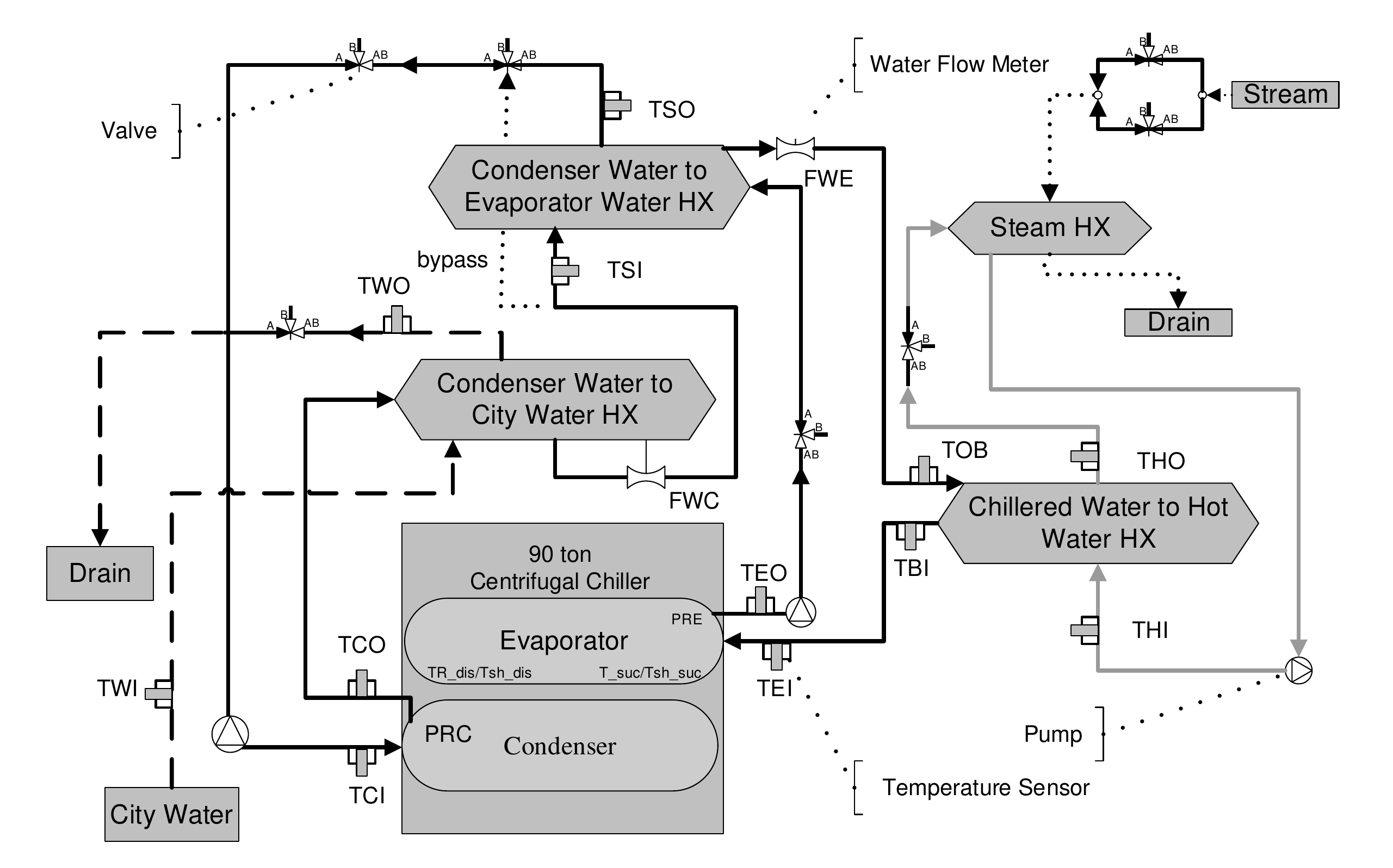}
    \caption{A schematic of the cooling system test facility and sensors mounted in the related water circuits~\cite{li2017fault}.}
    \label{fig:chiller-sensor}
\end{figure}

The RP-1043 chiller dataset~\cite{comstock1999development} is not public but is available for purchase from ASHRAE. The 90-ton chiller studied in the RP-1043 chiller dataset is representative of chillers used in larger installations~\cite{Comstock2002fault}, and consisted of the following parts: evaporator, compressor, condenser, economizer, motor, pumps, fans, and distribution pipes etc. with multiple sensor mounted in the system. Fig.~\ref{fig:chiller-sensor} depicts the cooling system with sensors mounted in both evaporation and condensing circuits.

The same sixteen features and six fault types as used in our previous work~\cite{tan2019encoder} were used to train our models in our empirical study. We also attempted to use other sets of selected features than the sixteen features in our case study, e.g., the features identified in Li~et~al.'s previous work~\cite{li2016fault}, and similar results were obtained. Table~\ref{tbl:features} and Table~\ref{tbl:faults} give detailed descriptions of the sixteen features and the six fault types used in this study, respectively. In the chiller dataset, each fault was introduced at four \acp{SL}, and we included fault data of all four \acp{SL} in the dataset for our experiment. The condenser fouling (FT-CF) fault was emulated by plugging tubes into condenser. The reduced condenser water flow rate (FT-FWC) fault and reduced evaporator water flow rate (FT-FWE) fault were emulated directly by reducing water flow rate in the condenser and evaporator. The refrigerant overcharge (FT-RO) fault and refrigerant leakage (FT-RL) fault were emulated by reducing or increasing the refrigerant charge respectively. The excess oil (FT-EO) fault was emulated by charging more oil than nominal. And the non-condensable in refrigerant (FT-NC) fault was emulated by adding Nitrogen to the refrigerant.

\begin{table}[tb]
\caption{Descriptions of variables used as features in the chiller dataset}
\label{tbl:features}
\begin{tabular}{lll}
\hline
Sensor & Description & Unit \\ \hline
TEI & Temperature of entering evaporator water & \degree F \\
TEO & \begin{tabular}[c]{@{}l@{}}Temperature of leaving evaporator water\end{tabular} & \degree F \\
TCI & \begin{tabular}[c]{@{}l@{}}Temperature of entering condenser water\end{tabular} & \degree F \\
TCO & \begin{tabular}[c]{@{}l@{}}Temperature of leaving condenser water\end{tabular} & \degree F \\
Cond Tons & \begin{tabular}[c]{@{}l@{}}Calculated Condenser Heat Rejection Rate\end{tabular} & Tons \\
Cooling Tons & Calculated City Water Cooling Rate & Tons \\
kW & Compressor motor power consumption & kW \\
FWC & Flow Rate of Condenser Water & gpm \\
FWE & Flow Rate of Evaporator Water & gpm \\
PRE & \begin{tabular}[c]{@{}l@{}}Pressure of refrigerant in evaporator\end{tabular} & psig \\
PRC & \begin{tabular}[c]{@{}l@{}}Pressure of refrigerant in condenser\end{tabular} & psig \\
TRC & Subcooling temperature & \degree F \\
T\_suc & Refrigerant suction temperature & \degree F \\
Tsh\_suc & \begin{tabular}[c]{@{}l@{}}Refrigerant suction superheat temperature\end{tabular} & \degree F \\
TR\_dis & Refrigerant discharge temperature & \degree F \\
Tsh\_dis & \begin{tabular}[c]{@{}l@{}}Refrigerant discharge superheat temperature\end{tabular} & \degree F \\ \hline
\end{tabular}
\end{table}

\begin{table}[tb]
\caption{The six chiller faults in our study}
\label{tbl:faults}
\begin{tabular}{lll}
\hline
\textbf{Fault Types} & \textbf{Identity} & \textbf{Normal Operation} \\ \hline
Reduced Condenser Water Flow & FT-FWC & 270 gpm \\
Reduced Evaporator Water Flow & FT-FWE & 216 gpm \\
Refrigerant Leak & FT-RL & 300 lb \\
Refrigerant Overchange & FT-RO & 300 lb \\
Condenser Fouling & FT-CF & 164 tubes \\
Non-condensables in System & FT-NC & No nitrogen \\ \hline
\end{tabular}%
\end{table}

\section{RP-1312 AHU Dataset}\label{sec:app-AHU}
The \acs{AHU} dataset included 16 fault types in total that are distributed across three seasons: spring, summer and winter. A detailed list of the 16 fault types studied by the \acs{AHU} dataset is given in Table~\ref{tab:AHU-faults}, where each fault is assigned a unique identifier. We can also see that the faults appearing in different seasons do not fully overlap; there are faults that exist only in spring but not in summer or winter (e.g., SP-FT-1) and also faults that appear in all three seasons such as the ``exhaust air damper stuck'' fault. When building \ac{ML} models for each season, we only considered faults that appear in that season. For example, the fault (positive) class for the AHU-spring dataset encompasses $11$ faults; these fault types constituted the $11$ subgroups (strata) in the fault class.

\begin{figure}[tb]
  \begin{subfigure}[t]{0.60\linewidth}
    \centering
    \includegraphics[height=5.5cm]{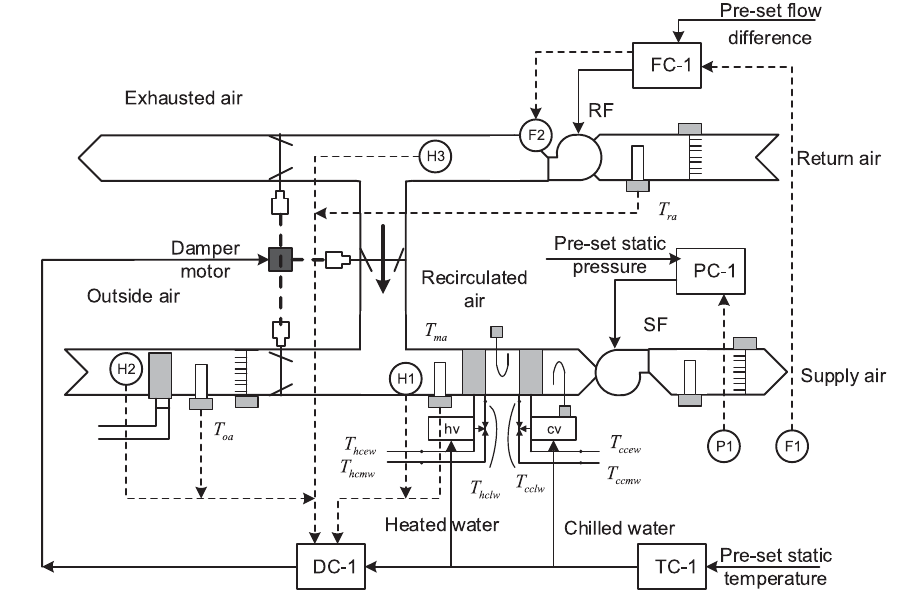}
    \caption{\acs{AHU} system schematic}
    \label{fig:AHU-schematic}
  \end{subfigure}
  \begin{subfigure}[t]{0.35\linewidth}
    \centering
    \includegraphics[height=5.5cm]{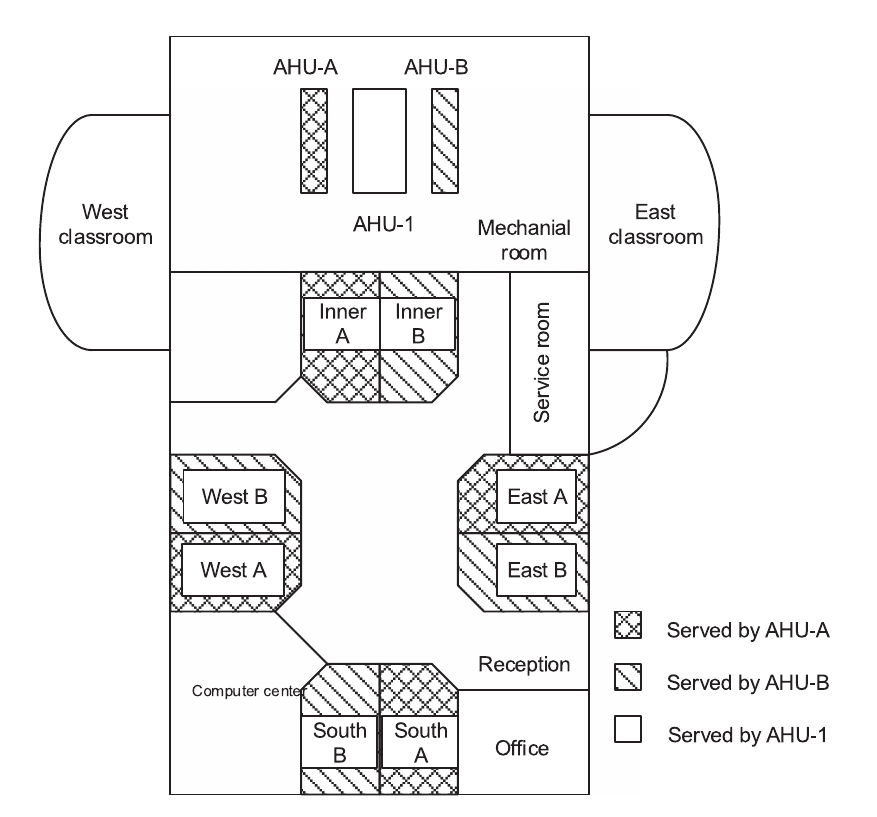}
    \caption{RP-1312 testing site layout}
    \label{fig:AHU-testing-site}
  \end{subfigure}
  \caption{We show in (a) a typical single-duct \ac{VAV} \acs{AHU} system~\cite{li2017fault}, and in (b) the schematic of the testing site used for creating the RP-1312 \acs{AHU} Dataset~\cite{wen2012rp}.}
\end{figure}

The schematic of a typical \ac{AHU} system is shown in Fig.~\ref{fig:AHU-schematic} that is configured for a \ac{VAV} system. The \ac{VAV} system maintains the supply air temperature to the terminals for air-conditioning. The testing site for creating the RP-1312 AHU Dataset involved two \acp{AHU}, i.e., AHU-A and AHU-B as shown in Fig.~\ref{fig:AHU-testing-site} that were operated under real weather and building load conditions. Faults were manually introduced into the air-mixing box, the coils, and the fan sections of AHU-A (treatment group), while AHU-B was operated at nominal states to serve as the control group.

% Details
\begin{table}[tb]
\centering
\caption{Fault Types Studied in the RP-1312 \ac{AHU} Dataset}
\label{tab:AHU-faults}
\resizebox{0.8\textwidth}{!}{%
\begin{tabular}{llll}
\hline
\textbf{Fault Types} & \textbf{Spring} & \textbf{Summer} & \textbf{Winter} \\ \hline
\multicolumn{1}{l}{Outside air damper leak} &  & \textit{SU-FT-1} & \textit{WT-FT-1} \\
Outside air temperature sensor bias & \textit{SP-FT-1} &  &  \\
Outside air damper stuck & \textit{SP-FT-2} &  & \textit{WT-FT-2} \\
Exhaust air damper stuck & \textit{SP-FT-3} & \textit{SU-FT-2} & \textit{WT-FT-3} \\
Cooling coil valve control unstable & \textit{SP-FT-4} & \textit{SU-FT-3} &  \\
Cooling coil valve reverse action &  & \textit{SU-FT-4} &  \\
Cooling coil valve stuck & \textit{SP-FT-5} & \textit{SU-FT-5} & \textit{WT-FT-4} \\
Heating coil valve leaking &  & \textit{SU-FT-6} &  \\
Return fan at fixed speed & \textit{SP-FT-6} & \textit{SU-FT-7} &  \\
Return fan complete failure & \textit{SP-FT-7} & \textit{SU-FT-8} &  \\
Air filter area block fault & \textit{SP-FT-8} &  &  \\
Mixed air damper unstable & \textit{SP-FT-9} &  &  \\
Sequence of  heating and cooling unstable & \textit{SP-FT-10} &  &  \\
Supply fan control unstable & \textit{SP-FT-11} &  &  \\
Heating coil fouling &  &  & \textit{WT-FT-5} \\
Heating coil reduced capacity &  &  & \textit{WT-FT-6} \\ \hline
\end{tabular}%
}
\end{table}

\section{Power System Faults Dataset}\label{sec:app-power}

% \subsection{Feeder Loading}\label{sec:feederloading}

% % Table \ref{tab:feederloading} shows the feeder loading per phase of the system under study. 

% \begin{table}[hb]
% 	\renewcommand{\arraystretch}{1.3}
% 	\caption{System Configuration under Different DER Technologies.}
% 	\label{tab:feederloading}
% 	\centering
% 	\begin{tabular}{ccc}
% 		\hline
% 		\hline
% 		Phase & Active power (kW) & Reactive power (kVar)\\
% 		\hline
% 		A & $3,297$ & $745$\\
% 		\hline
% 		B & $3,052$ & $671$\\
% 		\hline
% 		C & $4,425$ & $987$\\		
% 		\hline
% 		Total & $10,774$ & $2,403$\\
% 		\hline
% 		\hline
% 	\end{tabular}
% \end{table}
% %\vspace{-3mm}

% \subsection{Benchmark System}
% \label{sec:benchmarksys}

\begin{figure}[!htb]
	\centering
	\includegraphics[width=5.5in]{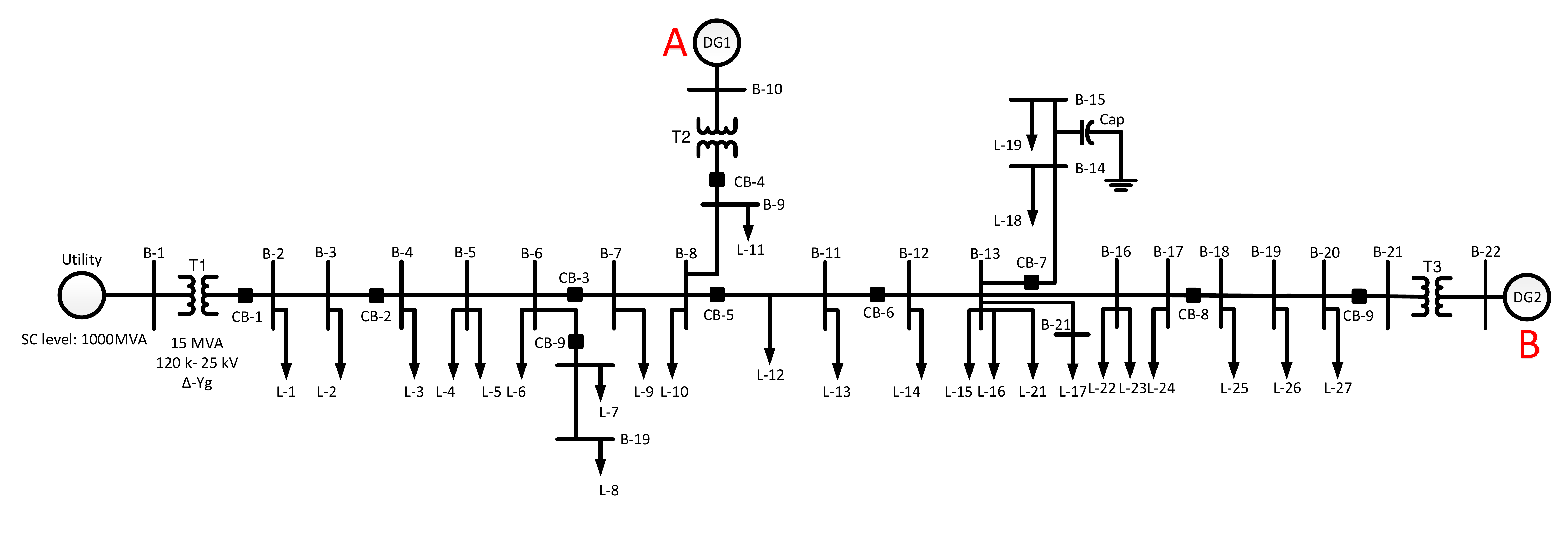}
	\centering
	\caption{A single line diagram of the distribution feeder under study \cite{cui2019feature}.}
	\label{fig:feeder_HIF}
\end{figure}

The benchmark system that generates this dataset can be found in Fig.~\ref{fig:feeder_HIF}. In this system, we simulate three system configuration under different distributed energy resource (DER) technologies, namely, synchronous-machine-based-system (synchronous machine at location A), inverter-based system (the inverter-interfaced wind farm at location A), and the hybrid system (synchronous machine at location A and the inverter-interfaced wind farm at location B). The wind farm is type 4 and rated at 575 V, 6.6 MVA. According to IEEE Standard~1547, the wind farm adopts constant power control with LVRT capability. The maximum fault current is limited to $1.5$~pu.

% \subsubsection{Events Under Study}
% \label{sec:syseve}

Table~\ref{tab:eventcategory} lists the event category and event type under study. The event numbers are explained as follows. First, $8$ loading conditions and $3$ DER technologies are examined respectively on top of the base case scenario. Furthermore, the $10$ events from Type 1 is associated with the undowned conductor, where $3$ SLG (AG, BG, CG), $3$ LLG (ABG, ACG, BCG), $3$ LL (AB, BC, AC), and $1$ LLLG (ABCG) faults are included. The $3$ events of Type 2 fault are the downed conductor for each phase. The fault impedance values includes $50, 150, 250, 350, 450$, and \SI{550}{\Omega} in this paper. In load switching, the $6$ types of non-fault events include $4$ single load switching (L-4, L-9, L-19, L-23) and $2$ combinational load switching ((L-2,L-4,L-5) and (L-9, L-10)) events. Additionally, the $2$ capacitor switching events have both the on and off status of the capacitor bank near bus B-15. A loading condition ranging from $30\%$ to $100\%$, in a step of $10\%$, is simulated.

% Therefore, the number of fault and non-fault events are calculated as follows:
% \begin{itemize}
%   \item Fault event: since two types of fault, summing up to $13$ cases, are included, the number of fault events with one fault impedance, one fault location and one fault impedance is $(10+3)\times8\times3=312$. Given $6$ simulated fault impedance values, $4$ fault inception angles, and $3$ fault locations, the total number of fault events add up to $312\times6\times4\times3=22464$.
%   \item Non-fault event: it comprises normal state, load switching (adding and shedding) and capacitor switching events. Therefore the total number of non-fault events equals to $(1+6+2)\times8\times3=216$. 
% \end{itemize}

% The above event number results in an imbalanced dataset, where the number of data points belonging to the minority class (``non-fault``) is far smaller than the number of the data points belonging to the majority class (``fault``). Under this circumstance, an algorithm gets insufficient information about the minority class to make an accurate prediction. Therefore, the synthetic minority over-sampling technique (SMOTE) is employed to generate synthetic samples and shift the classifier learning bias towards minority class \cite{chawla2002smote}.

%Explain the unbalanced load system. 
%In Table \ref{tab:sysconfig} and Table \ref{tab:DISI}.
%Explain the unbalanced load system. 

\begin{table}[!hbt]
%\vspace{-4mm}
\caption{Event Category of the System Under Study \cite{cui2019feature}.}
\label{tab:eventcategory}
\centering
\begin{tabular}{lll}
\hline
    Event Category & Event Type & Event Nunber \\ \hline
    \multirow{2}{*}{\begin{tabular}[c]{@{}l@{}}System Operating Condition\end{tabular}} & Loading Condition (30\%-100\%) & 8 \\
     & DER Tech. (SG, inverter, hybrid) & 3 \\ \hline
    \multirow{5}{*}{Fault Event} & Type 1: SLG, LLG, LL, LLLG & 10 \\
     & Type 2: Downed conductor & 3 \\
     & Fault impedance & 6 \\
     & Inception Angle ($\SI{0}{\degree}$, $\SI{30}{\degree}$, $\SI{60}{\degree}$, $\SI{90}{\degree}$) & 4 \\
     & Fault location & 3 \\ \hline
    \multirow{3}{*}{Non-fault Event} & Normal State & 1 \\
     & Load Switching & 6 \\
     & Capacitor Switching & 2 \\ \hline
\end{tabular}%
\end{table}

\end{document}